\documentclass{article}

\usepackage{microtype}
\usepackage{graphicx}
\usepackage{subcaption}
\usepackage{booktabs} 
\usepackage{glossaries}%
\usepackage{pifont}
\usepackage{wrapfig}
\usepackage{enumitem}
\usepackage{hyperref}

\usepackage[preprint]{icml2026}

\usepackage{amsmath}
\usepackage{amssymb}
\usepackage{mathtools}
\usepackage{amsthm}
\usepackage{tabularx}
\usepackage{multirow}
\usepackage{listings}
\usepackage[capitalize,noabbrev]{cleveref}

\newcommand{\anon}[1]{\textcolor{orange}{ANONYMIZED}}

\definecolor{red}{RGB}{213, 62, 79}
\definecolor{orange}{RGB}{244, 109, 67}
\definecolor{lightorange}{RGB}{253, 174, 97}
\definecolor{yellow}{RGB}{254, 224, 139}
\definecolor{dataprep}{RGB}{38, 189, 128}
\definecolor{modeltraining}{RGB}{38, 189, 128}
\definecolor{backgroundcol}{RGB}{20, 20, 20}
\definecolor{gold}{rgb}{225, 215, 200} 
\definecolor{navyblue}{rgb}{0, 0, 128} 

\usepackage{bbding}
\definecolor{darkgreen}{RGB}{0, 100, 0}
\newcommand\greencheck{\textcolor{darkgreen}{\ding{52}}}

\newcommand{\TokenCircle}[1][0.9]{%
    \tikz[baseline=(char.base)]\node[shape=rectangle,draw=black,inner sep=2pt,line width=1pt,fill=red,text=white,scale=#1] (char) {Token};\hspace{-1pt}
}
\newcommand{\SPCircle}[1][0.9]{%
    \tikz[baseline=(char.base)]\node[shape=rectangle,draw=black,inner sep=2pt,line width=1pt,fill=orange,text=white,scale=#1] (char) {SP};\hspace{-1pt}
}
\newcommand{\CrossCircle}[1][0.9]{%
    \tikz[baseline=(char.base)]\node[shape=rectangle,draw=black,inner sep=2pt,line width=1pt,fill=lightorange,text=white,scale=#1] (char) {XAttn};\hspace{-1pt}
}
\newcommand{\AdapterCircle}[1][0.92]{%
    \tikz[baseline=(char.base)]\node[shape=rectangle,draw=black,inner sep=2pt,line width=1pt,fill=yellow,text=white,scale=#1] (char) {Adapter};\hspace{-1pt}
}

\newcommand{\TextReasoningCircle}[1][0.92]{%
    \tikz[baseline=(char.base)]\node[shape=rectangle,draw=black,inner sep=2pt,line width=1pt,fill=navyblue,text=white,scale=#1] (char) {TR};\hspace{-1pt}
}
\newcommand{\ClassificationCircle}[1][0.92]{%
    \tikz[baseline=(char.base)]\node[shape=rectangle,draw=black,inner sep=2pt,line width=1pt,fill=modeltraining,text=white,scale=#1] (char) {CL};\hspace{-1pt}
}

\newcommand{\ForecastingCircle}[1][0.92]{%
    \tikz[baseline=(char.base)]\node[shape=rectangle,draw=black,inner sep=2pt,line width=1pt,fill=backgroundcol,text=white,scale=#1] (char) {FC};\hspace{-1pt}
}

\usepackage{adjustbox,float}
\usepackage{array,multirow,graphicx}

\newcolumntype{R}[2]{%
    >{\adjustbox{angle=#1,lap=\width-(#2)}\bgroup}%
    l%
    <{\egroup}%
}

\theoremstyle{plain}

\theoremstyle{definition}

\theoremstyle{remark}

\usepackage[textsize=tiny]{todonotes}

\icmltitlerunning{OpenTSLM: Time-Series Language Models for Reasoning over Multivariate Medical Text- and Time-Series Data}

\makeglossaries

\newacronym{EHR}{EHR}{electronic health record}
\newacronym{LLM}{LLM}{large language model}
\newacronym{HAR}{HAR}{human activity recognition}
\newacronym{ECG}{ECG}{electrocardiogram}
\newacronym{EEG}{EEG}{electroencephalogram}
\newacronym{CoT}{CoT}{chain of thought}
\newacronym{PPG}{PPG}{photoplethysmography}
\newacronym{EmbedHealth}{OpenTSLM}{OpenTSLM}
\newacronym{SP}{SP}{Soft Prompting}
\newacronym{MLP}{MLP}{multi-layer perceptron}
\newacronym{TSQA}{TSQA}{Time-Series Question Answering}
\newacronym{TSLM}{TSLM}{Time-Series Language Model}
\newacronym{OpenTSLM}{OpenTSLM}{}
\newacronym{QA}{QA}{Question Answering}
\newacronym{REM}{REM}{rapid eye movement}
\newacronym{VRAM}{VRAM}{video random access memory}
\newacronym{LoRA}{LoRA}{low rank adaptation}
\newacronym{TSFM}{TSFM}{time-series foundation model}

\glsunset{EmbedHealth}
\glsunset{OpenTSLM}
\glsdisablehyper

\newcommand{\capitalizeautoref}[1]{\expandafter\MakeUppercase\expandafter{#1}}

\begin{document}

\twocolumn[
  \icmltitle{OpenTSLM: Time-Series Language Models for Reasoning over Multivariate Medical Text- and Time-Series Data}



  \icmlsetsymbol{equal}{*}

 \begin{icmlauthorlist}
  \icmlauthor{Patrick Langer}{stanfordbio,ethzcdhi,ethzagentic}
  \icmlauthor{Thomas Kaar}{stanfordbio,ethzagentic}
  \icmlauthor{Max Rosenblattl}{stanfordbio,ethzagentic}
  \icmlauthor{Maxwell A. Xu}{uiuc,google}
  \icmlauthor{Winnie Chow}{stanford}
  \icmlauthor{Martin Maritsch}{amazon}
  \icmlauthor{Robert Jakob}{ethzagentic}
  \icmlauthor{Ning Wang}{ethzagentic}
  \icmlauthor{Juncheng Liu}{microsoft,nus}
  \icmlauthor{Aradhana Verma}{stanfordcardio}
  \icmlauthor{Brian Han}{pedcardio}
  \icmlauthor{Daniel Seung Kim}{uwcardio}
  \icmlauthor{Henry Chubb}{pedcardio}
  \icmlauthor{Scott Ceresnak}{pedcardio}
  \icmlauthor{Aydin Zahedivash}{stanfordbio}
  \icmlauthor{Alexander Tarlochan Singh Sandhu}{stanfordcardio}
  \icmlauthor{Fatima Rodriguez}{stanfordcardio}
  \icmlauthor{Daniel McDuff}{google,uw}
  \icmlauthor{Elgar Fleisch}{ethzcdhi,ethzagentic,unisg}
  \icmlauthor{Oliver Aalami}{stanfordbio}
  \icmlauthor{Filipe Barata}{ethzcdhi,equal}
  \icmlauthor{Paul Schmiedmayer}{stanfordbio,equal}
\end{icmlauthorlist}
\icmlaffiliation{stanfordbio}{Stanford Mussallem Center for Biodesign, Stanford University, USA}
\icmlaffiliation{ethzcdhi}{Centre for Digital Health Interventions, ETH Zurich, Switzerland}
\icmlaffiliation{ethzagentic}{Agentic Systems Lab, ETH Zurich, Switzerland}
\icmlaffiliation{nus}{National University of Singapore, Singapore}
\icmlaffiliation{microsoft}{Microsoft}
\icmlaffiliation{uiuc}{University of Illinois Urbana-Champaign, USA}
\icmlaffiliation{google}{Google Research, USA}
\icmlaffiliation{stanford}{Stanford University, USA}
\icmlaffiliation{amazon}{Amazon, USA}
\icmlaffiliation{unisg}{Centre for Digital Health Interventions, University of St.\ Gallen, Switzerland}
\icmlaffiliation{stanfordcardio}{Division of Cardiovascular Medicine, Stanford University, USA}
\icmlaffiliation{uwcardio}{Division of Cardiology, University of Washington, USA}
\icmlaffiliation{pedcardio}{Pediatric Cardiology, Stanford University, USA}
\icmlaffiliation{uw}{University of Washington, USA}
\icmlcorrespondingauthor{Patrick Langer}{planger@stanford.edu}

  \icmlkeywords{Machine Learning, ICML}

  \vskip 0.3in
]



\printAffiliationsAndNotice{\icmlEqualContribution}

\begin{abstract}
\Glspl{LLM} have shown strong capability in interpreting multimodal data but remain limited in their ability to natively handle time-series data. Addressing this limitation could enable the translation of longitudinal and wearable sensing data into actionable insights and patient-facing digital health applications.
We propose OpenTSLM, a family of \glspl{TSLM} that integrate time-series as a native modality into pretrained \glspl{LLM}, enabling natural-language prompting and reasoning over multiple time-series. We implement two OpenTSLM variants based on soft prompting~(OpenTSLM-SoftPrompt) and cross-attention~(OpenTSLM-Flamingo). 
To conduct comprehensive experiments on reasoning over medical text and time-series, we introduce three \gls{CoT} datasets: HAR-CoT (human activity recognition), Sleep-CoT (sleep staging), and ECG-QA-CoT (electrocardiogram question answering). 
Across tasks, OpenTSLM models consistently outperform baselines.
OpenTSLMs with time-series encoders trained from scratch achieve 69.88\% in sleep staging and 65.44\% in HAR, while OpenTSLM combined with \glspl{TSFM} achieve 68.33\% and 67.64\%, compared to 9.05\% and 60.44\% for fine-tuned text-only baselines.
Additionally, we conduct expert evaluations with cardiologists, which show that OpenTSLMs exhibit strong reasoning capabilities and temporal understanding on raw sensor data for ECG-QA. 
We further show that OpenTSLM-Flamingo models scale better in memory as the number and length of time-series increase. 
To facilitate further research, we release all code, datasets, and models as open-source resources.
\end{abstract}

\section{Introduction}
\label{sec:introduction}
\glsresetall

Medicine is inherently temporal: assessment, diagnosis, and treatment depend on how signs, symptoms, and biomarkers evolve over time~\cite{identifying_temporal_patterns_population_study, health_and_illness_over_time, disease_trajectories_from_healthcare}.
Clinical decision-making relies on temporal patterns—tracking vital signs, medication responses, laboratory values, and disease progression markers to guide diagnosis, prognosis, and therapeutic interventions. 
As time-series data from electronic health records and continuous monitoring proliferate~\cite{the_promise_of_digital_health, lancet_challenges_of_digital_health, digital_healthcare_technologies}, human-legible representations become essential for interpreting and managing this information~\cite{clinical_reasoning_temporal_domain, jmir_missing_data_in_ehrs, temporal_reasoning_system_discharge_summaries}.  
Clinical summaries must translate complex temporal patterns, e.g., biomarker trajectories, into interpretable assessments that support evidence-based decision-making and care coordination.

Recent advances in multimodal \glspl{LLM} allow users to interpret complex data through natural language, synthesizing information across text, images, audio, and video~\cite{multimodal_llms, multimodal_llms_healthcare}.
However, reasoning over time-series data remains a critical blind spot among currently supported modalities.
Prior work has attempted to integrate time-series as plain text tokens~\cite{llmtime, yubin_kim_2024_health_llm, liu_2023_large_language_models_are_few_shot_health_learners}; however results have been limited~\cite{llms_time_series_struggle}.
Other approaches reprogram \glspl{LLM} to act as feature extractors for classification heads, which then output a fixed set of classes or values, thereby losing text-generation capabilities~\cite{sensor_llm,a_time_series_is_64_words,time2lang, medualtime}.
More recently, soft prompting has been explored, concatenating learnable time-series tokens with text tokens to preserve generation~\cite{apple_health_ai_time_series_reasoning}.
Yet, longer series may require more tokens, increasing context length~\cite{2025_goetz_byte_encoding_ts_tokenization,a_time_series_is_64_words} and compute due to the quadratic cost of self-attention~\cite{a_time_series_is_64_words, attention_is_all_you_need}.

To overcome prior limitations, we propose \glspl{TSLM}, which integrate time-series as a native modality in \glspl{LLM}.
\glspl{TSLM} provide a natural interface to complex medical data, enabling clinicians and patients to query, interpret, and reason about longitudinal health information directly through natural language.
We introduce \gls{OpenTSLM}, a family of \glspl{TSLM} built by extending pretrained \glspl{LLM} with time-series inputs. 
A central design question in building \glspl{TSLM} is how to represent time-series signals. 
Prior work has primarily used soft prompting, encoding time-series as learned token embeddings concatenated with text tokens. 
While lightweight, this captures temporal dependencies only implicitly, as additional tokens in the context, and may scale poorly to longer or multiple sequences. 
We hypothesize that explicit multimodal fusion via cross-attention may be more effective for modeling temporal structure. 
To compare both approaches, we explore two variants for \gls{OpenTSLM}.
The first, \textbf{\gls{OpenTSLM}-SoftPrompt} leverages soft prompting to concatenate text tokens with time-series tokens predicted from a learnable encoder, so the model processes both as a single sequence without distinguishing between them. 
The second, \textbf{\gls{OpenTSLM}-Flamingo}, by contrast, models time-series explicitly as a separate modality, using a cross-attention mechanism inspired by Flamingo~\cite{flamingo} to fuse time-series and text.
We construct \gls{OpenTSLM}-SoftPrompt and \gls{OpenTSLM}-Flamingo using Llama~\cite{llama} and Gemma~\cite{gemma} backbones, and pair them with time-series encoders trained from scratch as well as with \glspl{TSFM} such as Chronos-2~\cite{ansari2025chronos}. 
We benchmark these models against each other and against baselines including \glspl{LLM} with tokenized time-series inputs~\cite{llmtime}, fine-tuned tokenized time-series models, and vision-based approaches. 
Unlike prior classification-based approaches, our models are trained in text-based reasoning tasks, generating \gls{CoT} rationales before producing predictions.
For training and evaluation, we introduce three new datasets: \textbf{HAR-CoT}, \textbf{Sleep-CoT}, and \textbf{ECG-QA-CoT}. 
To foster reproducibility and further research on \glspl{TSLM}, we release OpenTSLM as an open-source framework, including models and datasets.
\footnote{Hidden during double-blind review.}
\section{Related Work}
\label{sec:related_work}

Creating \glspl{TSLM} remains an open research challenge.
The main barrier is the modality gap between continuous signals and discrete text representations~\cite{apple_health_ai_time_series_reasoning,time2lang, sensorlm}.  
Prior work has proposed three main strategies to bridge this gap, as summarized by \citet{time_series_llms_survey}:
tokenizing time-series as text (\autoref{sec:related_work:tokenization}),
applying soft prompting (\autoref{sec:related_work:softprompting}),
and using cross-attention mechanisms (\autoref{sec:related_work:crossattention}).
\autoref{tab:related_work_overview} provides an overview of relevant methods.

\begin{table}[h!t]
\centering
\scriptsize
\caption{Methods combining time-series data with LLMs.}
\begin{tabularx}{1.0\linewidth}{l XXXXXX}
\textbf{Name} & \textbf{Method} & \textbf{Task} & \textbf{Text Gen.} & \textbf{Multi-Sensor} & \textbf{Raw Data} & \textbf{SFT}\\
\midrule
FSHL\cite{liu_2023_large_language_models_are_few_shot_health_learners} & \TokenCircle & \ClassificationCircle & \greencheck & \greencheck & \greencheck &  \\
\cite{llmtime} & \TokenCircle & \ForecastingCircle & \greencheck &  & \greencheck &  \\
HealthLLM~\cite{yubin_kim_2024_health_llm} & \TokenCircle & \TextReasoningCircle & \greencheck & \greencheck & \greencheck & \greencheck \\
\midrule\citet{apple_health_ai_time_series_reasoning} & \SPCircle & \TextReasoningCircle & \greencheck & \greencheck & \greencheck & \greencheck \\
ChatTS~\cite{xie2025chatts} & \SPCircle & \TextReasoningCircle & \greencheck & \greencheck & \greencheck & \greencheck \\
ITFormer~\cite{wang2025itformerbridgingtimeseries} & \SPCircle & \TextReasoningCircle & \greencheck & \greencheck & \greencheck & \greencheck \\
InstrucTime~\cite{cheng2025instructime} & \SPCircle & \TextReasoningCircle & \greencheck & \greencheck & \greencheck & \greencheck \\
MedTsLLM~\cite{chan2024medtsllm} & \SPCircle & \ClassificationCircle & & \greencheck & \greencheck & \\
MedualTime~\cite{medualtime} & \SPCircle & \ClassificationCircle &  &  & \greencheck & \greencheck \\
SensorLLM~\cite{sensor_llm} & \SPCircle & \ClassificationCircle &  & \greencheck & \greencheck & \greencheck \\
Time2Lang~\cite{time2lang} & \SPCircle & \ClassificationCircle &  &  & \greencheck &  \\
\textbf{OpenTSLM-SP (ours)} & \SPCircle  & \TextReasoningCircle & \greencheck & \greencheck & \greencheck & \greencheck \\
\midrule
SensorLM~\cite{sensorlm} & \CrossCircle & \ClassificationCircle & \greencheck & \greencheck &  &  \\
\textbf{OpenTSLM-Flamingo (ours)} & \CrossCircle & \TextReasoningCircle & \greencheck & \greencheck & \greencheck & \greencheck \\
\end{tabularx}
\label{tab:related_work_overview}\\
\medskip
\noindent
\TokenCircle=Tokenization,\;
\SPCircle=Soft Prompting,\;
\CrossCircle=Cross-Attention,\;\quad\quad
\AdapterCircle=Adapter;
\ClassificationCircle=Classification,\;
\ForecastingCircle=Forecasting,\;
\TextReasoningCircle=Text Reasoning
\vspace{-2em}
\end{table}

\subsection{Tokenization of Time-Series as Text Inputs}
\label{sec:related_work:tokenization}

\citet{llmtime} has demonstrated that \glspl{LLM} can perform time-series forecasting by encoding values as text tokens and predicting future values without domain-specific tuning.
\citet{liu_2023_large_language_models_are_few_shot_health_learners} tokenize data from wearables and smartphones to enable \glspl{LLM} to infer clinical and wellness information through few-shot prompting.  
Similarly, \citet{yubin_kim_2024_health_llm} propose HealthLLM, a framework for health prediction using physiological signals (e.g., heart rate, sleep) combined with user context and medical knowledge embedded in prompts.

\subsection{Combining Text and Time-Series Token Embeddings (Soft Prompting)}
\label{sec:related_work:softprompting}

An alternative to manual tokenization is to encode time-series into embeddings that capture time-series information, using a time-series encoder as presented by~\citet{a_time_series_is_64_words}.
These embeddings can be input into a transformer directly or concatenated with text embeddings (soft prompting)~\cite{ apple_health_ai_time_series_reasoning,        a_time_series_is_64_words,time2lang, medualtime, xie2025chatts, wang2025itformerbridgingtimeseries, cheng2025instructime,chan2024medtsllm}.
\citet{time2lang} use this approach and train an encoder to produce soft prompts from time-series, which are then processed by a frozen \gls{LLM} for classification via a projection head; however, this disables free-form text generation.
\citet{medualtime} and \citet{chan2024medtsllm} similarly combine time-series and text-token embeddings, using a classification head~\cite{medualtime} and a task solver~\cite{chan2024medtsllm} for prediction.
\citet{wang2025itformerbridgingtimeseries} introduced ITFormer, a novel framework that combines any time-series encoder with any frozen LLM to support time-series question answering, also by combining text- and derived time-series tokens.
\citet{cheng2025instructime} introduce a framework that first aligns time-series and natural language in a general stage, and later finetunes for a specific domain to perform classification.
\citet{sensor_llm} integrate sensor and text embeddings in two stages:
First, they generate a caption-like summary of the time-series for free-form output;
Second, they classify the data via a projection head, therefore restricting free from output.
\citet{apple_health_ai_time_series_reasoning} and \citet{xie2025chatts} interleave time-series tokens with text tokens in the \gls{LLM} input, enabling free-form text reasoning.

\subsection{Cross-Attention for Time-Series Data}
\label{sec:related_work:crossattention}

Few studies use cross-attention to integrate time-series into \glspl{LLM}.  
\citet{sensorlm} apply cross-attention between a time-series encoder and a text encoder, aligned with contrastive loss, to extract statistical summaries (e.g., mean, max) from a single sensor.  
They train a new sensor encoder, text encoder, and multimodal text decoder, rather than adapting a pretrained \gls{LLM}~\cite{sensorlm}.

\subsection{Pretrained Time-Series Foundation Models}
Recent work on time-series foundation models (TSFMs) shows that large-scale pretraining enables strong zero-shot forecasting across domains. 
TimesFM~\citep{das2024decoder} proposes a practical decoder-only Transformer with patching, pretrained on large-scale real-world time-series data, and achieves competitive performance. 
Moirai2~\citep{liu2025moirai} adopts a decoder-only design with quantile-based objectives and multi-token prediction to improve probabilistic accuracy and inference efficiency. 
Chronos-2~\citep{ansari2025chronos} extends pretrained forecasting to multivariate and covariate-aware settings via group attention for sharing information across related series. 
While these TSFMs achieve strong results on established forecasting benchmarks (e.g., GIFT-Eval~\citep{aksu2024gift}), they are primarily designed for forecasting and typically lack explicit reasoning capabilities and text generation interfaces.

\section{Methods}
\label{sec:methods}

\begin{figure}[h!t]
  \centering
  \includegraphics[width=0.95\linewidth]{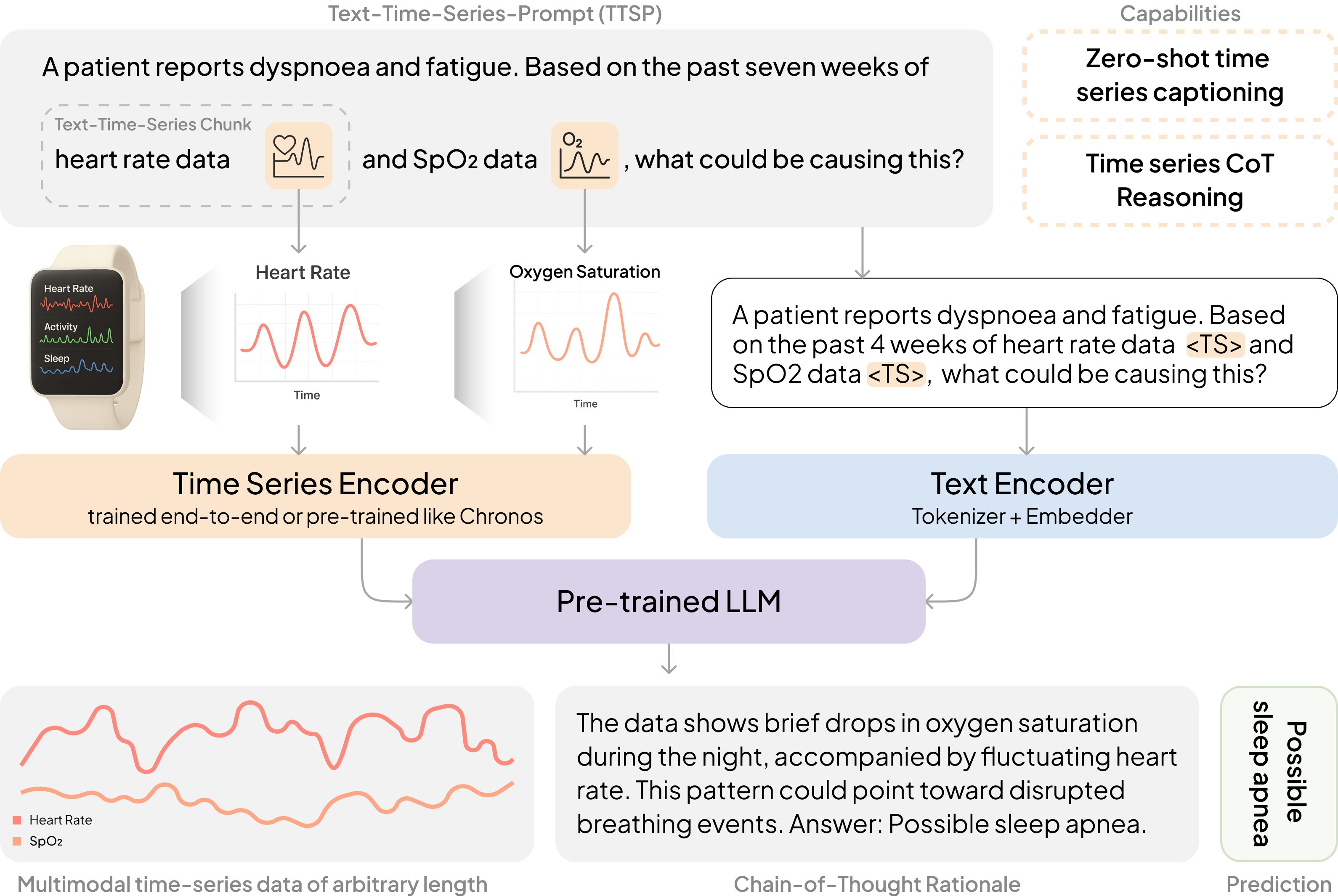}
  \caption{Overview of Text–Time-Series LLMs with support for multiple time-series inputs.}
  \label{fig:approach_overview}
\end{figure}

We present two architectures for \glspl{TSLM}, \gls{OpenTSLM}-\gls{SP}~(\autoref{sec:methods:OpenTSLM_sp}) and \gls{OpenTSLM}-Flamingo~(\autoref{sec:methods:OpenTSLM_flamingo}).
To support multiple time-series inputs, we design a prompt format that interleaves sensor data with accompanying textual descriptions (e.g., “Data from Sensor X over Y days:” followed by the data representation).
\autoref{fig:approach_overview} illustrates our approach.

\subsection{Time-Series encoder}
\label{sec:methods:time_series_encoder}
We use a time-series encoder to extract features that characterize the input time-series. We support either training an encoder from scratch, leveraging the PatchTST architecture~\cite{a_time_series_is_64_words}, or using pretrained \glspl{TSFM} such as Chronos-2~\cite{ansari2025chronos}.
When trained from scratch, the encoder is trained with our models, whereas for pretrained encoders the parameters are frozen and only a projection layer is trained alongside our models (see~\autoref{sec:methods:training}).

In the case of PatchTST, the encoder consists of a \texttt{PatchEncoder}, followed by either a \texttt{TransformerEncoder} for \gls{OpenTSLM}-\gls{SP} or a \texttt{PerceiverResampler} for \gls{OpenTSLM}-Flamingo, inspired by~\citet{flamingo, open_flamingo}.
We divide an input time-series $x \in \mathbb{R}^{L}$ into non-overlapping patches of size $p$, yielding $N = L/p$ patches. Each patch is then transformed into a patch embedding vector using a 1D convolution and combined with a positional encoding~\cite{a_time_series_is_64_words}:
\begin{equation}
\mathbf{E}i = \text{Conv1D}(x{i \cdot p:(i+1) \cdot p}) \in \mathbb{R}^{d_{\text{enc}}} + \mathbf{P}i ,
\end{equation}
where the convolution has kernel size and stride equal to $p$, mapping each patch to a $d{\text{enc}}$-dimensional embedding, and $\mathbf{P}_i$ denotes a learnable positional encoding. 
The resulting sequence of position-augmented embeddings is processed by the corresponding encoder (cf. Sections~\ref{sec:methods:OpenTSLM_sp} and~\ref{sec:methods:OpenTSLM_flamingo}).

\paragraph{Preserving scale and temporal information}

The \texttt{PatchEncoder} expects inputs normalized to $x \in [-1, 1]$. Since raw time-series differ in scale and resolution across modalities depending on the sensor. 
Consistent with prior work~\cite{apple_health_ai_time_series_reasoning, xie2025chatts} we preserve scale and temporal context by adding the original mean, standard deviation, and time scale to the textual description. 
For example: \textit{This is heart-rate data over \textbf{24 hours sampled at 50\,Hz} with \textbf{mean=61} and \textbf{std=12}.}

\begin{figure*}[h]
  \centering
  \includegraphics[width=0.7\linewidth]{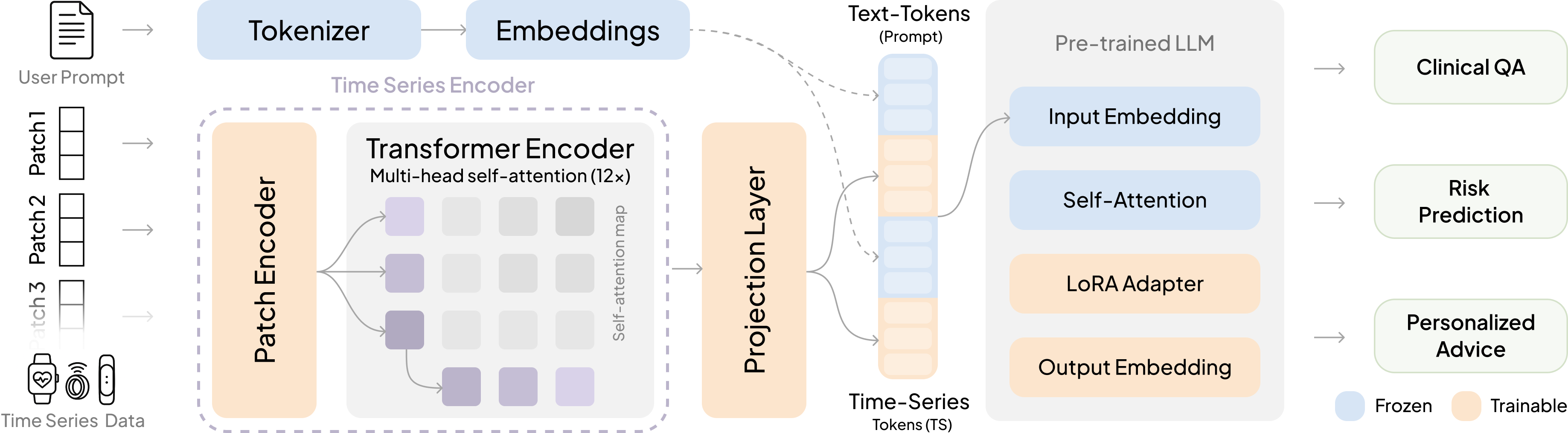}
  \caption{Architecture of \gls{OpenTSLM}-SoftPrompt}
  \label{fig:soft_prompting_architecture}
\end{figure*}

\subsection{Soft prompting architecture~(\gls{OpenTSLM}-\gls{SP})}
\label{sec:methods:OpenTSLM_sp}

\gls{OpenTSLM}-\gls{SP} has three components (\autoref{fig:soft_prompting_architecture}):  
(1) a time-series encoder that transforms raw data into patch embeddings,  
(2) a projection layer mapping embeddings to the \gls{LLM} hidden space,  
(3) a pretrained \gls{LLM}, fine-tuned using LoRA adapters~\cite{lora}.

\paragraph{Projecting Time-Series Tokens to Text Tokens}
We apply the patch embeddings to a transformer encoder and subsequently project the resulting tokens with an \gls{MLP} to align them with the embedding space of dimension $d_{\text{llm}}$, corresponding to the hidden size of the \gls{LLM}, following \citet{a_time_series_is_64_words} and \citet{apple_health_ai_time_series_reasoning}.
\begin{equation}
\mathbf{Z} = \text{MLP}(\mathbf{TransformerEncoder}(E_{1:N})) \in \mathbb{R}^{N \times d_{\text{llm}}}
\end{equation}
where $\mathbf{Z} \in \mathbb{R}^{N \times d_{\text{llm}}}$ denotes the projected time-series tokens in the \gls{LLM} embedding space.

\paragraph{Text-Time-Series integration via Soft Prompting}

We interleave any number of text and time-series tokens through a soft prompting mechanism.
A typical prompt consists of (1) an initial text segment (“pre-prompt”), (2) a sequence of interleaved time-series tokens and textual descriptions, and (3) a final text segment (“post-prompt”), often a question. 
Formally, the model input is:
\begin{equation}
\mathbf{X}_{\text{input}} = [\mathbf{T}_{\text{pre}}, \mathbf{Z}_1, \mathbf{T}_{\text{desc}_1}, \mathbf{Z}_2, \mathbf{T}_{\text{desc}_2}, \ldots, \mathbf{Z}_K, \mathbf{T}_{\text{desc}_K},
\mathbf{T}_{\text{post}}]
\end{equation}
where $\mathbf{T}_{\text{pre}}$, $\mathbf{T}_{\text{desc}_i}$, and $\mathbf{T}_{\text{post}}$ are token embeddings of text segments, and each $\mathbf{Z}_i$ is a projected time-series embedding aligned with the \gls{LLM} hidden space.  
We refer to each $(\mathbf{Z}_i, \mathbf{T}_{\text{desc}_i})$ as a text–time-series chunk. This approach implicitly integrates time-series through learned tokens.

\subsection{Cross-attention architecture~(\gls{OpenTSLM}-Flamingo)}
\label{sec:methods:OpenTSLM_flamingo}

\begin{figure*}[h]
  \centering
  \includegraphics[width=0.7\textwidth]{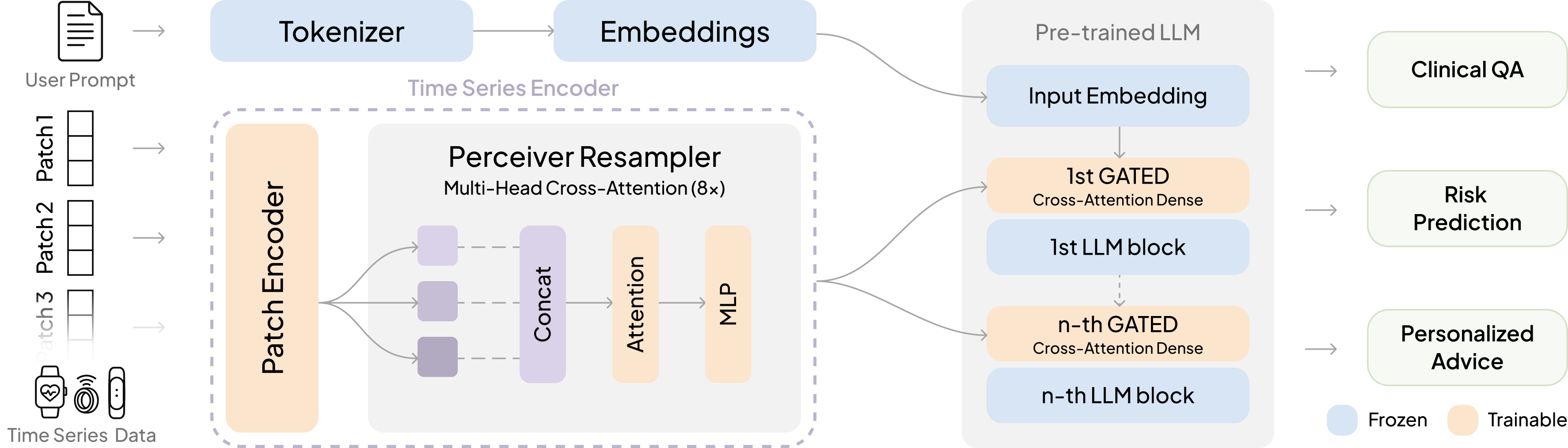}
  \caption{Architecture of \gls{OpenTSLM}-Flamingo}
  \label{fig:flamingo_architecture}
\end{figure*}

\gls{OpenTSLM}-Flamingo is inspired by the Flamingo model for vision–language tasks~\cite{flamingo, open_flamingo}.  
Following OpenFlamingo~\cite{open_flamingo}, we extend pretrained \glspl{LLM} with cross-attention layers to support time-series reasoning.  
\paragraph{Architecture Overview}
We replace the vision encoder of Flamingo with a time-series encoder and adapt the cross-attention mechanism for temporal data.  
The model consists of:  
(1) a time-series patch encoder,  
(2) a Perceiver Resampler,  
(3) gated cross-attention layers integrated into the LLM, and  
(4) the frozen language model backbone. \autoref{fig:flamingo_architecture} visualizes the architecture.

\paragraph{PerceiverResampler}
We use a \text{PerceiverResampler} inspired by Flamingo~\cite{open_flamingo} as Encoder for the time-series patches, yielding a fixed-size latent representation:
\begin{equation}
    \mathbf{Z}_{\text{latent}} = \text{PerceiverResampler}(\mathbf{E}_{1:N}) \in \mathbb{R}^{N_{\text{latent}} \times d_{\text{time}}},
\end{equation}
Here, $d_{\text{time}}$ is the dimensionality of the time-series features by the perceiver, in our case $(N, 1)$, encoding one time-series with one channel at a time. 

\paragraph{Text-Time-Series Gated Cross-Attention}

To integrate $\mathbf{Z}_{\text{latent}}$ into the \gls{LLM}, we add gated cross-attention layers every $N$~(hyperparameter) transformer blocks which compute:
\begin{equation}
\mathbf{Q}_{\text{text}} = \mathbf{x}\mathbf{W}_Q, \quad \mathbf{K}_{\text{ts}} = \mathbf{Z}_{\text{latent}}\mathbf{W}_K, \quad \mathbf{V}_{\text{ts}} = \mathbf{Z}_{\text{latent}}\mathbf{W}_V
\end{equation}
\begin{equation}
\text{Gated-XAttn}(\mathbf{x}, \mathbf{Z}_{\text{latent}}) = x + \gamma\cdot  \text{softmax}\left(\frac{\mathbf{Q}_{\text{text}}\mathbf{K}_{\text{ts}}^T}{\sqrt{d_k}}\right)\mathbf{V}_{\text{ts}}.
\end{equation}
where $\gamma_{\text{attn}}$ is a learnable parameter controlling the influence of the time-series, $\mathbf{x} \in \mathbb{R}^{T \times d_{\text{model}}}$, the LLM input,  
$\mathbf{W}_Q, \mathbf{W}_K, \mathbf{W}_V \in \mathbb{R}^{d_{\text{model}} \times d_k}$ learned projection matrices,  and $d_k$ the key dimension.

\paragraph{Conditioning Text-tokens on Time-Series via Special Tokens}

The \gls{LLM} processes tokens autoregressively, attending to previous inputs.  
Following OpenFlamingo~\cite{open_flamingo}, we introduce special tokens $\langle\text{TS}\rangle$ and $\langle\text{endofchunk}\rangle$ to indicate when time-series modalities should be incorporated. 
Upon encountering $\langle\text{TS}\rangle$, the model conditions on the corresponding latent representation $\mathbf{Z}_{\text{latent}}$ via gated cross-attention. A typical input prompt is
\begin{equation}
\begin{aligned}
\mathbf{X}_{\text{input}} ={}&
[\text{pre\_prompt}, \langle\text{TS}\rangle, \text{ts\_desc}_1,
\langle\text{endofchunk}\rangle, \\
& \langle\text{TS}\rangle, \text{ts\_desc}_2,
\langle\text{endofchunk}\rangle, \text{post\_prompt}]
\end{aligned}
\end{equation}

where $\langle\text{TS}\rangle$ triggers multimodal conditioning and $\langle\text{endofchunk}\rangle$ signals the end of text describing a time-series. 
This setup enables interleaving multiple text and time-series segments~\cite{open_flamingo}. The embeddings of the special tokens are learned.

\section{Experiments}
\label{sec:experiments}

In the following, we outline our training methodology and report results on multiple-choice \gls{TSQA} and time-series reasoning datasets. 
We compare \gls{OpenTSLM}-SoftPrompt and \gls{OpenTSLM}-Flamingo against each other and baselines in terms of performance, and report \gls{VRAM} requirements for training \gls{OpenTSLM}. 
We present sample model outputs across datasets and an evaluation for \gls{ECG} rationales by expert cardiologists.

\subsection{Multi-Stage Curriculum Learning – Teaching \glspl{LLM} Time-Series}
\label{sec:methods:training}

Following~\cite{apple_health_ai_time_series_reasoning}, we adopt a two-stage curriculum to train \glspl{TSLM}. In stage one (encoder warmup), we use two synthetic time-series datasets to pretrain the encoder: 
\begin{itemize}[left=7pt, itemsep=0pt, topsep=0pt, nosep]
    \item \textbf{\gls{TSQA}~\cite{tsqa}}  
Multiple-choice time-series question answering on synthetic data for learning simple temporal patterns (e.g., ascending/descending trends).
    \item \textbf{Time-Series Captioning (M4-Captions)}
We generate pseudo-labeled captions using ChatGPT, prompted with plots of time-series of the M4 dataset~\cite{m4_dataset}~(see \autoref{sec:appendix:m4}).
\end{itemize}

\noindent In stage two, we introduce three new \gls{CoT} time-series datasets covering \gls{HAR}, sleep staging, and \gls{ECG} \gls{QA}. 
We generated these using GPT-4o by providing a plot and ground-truth answer for each sample, then asking the model to produce rationales leading to the correct response. 
Further details are provided in \autoref{sec:appendix:datasets}.
\begin{itemize}[left=7pt, itemsep=0pt, topsep=0pt]
\item \textbf{\gls{HAR}-\gls{CoT}}  
three-axis accelerometer data combined from DaLiAc~\cite{daliac_dataset}, DOMINO~\cite{domino_dataset}, HHAR~\cite{hhar_dataset}, PAMAP2~\cite{pamap}, RealWorld~\cite{realworld_har_dataset}, and datasets from~\cite{shoaib2013_uthar,shoaib2014_utsensors,shoaib2016_utcomplex}. 
Sampled at 50 Hz, split into 2.56s windows, 8 activities: sitting, standing, lying, walking, running, biking, walking upstairs, walking downstairs. See~\autoref{sec:appendix:har_generation} for detailed description.
\item \textbf{Sleep-\gls{CoT}}  
Based on SleepEDF~\cite{sleepedf,physionet}, using 30s \gls{EEG} segments for sleep staging. 
Following prior work~\cite{apple_health_ai_time_series_reasoning,pouliou2025_sleep_stage_non_rem_merge}, Non-\gls{REM} stages 3 and 4 are merged, yielding five classes: Wake, \gls{REM}, Non-\gls{REM}1, Non-\gls{REM}2, Non-\gls{REM}3. See~\autoref{sec:appendix:sleep_generation} for details.

\item \textbf{\gls{ECG}-QA-\gls{CoT}}  
Based on ECG-QA~\cite{ecg_qa}, which provides 12-lead 10s ECGs and clinical context, we excluded comparison questions, retaining 42/70 templates. 
This yielded 3,138 unique questions across 240k samples (see~\autoref{sec:appendix:ecg_qa_generation}).

\end{itemize}
All datasets are split into \textbf{80/10/10 train/validation/test} sets. \autoref{tab:datasets} in ~\autoref{sec:appendix:training_details} summarizes number of samples in the datasets, number of time-series and lengths.

\paragraph{Training objective}

In all stages, we frame the task as an autoregressive language modeling problem. 
During training and evaluation, the model is prompted to generate outputs in a structured format, consisting of a free-form rationale followed by the final prediction: \texttt{<reasoning> Answer: <final answer>}. Formally, the loss is defined by \autoref{eq:loss}, where $\mathbf{Z}_{\text{ts}}$ are the
\noindent
\begin{equation}
    \mathcal{L}_{\text{LM}} = - \sum_{t=1}^{T} \log P(y_t \mid y_{<t}, \mathbf{x}_{1:t}, \mathbf{Z}_{\text{ts}}; \Theta)
    \label{eq:loss}
\end{equation}\noindent
time-series features, and $\Theta$ the learnable weights, i.e., the TimeSeriesEncoder, MLP, and LoRA in \gls{OpenTSLM}-SoftPrompt, and TimeSeriesEncoder and cross-attention in \gls{OpenTSLM}-Flamingo.

\subsection{Baselines}
\label{sec:experiments:baseline}

We compare OpenTSLM against three baselines using the same open-weight \glspl{LLM}, i.e., Llama-3.2(1B, 3B) and Gemma3 (270M, 1B-PT), and additionally GPT-4o~(gpt-4o-2024-08-06).

\begin{enumerate}[left=2pt,, itemsep=0pt, topsep=0pt, nosep]
\item \textbf{Tokenized time-series}: 
Using the open-source code provided by~\citet{llmtime}, we tokenize time-series into text inputs and report zero-shot performance on the test set.
\item \textbf{Tokenized finetuned}: Same as 1.~(excluding GPT-4o), but finetuned with LoRA~\cite{lora} on the training set. We choose best model by validation loss, and report performance on test set.
\item \textbf{Image~(Plot)}: We convert time-series into as input for GPT-4o and Gemma-4b-pt~(since the smaller Gemma 3 variants do not support image input).
\item \textbf{Random baseline}: For comparison, we report the expected performance of a predictor that selects labels uniformly at random, adjusted to each dataset’s label distribution.
\end{enumerate}

\subsection{Quantitative results on time-series classification}
\label{sec:results:classification}
We present performance on the test splits of TSQA, \gls{HAR}-\gls{CoT}, Sleep-\gls{CoT}, and ECG-QA-CoT and report macro-F1 score and accuracy in \autoref{tab:classification_results}.
\begin{table}[h!t]
    \scriptsize
    \centering
    \caption{Performance comparison on time-series question answering~(TSQA) and time-series reasoning (HAR-CoT, Sleep-CoT, ECG-QA-CoT) tasks between OpenTSLM models and baselines.}
    \vspace{-0.5em}

    \setlength{\tabcolsep}{3pt}
    \renewcommand{\arraystretch}{1.05}

  \begin{tabularx}{\linewidth}{X p{1.6cm} X *{7}{X}}
  \toprule
    \textbf{ } & \textbf{Model} 
        & \multicolumn{2}{c}{\textbf{TSQA}} 
        & \multicolumn{2}{c}{\textbf{HAR-\gls{CoT}}} 
        & \multicolumn{2}{c}{\textbf{Sleep-\gls{CoT}}} 
        & \multicolumn{2}{c}{\textbf{ECG-QA-CoT}} \\
    & & F1 & Acc & F1 & Acc & F1 & Acc & F1 & Acc \\
    \midrule
    \multicolumn{2}{l}{Random Baseline} & 33.33 & 33.33 & 11.49 & 12.50 & 17.48 & 20.00 & 16.47 & 20.18 \\
    \midrule
    \multirow{5}{*}{\rotatebox{90}{\shortstack{Tokenized\\Time-Series}}} 
        & Llama3.2-1B      & 16.01 & 31.04 & 0.00\textsuperscript{*1}  & 0.00 & 2.14 & 0.65 & 0.00 & 0.00 \\
        & Llama3.2-3B      & 16.24 &  32.06  & 0.00 & 0.00 & 5.66 & 12.15 & 0.00 & 0.00 \\
        & Gemma3-270M      & 10.52   & 9.58 & 0.00  & 0.00 & 0.00  & 0.00 & 0.00 & 0.00 \\
        & Gemma3-1B-pt        & 11.76   & 12.92 & 0.00  & 0.00 & 0.00  & 0.00 & 0.00 & 0.00 \\
        & GPT-4o       & 45.32 & 45.29 & 2.95 & 11.74 & 15.47 & 16.02 & 18.19 & 28.76 \\
    \midrule
    \multirow{4}{*}{\rotatebox{90}{\shortstack{Tokenized\\Finetuned}}} 
        & Llama3.2-1B      & 83.74 & 81.40 & 51.28 & 62.71 & 9.05 & 24.19 & OOM & OOM\textsuperscript{*2} \\
        & Llama3.2-3B      & 84.54 & 82.06 & 60.44 & 66.87 & 5.86 & 14.30 & OOM & OOM \\
        & Gemma3-270M      & 68.05 & 65.40 & 40.66 & 54.56 & 0.00 & 0.00 & OOM & OOM \\
        & Gemma3-1B-pt        & 82.85 & 83.42 & 52.15 & 63.90 & 0.00 & 0.00 & OOM & OOM \\
    \midrule
    \multirow{2}{*}{\rotatebox{90}{\shortstack{Image\\(Plot)}}} 

        & Gemma3-4B-pt        & 48.77 & 50.60 & 1.72 & 0.89 & 6.75 & 14.95 & 1.90 & 1.03 \\
        & {Gemma3-4B FT}      & 94.94 & 94.85 & 44.01 & 51.02 & 18.56 & 38.28 & 26.17 & 38.10 \\
        & GPT-4o       & 59.24 & 62.10 & 10.83 & 13.90 & 4.82 & 10.75 & 24.95 & 33.30 \\

    \midrule
    \multirow{4}{*}{\rotatebox{90}{\shortstack{OpenTSLM\\SoftPrompt}}} 
        & Llama3.2-1B      & \textbf{97.50} & \textbf{97.54} & \textbf{65.44} & \textbf{71.48} & \textbf{69.88} & \textbf{81.08} & 32.84 & 35.49 \\
        & Llama3.2-3B      & 97.37 & 97.33 & 64.87 & 67.89 & 54.40 & 72.04 & 33.67 & 36.25 \\
        & Gemma3-270M      & 40.32 & 26.79 & 1.43 & 0.55 & 7.96 & 5.91 & 1.29 & 1.11 \\
        & Gemma3-1B-pt        & 87.29 & 89.18 & 40.52 & 45.17 & 30.99 & 36.56 & 27.86 & 34.76 \\ 
    \midrule
    \multirow{4}{*}{\rotatebox{90}{\shortstack{OpenTSLM\\Flamingo}}} 
        & Llama3.2-1B      & 94.08 & 94.00 & 62.93 & 69.27 & 49.33 & 67.31 & 34.62 & 38.14 \\
        & Llama3.2-3B      & 90.14 & 90.10 & 62.77 & 69.03 & 45.45 & 69.14 & \textbf{40.25} & \textbf{46.25} \\
        & Gemma3-270M      & 77.86 & 78.12 & 57.75 & 63.43 & 51.38 & 68.49 & 32.71 & 35.50 \\
        & Gemma3-1B-pt        & 92.56 & 92.46 & \textbf{65.44} & \textbf{71.48} & 43.69 & 60.67 & 35.31 & 37.79 \\
    \bottomrule
    \addlinespace[5pt]
    \end{tabularx}
    \caption*{\footnotesize Note: Gemma models have smaller context than Llama (32k vs. 128k); soft prompt uses up context, performing worse. \textsuperscript{*1}0.00 model failed to produce \texttt{Answer: \{answer\}} template, often repeating input prompt~(see ~\autoref{appendix:baseline_fails}).\textsuperscript{*2}OOM - Out of memory: 12 ECG leads of 10s tokenize to 80k tokens, requiring \textgreater 100GB VRAM. Gemma3-4B FT refers to Gemma-3-4B-pt finetuned.
}
\label{tab:classification_results}
\vspace{-2em}
\end{table}

OpenTSLM models achieve the highest performance across benchmarks, while most tokenized text-only baselines fail to produce valid outputs, not answering in the expected template but merely repeating inputs or starting to count (see \autoref{appendix:baseline_fails}), resulting in 0.00 F1 on HAR for all models except for GPT-4o~(2.95). 
GPT-4o yields only 2.95 F1 with text but improves substantially with plots (e.g., 10.83 on HAR, 59.24 on TSQA). Using plot inputs, Gemma3-4B achieves 48.77 F1 on TSQA, 1.72 on HAR-CoT, 6.75 on Sleep-CoT, and 1.90 on ECG-QA-CoT, but improves significantly after fine-tuning on plots, achieving 94.94,0 44.01, 18.56, and 26.17, respectively.
Llama models achieve 2.14 and 5.65F1 on Sleep, respectively, while Gemma models again achieve 0.00, likely due to their smaller context window (32k vs. 128k). 
By contrast, OpenTSLM–SoftPrompt with Llama3.2-1B attains 97.50 F1 score (97.54 accuracy) on TSQA, with Llama3.2-3B at 97.37 (97.33); Flamingo variants are close (e.g., Llama3.2-1B 94.08 (94.00)), while the strongest tokenized-finetuned baseline reaches 84.54 (82.06) and GPT-4o with image inputs at 59.24 (62.10). 
On HAR-CoT, the strongest results are 65.44F1 (71.48 accuracy) for OpenTSLM–SoftPrompt (Llama3.2-1B) and 65.44 (71.48) for OpenTSLM–Flamingo (Gemma3-1B-pt); the best tokenized-finetuned baseline records 60.44 (66.87). 
On Sleep-CoT, OpenTSLM–SoftPrompt (Llama3.2-1B) achieves 69.88 (81.08), followed by OpenTSLM–SoftPrompt (Llama3.2-3B) at 54.40 (72.04) and Flamingo (Gemma3-270M) at 51.38 (68.49); tokenized-finetuned baselines remain lower (best 9.05 (24.19)). 
On ECG-QA-CoT, OpenTSLM–Flamingo (Llama3.2-3B) leads with 40.25 (46.25).

\subsection{Evaluation of memory use during training}

\begin{figure}[h!t]
  \vspace{-.5em}
  \centering
  \includegraphics[width=0.80\linewidth]{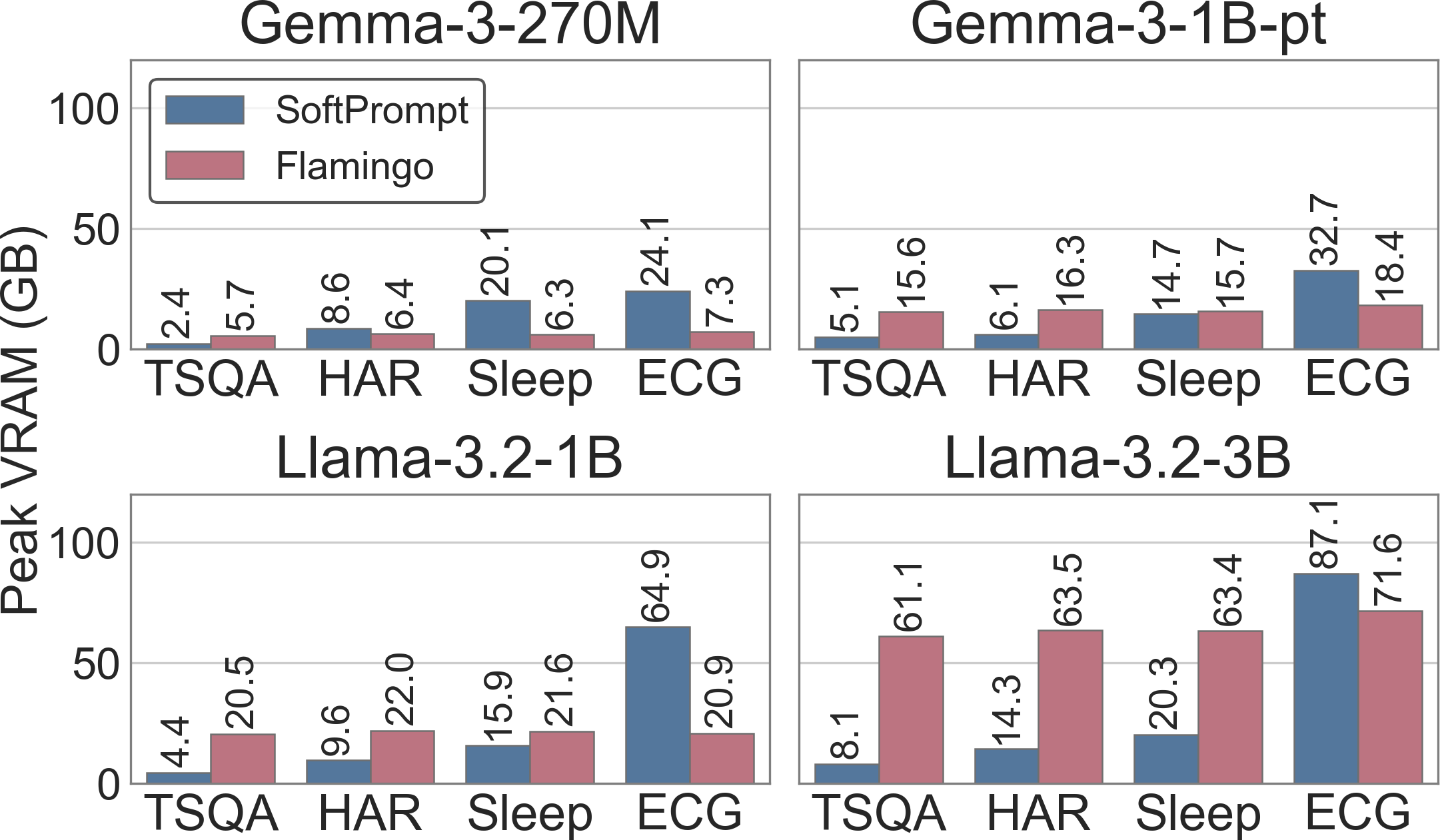}
  \caption{VRAM memory usage in training across datasets.}
  \label{fig:memory_facet} 
  \vspace{-.5em}
\end{figure}

\noindent
We evaluate peak \gls{VRAM} usage during training for both OpenTSLM variants. \autoref{fig:memory_facet} summarizes peak \gls{VRAM} on TSQA, HAR\textendash CoT, SleepEDF\textendash CoT, and ECG\textendash QA\textendash CoT. 
OpenTSLM-Flamingo shows near-constant memory across datasets,
while OpenTSLM-SoftPrompt vary substantially with the dataset,
suggesting that the memory cost of OpenTSLM-SoftPrompt is highly related to the length of time-series and the text tokens. 

To further investigate memory scaling, we train models on a simulated dataset~(see \autoref{sec:appendix:memory_use_simulation})  with various time-series input shapes. 
We report max \gls{VRAM} usage in the appendix (see \autoref{fig:memory_per_length} and  \autoref{tab:vram_usage_simulation}). 
In summary, we observe that VRAM usage of OpenTSLM-Flamingo stays constant as the number of time-series $N$ and the sequence length $L$ increase. In contrast, the VRAM usage of OpenTSLM-SoftPrompt scales with both $N$ and $L$, suggesting poor memory scalability, which can become a practical bottleneck for longer contexts.



\subsection{Qualitative results and expert evaluation of ECG rationales}
\label{sec:methods:qualitative_results}
\begin{figure*}[h!t]
    \centering
    \begin{subfigure}{0.3\linewidth}
        \includegraphics[width=1.0\linewidth]{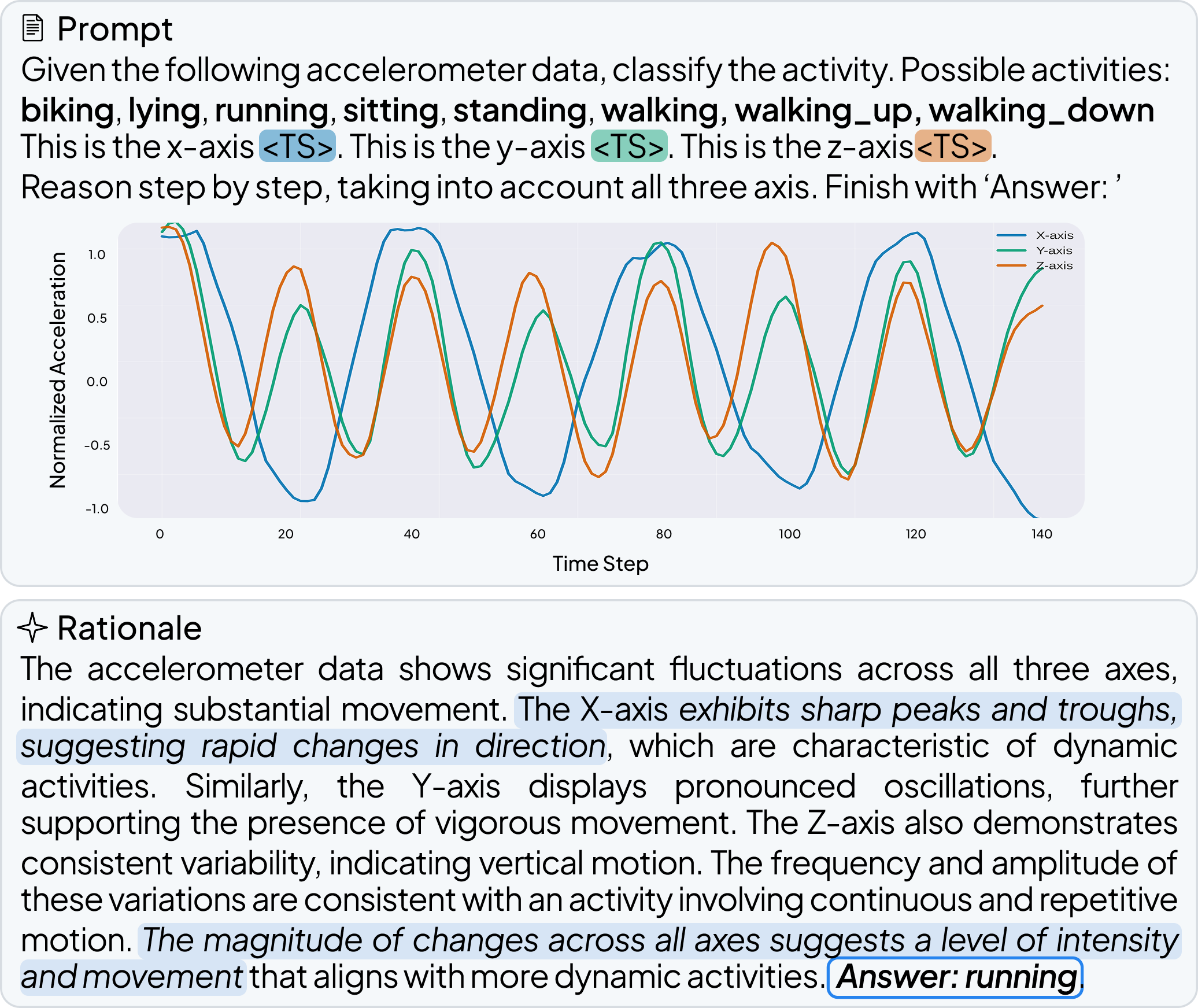}
        \caption{Human Activity Recognition \gls{CoT}}
        \label{fig:cot_rationale:har}
    \end{subfigure}
    \begin{subfigure}{0.3\linewidth}
        \includegraphics[width=1.0\linewidth]{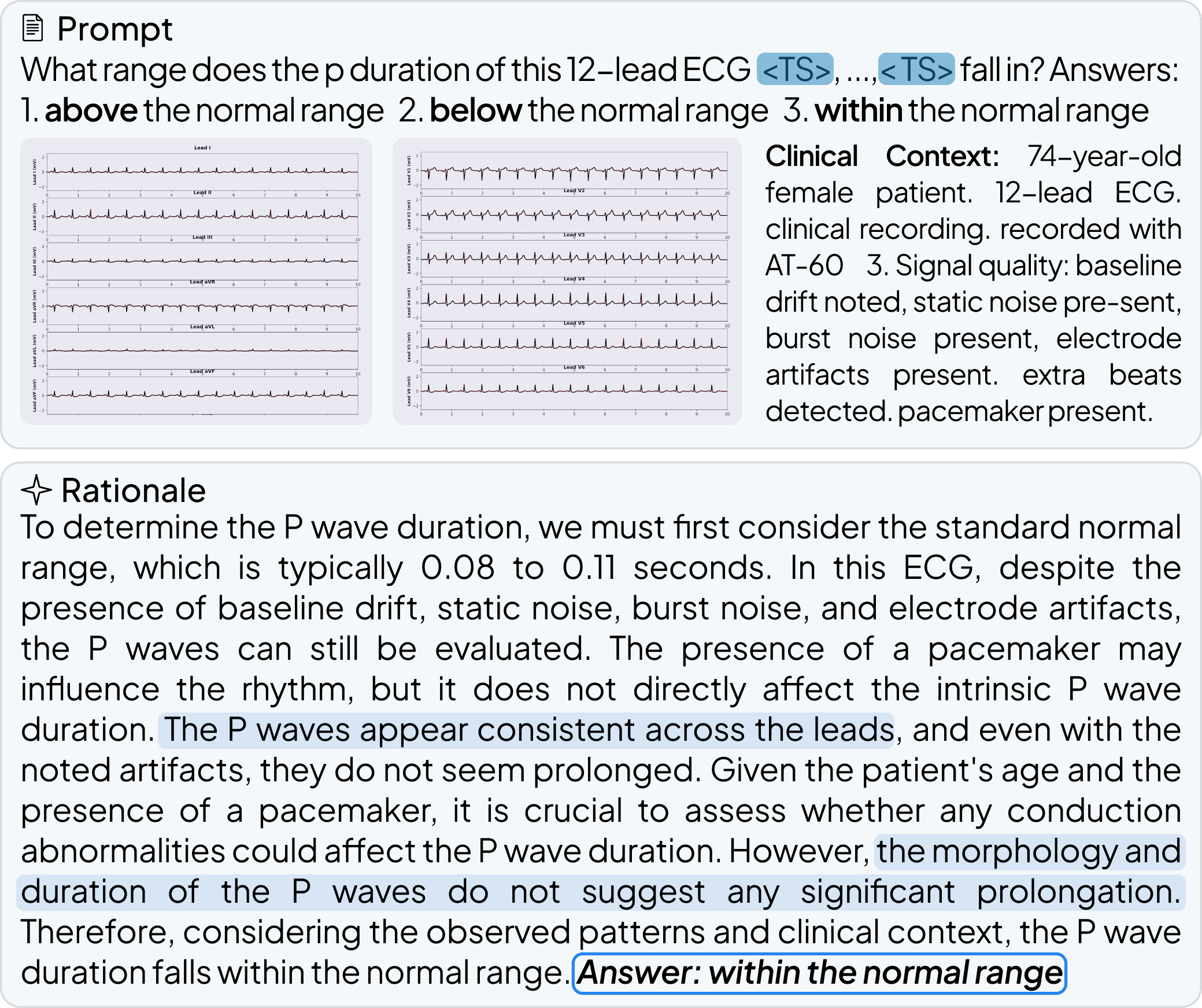}
        \caption{ECG Interpretation \gls{CoT}}
        \label{fig:cot_rationale:ecg}
    \end{subfigure}
    \begin{subfigure}{0.3\linewidth}
        \includegraphics[width=1.05\linewidth]{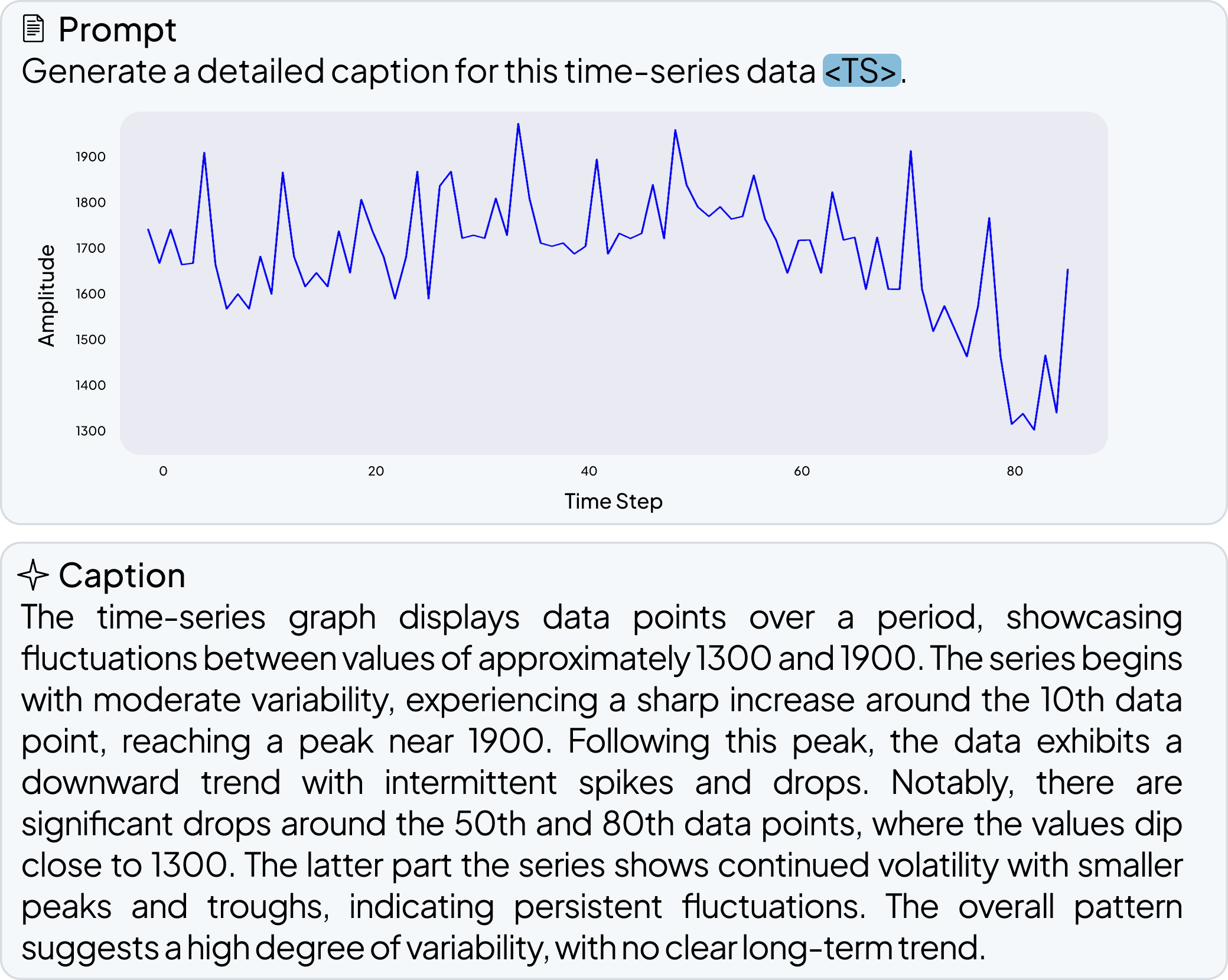}
        \caption{M4 Time-Series Captioning}
        \label{fig:captioning:m4}
    \end{subfigure}

    \caption{Example \gls{CoT} rationales for HAR, Sleep Staging, ECG-QA and M4 captioning, generated with \gls{OpenTSLM}-Flamingo/Llama3.2-1B.  
    More examples are provided in \autoref{sec:appendix:rationales}.}
    \label{fig:cot_rationale}
\end{figure*}

Both \textbf{\gls{OpenTSLM}} variants remain text models, trained to generate rationales for classification rather than outputting only a class label.  
\autoref{fig:cot_rationale:har} shows example rationales for \textit{human activity recognition} and
\autoref{fig:cot_rationale:ecg} for ECG-QA.
\autoref{fig:captioning:m4} shows time-series captioning on M4 data.

To evaluate the quality of model rationales, we conducted an expert review with five cardiologists from \anon{Stanford} Hospital on rationales generated by OpenTSLM-Flamingo-Llama3.2-3B (best model) for ECG-QA.
\begin{figure}[h!t]
    \centering
    \begin{subfigure}{0.45\linewidth}
        \centering
        \includegraphics[width=\linewidth]{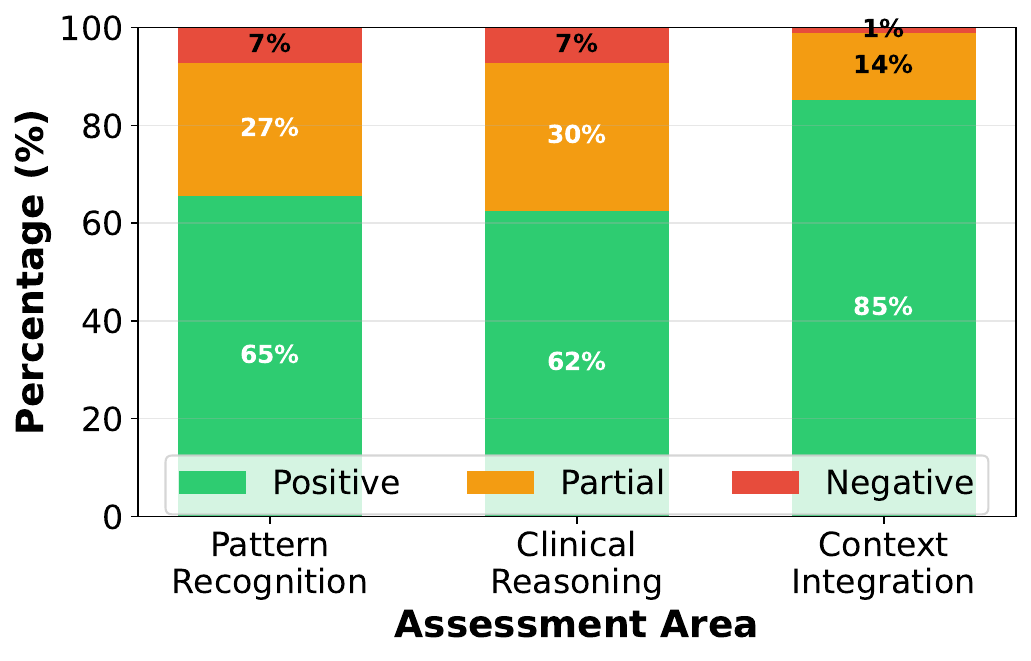}
        \caption{Performance by Area}
        \label{fig:performance_by_area}
    \end{subfigure}%
    \hspace{0.5cm}
    \begin{subfigure}{0.45\linewidth}
        \centering
        \includegraphics[width=\linewidth]{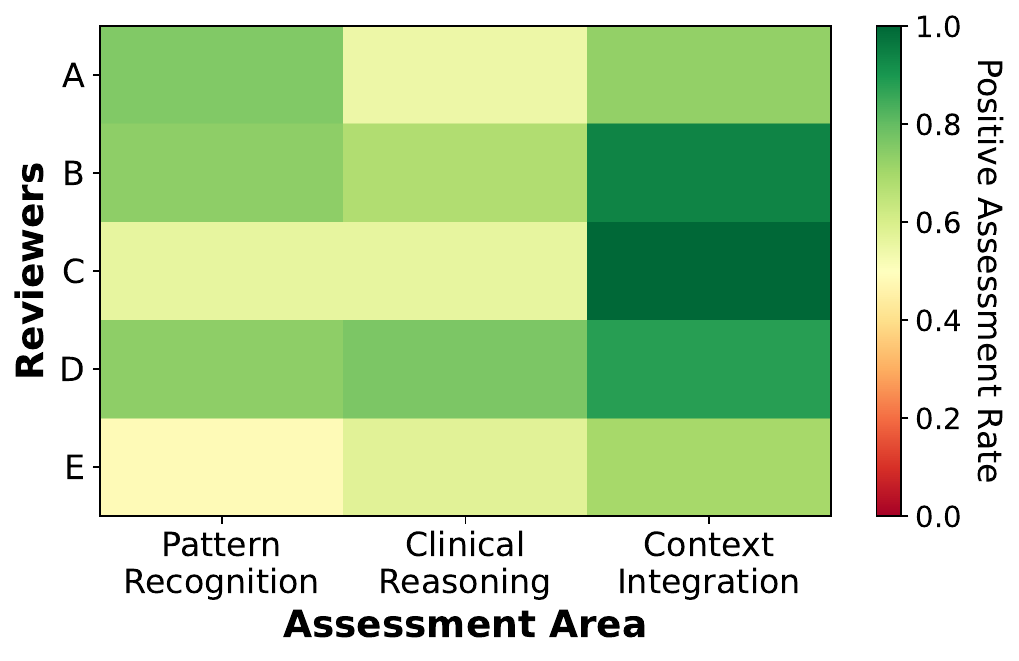}
        \caption{Eval Distribution}
        \label{fig:overall_distribution}
    \end{subfigure}
    \caption{Qualitative evaluation of \gls{CoT} rationales and inter-reviewer agreement patterns.}
    \vspace{-2em}
    \label{fig:model_assessment_overview}
\end{figure}
We randomly sampled two examples per template (84 total), each reviewed by at least two cardiologists. 
Evaluation followed a rubric derived from the American College of Cardiology/American Heart Association Clinical Competence Statement on \glspl{ECG}~\cite{L1999Nov, ACCAHA} and based on the RIME (“Reporter–Interpreter–Manager–Educator”) framework~\cite{L1999Nov} (see \ref{sec:appendix:ecgEval}), assessing whether the model:
(1) correctly identified relevant ECG features;
(2) appropriately connected them to the final answer;
(3) incorporated patient context (age, artifacts, ...).
Overall, the model gave a correct or partially correct ECG interpretation in 92.9\% of cases, spanning ECG recognition, reasoning, and contextualization. 
\gls{OpenTSLM} showed strongest performance in clinical context integration (85.1\% positive) compared to ECG pattern recognition (65.5\% positive) and clinical reasoning (62.5\% positive) (\autoref{fig:performance_by_area}).
Assessment patterns varied notably across reviewers, with some reviewers consistently more favorable across all evaluation areas (\autoref{fig:overall_distribution}).
Reviewer disagreement was most common for clinical reasoning, where moderate disagreements were observed between adjacent assessment categories.
Complete disagreements between positive and negative assessments were relatively rare across all areas (\autoref{fig:disagreement_patterns} in \ref{sec:appendix:ecgEval}).

\subsection{OpenTSLM with pretrained time-series encoder}

Motivated by the strong performance of existing pretrained \glspl{TSFM}, such as Chronos-2~\cite{ansari2025chronos}, we further investigate their compatibility with the OpenTSLM framework~(see~\autoref{sec:methods:time_series_encoder}).
Table~\ref{tab:pretrained_TS_enc} shows that using OpenTSLM-Flamingo with pretrained Chronos-2 generally improves performance across all LLM backbones and datasets. In particular, for TSQA, even relatively small LLMs (e.g., Gemma3-270M) benefit substantially, with performance improving from 77.86\% to 99.69\%. For Sleep-CoT, pretrained TSFMs lead to substantial F1 gains across all backbones, e.g., improvements from 49.33 to 64.57 for Llama3.2-1B and from 51.38 to 68.33 for Gemma3-270M. 
We also observe moderate  F1 improvements on ECG-QA-CoT (e.g., from 34.62 to 41.92 for Llama3.2-1B).

\begin{table}[h!t]
\centering
\caption{OpenTSLM-Flamingo with different LLM bases and a pretrained encoder time-series encoder. }
\label{tab:pretrained_TS_enc}
\resizebox{\linewidth}{!}{%
\begin{tabular}{@{}lllllllll@{}}
\toprule
& \multicolumn{2}{c}{TSQA} & \multicolumn{2}{c}{HAR-CoT} & \multicolumn{2}{c}{Sleep-CoT} & \multicolumn{2}{c}{ECG-QA-CoT} \\ 
& F1          & Acc        & F1           & Acc          & F1            & Acc           & F1             & Acc           \\
\midrule
Llama3.2-1B          & 94.08       & 94.00      & 62.93        & 69.27        & 49.33         & 67.31         & 34.62          & 38.14         \\
Llama3.2-1B-Chronos2 & \textbf{99.71}       & \textbf{99.71}      & 66.82        & 70.61        & 64.57         & 78.54        & \textbf{41.92}               & 44.01               \\
\midrule
Llama3.2-3B      & 90.14 & 90.10 & 62.77 & 69.03 & 45.45 & 69.14 & 40.25 & \textbf{46.25} \\

Llama3.2-3B-Chronos2 & 99.58       & 99.58       & \textbf{67.64}        & \textbf{72.20}        & 61.52          & 78.31       & 41.89                & 44.27                \\
\midrule
Gemma3-270M          & 77.86       & 78.12      &  57.75       & 63.43         & 51.38         & 68.49         &  32.71              & 35.50               \\
Gemma3-270M-Chronos2 & 99.69       & 99.69      & 58.65       & 63.90        & \textbf{68.33}         & \textbf{80.69}         & 35.92             & 38.37             \\
\midrule
Gemma3-1B-pt & 92.56       & 92.46       & 65.44        & 71.48         & 43.69         & 60.67          & 35.31                & 37.79              \\
Gemma3-1B-pt-Chronos2 & \textbf{99.71}       & \textbf{99.71}      & 62.72        & 67.50        & 61.13         & 75.96         &  40.25              & 42.47              \\
\end{tabular}
}
\vspace{-1.5em}
\end{table}
\section{Discussion}
\label{sec:discussion}

All OpenTSLM models consistently outperform baselines. Text-only models often fail to follow the answer template and thus perform at or below chance~(\autoref{sec:methods:training}). 
Finetuned baselines improve substantially on HAR-CoT (60.44\% F1 vs. 0\% for Llama-3.2-1B) but only slightly on Sleep-CoT (9.05 vs. 2.14). ECG-QA finetuning was infeasible due to high VRAM demands (80k tokens require \textgreater 100GB per sample). 
OpenTSLM-SoftPrompt performs best on shorter sequences (Sleep-CoT, TSQA) but becomes impractical as VRAM requirements grow with sequence length (\textgreater 180GB in simulations with 10,000-length series). With soft prompting, smaller models like Gemma-3 270M and 1B quickly exhaust their context and underperform. 
In contrast, OpenTSLM-Flamingo sustains stable memory across sequence lengths and series (up to 60GB for Llama-3.2-3B with five 10,000-length series). 
This allows even tiny models, such as Gemma-270M, to deliver strong results, highlighting the efficiency of cross-attention for treating time-series as a native modality.
Leveraging pretrained \glspl{TSFM} such as Chronos-2 further improves performance, in some cases substantially, for example, TSQA accuracy increases from 77.86\% to 99.69\% with Gemma-3 270M. OpenTSLM is compatible with any pretrained \gls{TSFM} via a trainable projection layer.

\noindent\textbf{Practical implications.}
Our results show that even frontier \glspl{LLM} like GPT-4o are poorly suited for time-series reasoning and that time-series must be treated as a distinct modality. 
With OpenTSLM, even small models like Gemma3 270M outperform GPT-4o ($\sim$200B parameters~\cite{chatgpt_4o_estimation}) at a fraction of the compute and cost, enabling efficient on-device or mobile deployment. 
We recommend using OpenTSLM-SoftPrompt for short time-series, where it delivers strong performance while requiring only a small number of additional parameters during finetuning. 
However, because SoftPrompt’s memory usage grows exponentially with sequence length, it becomes impractical for longer horizons or multi-series inputs.
In contrast, we recommend OpenTSLM-Flamingo for longer time-series and multivariate sensor data (e.g, 12-lead ECG, 3-axis IMU) and as a general-purpose solution, as it maintains nearly constant memory consumption across extended or multi-series contexts, and offers better performance on complex datasets~(like ECG-QA).
Perhaps the greatest advantage of \glspl{TSLM} is the interface they provide for contextualizing results. 
In ECG-QA, OpenTSLM correctly identified the relevant ECG features in most cases, with missing context only 7.1\% of the time.
The model demonstrated particularly strong clinical context integration (85.1\% positive assessments), thereby offering clinicians and researchers a transparent window into the model's reasoning.
As trust is important in medicine, this transparency underscores the value of applying \glspl{LLM} to time-series.\\
\noindent\textbf{Comparison with prior work.}
Our approach differs from prior work in several ways. 
First, we introduce time-series as a new modality for \glspl{LLM}, unlike ~\cite{health_prompt} and \cite{yubin_kim_2024_health_llm}, which tokenize time-series. 
Second, we frame tasks as joint text–time-series reasoning, training models to generate rationales that integrate temporal information. 
This contrasts with MedualTime~\cite{medualtime} and Time2Lang~\cite{time2lang}, which reprogrammed \glspl{LLM} with fixed classification or forecasting heads, removing language generation capabilities. 
Notably, OpenTSLM achieves 40.25 F1 on ECG-QA-CoT, producing rationales across 3,138 questions and 42 templates with diverse answer options. 
By comparison, \citeauthor{medualtime} report 76 F1 on PTB-XL~(underlying dataset of ECG-QA) with only four classes and a fixed classification head~\cite{medualtime}. 
Third, unlike SensorLM~\cite{sensorlm}, which is trained from scratch, our models build on pretrained open-weight \glspl{LLM}, retaining pretrained knowledge. 
Fourth, while prior work used soft prompting~\cite{apple_health_ai_time_series_reasoning, wang2025itformerbridgingtimeseries} to model time-series implicitly by concatenating text-tokens with derived time-series tokens, we find that this approach scales poorly in memory use. 
In contrast, our \gls{OpenTSLM}-Flamingo approach models time-series explicitly via a separate encoding integrated via cross-attention, scaling better to long sequences.\\
\noindent\textbf{Limitations.}  
We acknowledge several limitations. 
First, our method of encoding time-series may not be optimal, as we rely on including the mean and standard deviation in accompanying texts to preserve the temporal scale. 
Second, we generated CoT datasets using GPT-4o on plots, which we have shown to perform poorly on these plots alone. Curated datasets likely lead to better rationales. 
Third, framing tasks as natural language generation does not ensure that the model prioritizes the correct label after presenting its rationale, underscoring the need for loss functions that explicitly enforce correct answers.
Finally, we only assessed the impact of Chronos-2 as a pretrained \gls{TSFM} and did not evaluate the impact of using other state-of-the-art models as encoders on performance, which we leave for future work.
\section{Conclusion}
\label{sec:conclusion}

Our results show that both \gls{OpenTSLM} variants enable small-scale \glspl{LLM} to outperform much larger text-only models on time-series tasks, demonstrating that lightweight, domain-adapted architectures can achieve strong performance without massive model scales. 
With \gls{OpenTSLM}, we extend open-weight pretrained \glspl{LLM} to process time-series, retaining knowledge while adapting them to temporal domains. 
This work may lay the foundation for general-purpose \glspl{TSLM} capable of handling diverse time-series datasets. 
Although our focus is healthcare, the ability to reason over longitudinal data has broad relevance in domains such as finance, supply chain management, and industrial monitoring.

\bibliography{iclr2026_conference}
\bibliographystyle{icml2026}

\newpage
\appendix
\onecolumn
\section{Appendix}
\label{sec:appendix}

\subsection{Training details}
\label{sec:appendix:training_details}

\autoref{tab:datasets} provides an overview of the datasets used during training.
\begin{table}[h!]
\centering
\footnotesize
\begin{tabularx}{\textwidth}{lllXXX}
 & \textbf{Dataset} & \textbf{\#Samples (Train/Val/Test)} & \textbf{Num series} & \textbf{Length} & \textbf{Frequency} \\
\midrule
\multirow{2}{*}{\rotatebox{90}{Stage 1}} 
  & \gls{TSQA}\textsuperscript{*1} & 38,400 / 4,800 / 4,800 & 1 & Hours to Years & Not specified \\
  & M4-Captions & 80,000 / 10,000 / 10,000 & 1 & 64-512 points & Not specified \\            \addlinespace[6pt]
\midrule
\multirow{3}{*}{\rotatebox{90}{Stage 2}} 
  & HAR-\gls{CoT} & 68,542 / 8,718 / 8,222 & 3 & 2.56s & 50Hz \\
  & Sleep-\gls{CoT} & 7,434 / 930 / 930 & 1 & 30s & 100Hz \\
  & ECG-QA-\gls{CoT} & 159,313 / 31,137 / 41,093 & 12 & 10s & 100Hz \\
\end{tabularx}
\caption{\textsuperscript{*1}TSQA~\cite{tsqa}
Overview of datasets used in Stage 1 (pretraining tasks) and Stage 2 (task-specific CoT reasoning). Datasets are split in 80/10/10 ration.}
\label{tab:datasets}
\end{table}
All data was split into ratios of 80/10/10 (train/val/test).

\subsubsection{Training Configuration}

The models were trained with the following configuration:

\begin{itemize}
    \item \textbf{Optimizer:} AdamW
    \item \textbf{Learning Rates:}
    \begin{itemize}
        \item \textbf{OpenTSLM-SP:}  
        \begin{itemize}
            \item Time-series encoder: $2 \times 10^{-4}$
            \item LoRA: $2 \times 10^{-4}$
            \item Projector: $1 \times 10^{-4}$
        \end{itemize}
        \item \textbf{OpenTSLM-Flamingo:}  
        \begin{itemize}
            \item Encoder: $2 \times 10^{-4}$
            \item Cross-attention layers: $2 \times 10^{-4}$
        \end{itemize}
    \end{itemize}
    \item \textbf{Scheduler:} Linear learning rate schedule with warmup  
    \item \textbf{Warmup:} $10\%$ of total training steps
    \item \textbf{Gradient Clipping:} $\ell_2$-norm capped at 1.0
    \item \textbf{Weight Decay:} $0.01$
    \item \textbf{Training Length:} Up to 200 epochs with early stopping (patience = 5 epochs)
\end{itemize}

Learning rate choices were informed by~\cite{apple_health_ai_time_series_reasoning}. For pretrained Chronos-2~\citep{ansari2025chronos}, we directly download the model from \href{https://huggingface.co/amazon/chronos-2}{https://huggingface.co/amazon/chronos-2}. For learning rates for OpenTSLM-Flamingo with chronos-2, we keep as training from scratch.

\subsection{Generation of multivariate time-series \gls{CoT} datasets}
\label{sec:appendix:datasets}

This section provides detailed descriptions of the \gls{CoT} datasets generated for our study: Human Activity Recognition (HAR-CoT), Sleep Stage Classification (SleepEDF-CoT), and Electrocardiogram Question Answering (ECG-QA-CoT).

Our objective was to enable \glspl{TSLM} not only to classify time-series but also to generate explicit reasoning chains. Since few datasets include \gls{CoT} text, we generated our own multivariate time-series \gls{CoT} datasets using widely adopted benchmarks in \gls{HAR}, sleep staging, and \gls{ECG}-\gls{QA}, following a similiar approach as proposed by~\cite{apple_health_ai_time_series_reasoning}.

For each dataset, we generated rationales with GPT-4o by providing a plot of the data along with the correct label, and prompting the model to produce a rationale leading to that label. The exact prompts are described in Sections~\ref{sec:appendix:har_generation}, \ref{sec:appendix:sleep_generation}, and \ref{sec:appendix:ecg_qa_generation}. We carefully engineered the prompts and manually reviewed a subset of samples to ensure the generated rationales were consistent and sensible. When plotting, original data was used without normalization. If multiple time-series were present in a sample (e.g., three in HAR or twelve in ECG), all were plotted as separate subplots but combined into a single figure.

\begin{itemize}
    \item \textbf{GPT-4o snapshot:} gpt-4o-2024-08-06
    \item \textbf{Temperature:} 0.3
    \item \textbf{Seed:} 42
\end{itemize}

The following subsections describe dataset-specific methodologies, data processing, prompts, answer selection, and final class distributions.

\subsubsection{Human Activity Recogniton~(HAR) CoT}
\label{sec:appendix:har_generation}

We merged multiple \gls{HAR} datasets spanning DaLiAc~\cite{daliac_dataset}, DOMINO~\cite{domino_dataset}, HHAR~\cite{hhar_dataset}, 
PAMAP2~\cite{pamap}, RealWorld~\cite{realworld_har_dataset}, and datastes from \cite{shoaib2013_uthar,shoaib2014_utsensors, shoaib2016_utcomplex}. We retain only those activity classes present in all datasets. The final dataset includes eight activity classes: sitting, walking, standing, running, walking up stairs, walking down stairs, lying, and biking. The data is split into 2.56 second windows.

\paragraph{Data Processing}

The dataset was processed to create 2.56-second windows of triaxial accelerometer data (X, Y, Z axes). Each sample was visualized as a multi-panel plot showing the acceleration signals across all three axes over the time window.

\paragraph{Prompt for CoT generation}
We generated CoT rationales by prompting the model with a correct and dissimilar label. The following prompt template was used for HAR-CoT generation:

\begin{lstlisting}
You are shown a time-series plot of accelerometer over a 2.56 second window. 
This data corresponds to one of two possible activities:
[CORRECT_ACTIVITY]
[DISSIMILAR_ACTIVITY]

Your task is to classify the activity based on analysis of the data.

Instructions:
- Begin by analyzing the time-series without assuming a specific label.
- Think step-by-step about what the observed patterns suggest regarding movement intensity and behavior.
- Write your rationale as a single, natural paragraph, do not use bullet 
  points, numbered steps, or section headings.
- Do not refer back to the plot or to the act of visual analysis in your  rationale; the plot is only for reference but you should reason about the 
  time-series data.
- Do **not** assume any answer at the beginning,  analyze as if you do not 
  yet know which class is correct.
- Do **not** mention either class label until the final sentence.
- Make sure that your last word is the answer. You MUST end your response 
  with "Answer: [CORRECT_ACTIVITY]":
\end{lstlisting}

\begin{figure}[h!t]
    \centering
    \includegraphics[width=\textwidth]{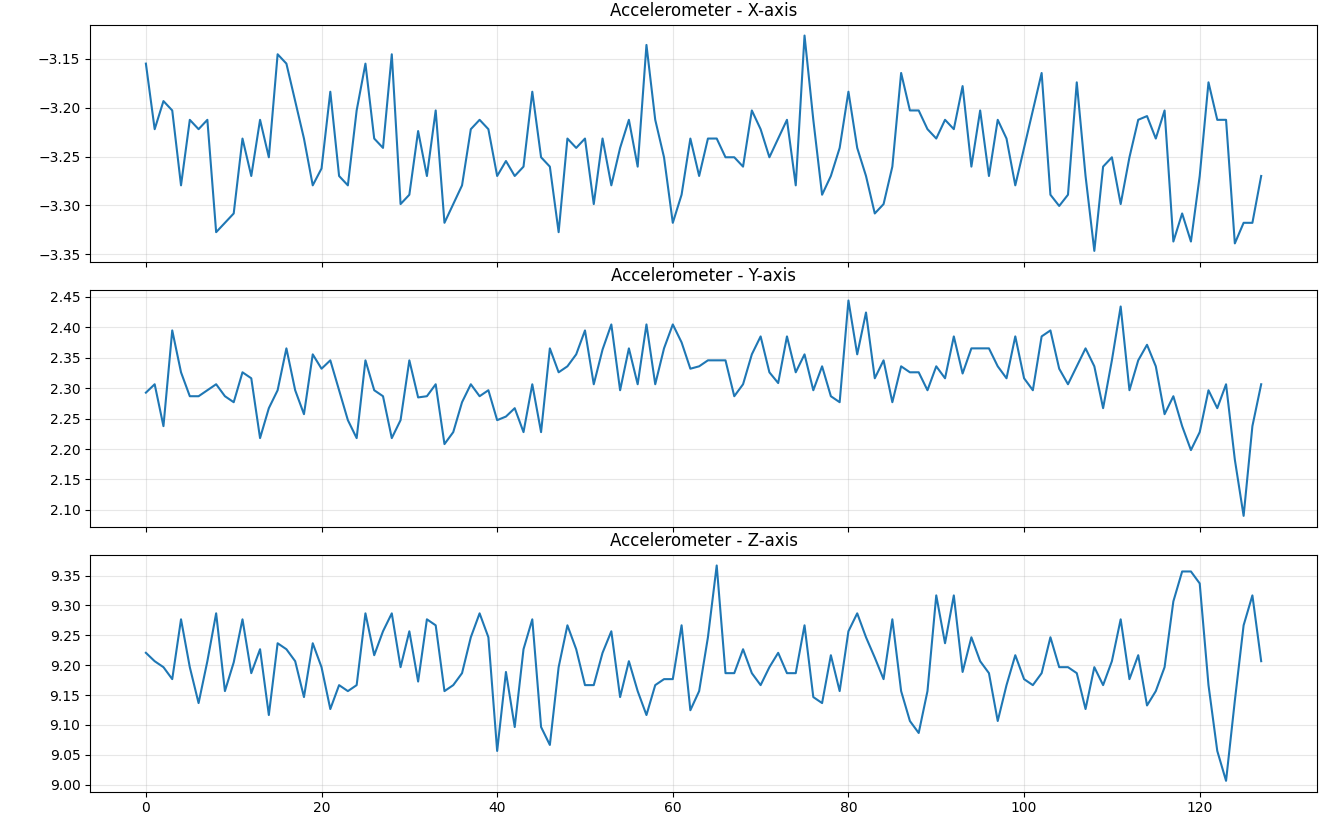}
    \caption{Sample HAR signal input to GPT-4o for rationale generation}
    \label{fig:har_cot_generation}
\end{figure}

\paragraph{Answer Selection Strategy}

For each sample, we implemented a dissimilarity-based answer selection strategy. Given a correct activity label, we selected the most dissimilar activity from a predefined mapping:

\begin{itemize}
\item \textbf{Sitting}: walking, running, biking, walking up, walking down
\item \textbf{Walking}: sitting, lying, standing, biking, running
\item \textbf{Standing}: walking, running, biking, walking up, walking down
\item \textbf{Running}: sitting, lying, standing, biking, walking
\item \textbf{Walking up}: sitting, lying, standing, biking, running
\item \textbf{Walking down}: sitting, lying, standing, biking, running
\item \textbf{Lying}: walking, running, biking, walking up, walking down
\item \textbf{Biking}: sitting, lying, standing, walking, running
\end{itemize}

This strategy ensured that the binary classification tasks were challenging and required genuine analysis of movement patterns rather than simple pattern recognition.

\paragraph{Label distribution}

\begin{table}[htbp]
\centering
\caption{Per-class sample distribution for HAR-CoT train, validation, and test sets}
\begin{tabularx}{\textwidth}{l *{3}{>{\raggedleft\arraybackslash}X}}
\toprule
\textbf{Class} & \textbf{Train (n=68542)} & \textbf{Val (n=8718)} & \textbf{Test (n=8222)} \\
\midrule
Biking               & 4037 (5.9\%)   & 435 (5.0\%)   & 473 (5.8\%)   \\
Lying                & 4305 (6.3\%)   & 682 (7.8\%)   & 444 (5.4\%)   \\
Running              & 8101 (11.8\%)   & 948 (10.9\%)   & 1057 (12.9\%)   \\
Sitting              & 18997 (27.7\%)   & 2315 (26.6\%)   & 2342 (28.5\%)   \\
Standing             & 11001 (16.1\%)   & 1449 (16.6\%)   & 1264 (15.4\%)   \\
Walking              & 12675 (18.5\%)   & 1611 (18.5\%)   & 1508 (18.3\%)   \\
Walking Down         & 4514 (6.6\%)   & 710 (8.1\%)   & 542 (6.6\%)   \\
Walking Up           & 4912 (7.2\%)   & 568 (6.5\%)   & 592 (7.2\%)   \\
\bottomrule
\end{tabularx}
\end{table}

\subsubsection{Sleep Stage Classification Chain-of-Thought (SleepEDF-CoT)}
\label{sec:appendix:sleep_generation}

The SleepEDF-CoT dataset was generated from the Sleep-EDF database, which contains polysomnography recordings with expert-annotated sleep stage labels. The dataset includes five sleep stages: Wake (W), Non-REM stage 1 (N1), Non-REM stage 2 (N2), Non-REM stage 3 (N3), and REM sleep (REM).

\paragraph{Data Processing}

The dataset was processed to create 30-second windows of EEG data from the Fpz-Cz channel. Each sample was visualized as a single-channel EEG plot showing brain activity patterns characteristic of different sleep stages.

\paragraph{Prompt for CoT generation}
We generated CoT rationales by prompting the model with a correct and dissimilar label. The following prompt template was used for SleepEDF-CoT generation:

\begin{lstlisting}
You are presented with a time-series plot showing EEG data collected over a 30-second interval. This signal corresponds to one of two possible sleep stages:
- [SLEEP_STAGE_1]
- [SLEEP_STAGE_2]

Your task is to determine the correct sleep stage based solely on the observed patterns in the time series.

Instructions:
- Analyze the data objectively without presuming a particular label.
- Reason carefully and methodically about what the signal patterns suggest 
  regarding sleep stage.
- Write your reasoning as a single, coherent paragraph. Do not use bullet points, lists, or section headers.
- Do not reference the plot, visuals, or the process of viewing the data in your explanation; focus only on the characteristics of the time series.
- Do not mention or speculate about either class during the rationale, only reveal the correct class at the very end.
- Never state that you are uncertain or unable to classify the data. You must always provide a rationale and a final answer.
- Your final sentence must conclude with: "Answer: [CORRECT_SLEEP_STAGE]"
\end{lstlisting}

\begin{figure}[h!t]
    \centering
    \includegraphics[width=\textwidth]{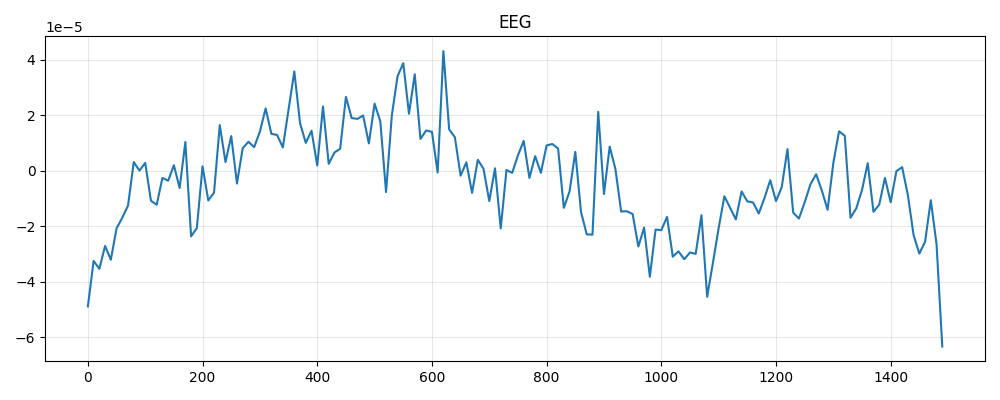}
    \caption{Sample EEG signal input to GPT-4o for sleep stage rationale generation}
    \label{fig:sleep_cot_generation}
\end{figure}

\paragraph{Answer Selection Strategy}

For sleep stage classification, we implemented a dissimilarity-based strategy that pairs physiologically distinct sleep stages:

\begin{itemize}
\item \textbf{Wake (W)}: N3, N4, REM
\item \textbf{N1}: W, N3, N4
\item \textbf{N2}: W, REM
\item \textbf{N3}: W, REM
\item \textbf{N4}: W, REM
\item \textbf{REM}: N2, N3, N4
\end{itemize}

This approach ensured that the binary classification tasks required understanding of fundamental differences in brain activity patterns between sleep stages.

\paragraph{Label distribution}
SleepEDF dataset
\begin{table}[htbp]
\centering
\caption{Per-class sample distribution for train, validation, and test sets (Sleep stages)}
\begin{tabularx}{\textwidth}{l *{3}{>{\raggedleft\arraybackslash}X}}
\toprule
\textbf{Label} & \textbf{Train (n=7434)} & \textbf{Val (n=930)} & \textbf{Test (n=930)} \\
\midrule
Non-REM 1     & 410 (5.5\%)   & 52 (5.6\%)   & 51 (5.5\%)   \\
Non-REM 2     & 2057 (27.7\%) & 257 (27.6\%) & 257 (27.6\%) \\
Non-REM 3     & 357 (4.8\%)   & 45 (4.8\%)   & 45 (4.8\%)   \\
Non-REM 4     & 299 (4.0\%)   & 37 (4.0\%)   & 38 (4.1\%)   \\
REM    & 944 (12.7\%)  & 118 (12.7\%) & 118 (12.7\%) \\
Wake      & 3367 (45.3\%) & 421 (45.3\%) & 421 (45.3\%) \\
\bottomrule
\end{tabularx}
\end{table}

\clearpage

\subsubsection{Electrocardiogram Question Answering Chain-of-Thought (ECG-QA-CoT)}
\label{sec:appendix:ecg_qa_generation}

The ECG-QA-CoT dataset was generated from the PTB-XL~\cite{ptbxl} database combined with the ECG-QA~\cite{ecg_qa} question templates. This dataset contains 12-lead ECG recordings with clinical questions covering various aspects of cardiac analysis, including rhythm analysis, morphology assessment, and diagnostic classification.

\paragraph{Data Processing}

The dataset was processed to create complete 12-lead ECG recordings (I, II, III, aVR, aVL, aVF, V1, V2, V3, V4, V5, V6) sampled at 100 Hz. Each ECG was visualized as a multi-panel plot showing all 12 leads simultaneously, enabling comprehensive cardiac analysis.

\paragraph{Prompt for CoT generation}

The following prompt template was used for ECG-QA-CoT generation:

\begin{lstlisting}
You are presented with a complete 12-lead ECG recording showing all standard leads (I, II, III, aVR, aVL, aVF, V1, V2, V3, V4, V5, V6).

Clinical Context: [CLINICAL_CONTEXT]

Question: [QUESTION]

This question has one of two possible answers:
- [ANSWER_OPTION_1]
- [ANSWER_OPTION_2]

Your task is to analyze the ECG and determine the correct answer based on the observed cardiac patterns. You may include the clinical context in your analysis if it helps you determine the correct answer.

Instructions:
- Analyze the ECG systematically without presuming a particular answer.
- Consider rhythm, rate, morphology, intervals, and any abnormalities you observe across all 12 leads.
- Think step-by-step about what the ECG patterns indicate regarding the clinical question above.
- Write your reasoning as a single, coherent paragraph. Do not use bullet points, lists, or section headers.
- Do not reference the visual aspects of viewing the ECG plot; focus on the cardiac characteristics and clinical significance.
- Do not mention or assume either answer option during your rationale, only reveal the correct answer at the very end.
- NEVER state uncertainty or inability to determine the answer. You MUST always provide clinical reasoning and a definitive answer.
- Your final sentence must conclude with: "Answer: [CORRECT_ANSWER]"
\end{lstlisting}

\begin{figure}[h!t]
    \centering
    \includegraphics[width=\textwidth]{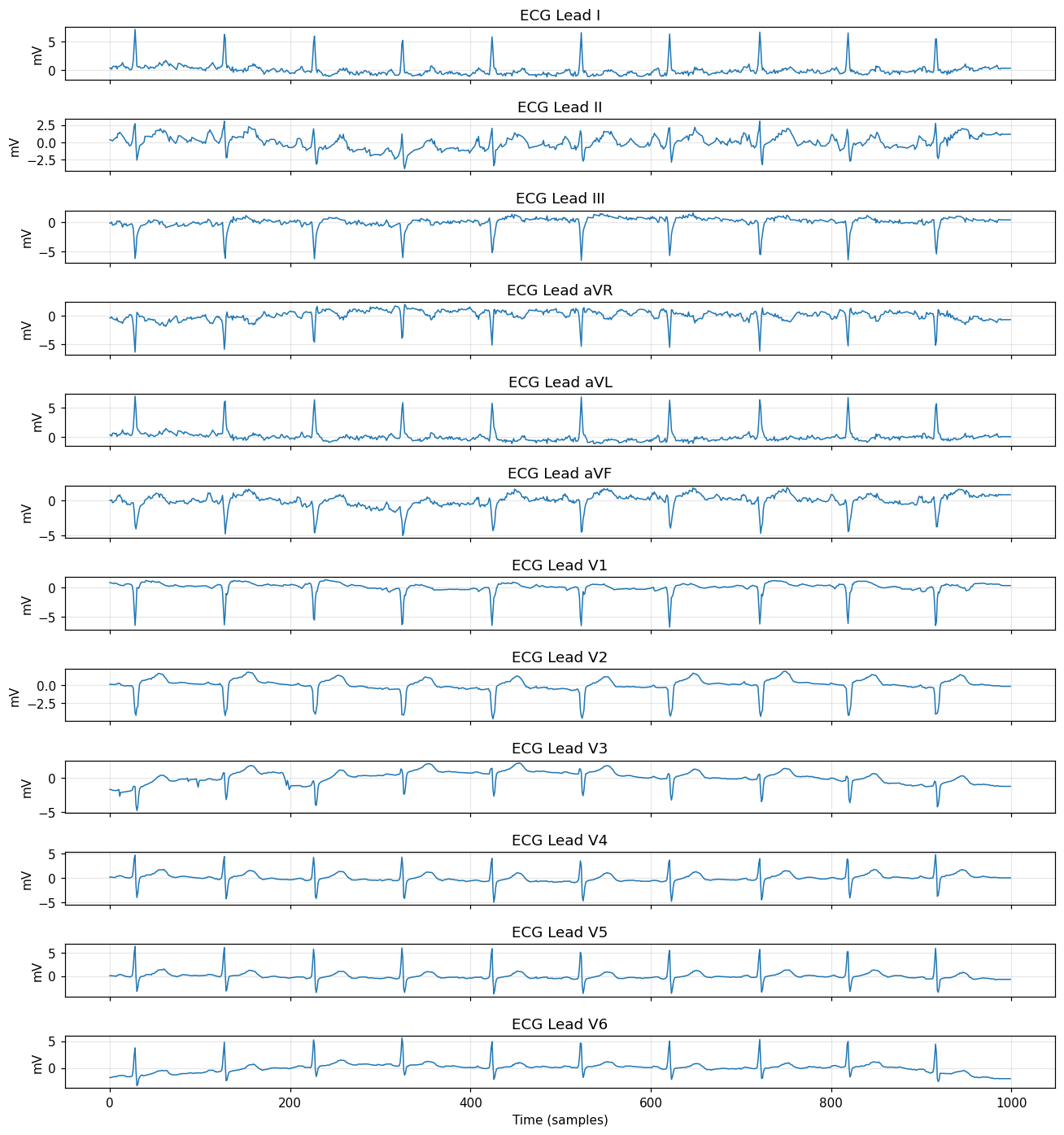}
    \caption{Sample ECG signal input to GPT-4o for rationale generation}
    \label{fig:ecg_cot_generation}
\end{figure}

\clearpage
\paragraph{Label distribution}

\begin{table}[h!t]
\centering
\caption{Per-template sample distribution for ECG-QA CoT train, validation, and test sets}
\begin{tabularx}{\textwidth}{l *{3}{>{\raggedleft\arraybackslash}X}}
\toprule
\textbf{Template ID} & \textbf{Train (n=159,306)} & \textbf{Val (n=31,137)} & \textbf{Test (n=41,093)} \\
\midrule
Template 1  & 17,089 (10.7\%) & 2,924 (9.4\%)   & 3,467 (8.4\%)   \\
Template 2  & 300 (0.2\%)     & 60 (0.2\%)      & 60 (0.1\%)      \\
Template 3  & 240 (0.2\%)     & 48 (0.2\%)      & 48 (0.1\%)      \\
Template 4  & 20,861 (13.1\%) & 3,782 (12.1\%)  & 4,096 (10.0\%)  \\
Template 5  & 20,104 (12.6\%) & 3,599 (11.6\%)  & 3,905 (9.5\%)   \\
Template 6  & 5,356 (3.4\%)   & 1,022 (3.3\%)   & 1,085 (2.6\%)   \\
Template 7  & 1,137 (0.7\%)   & 221 (0.7\%)     & 224 (0.5\%)     \\
Template 8  & 4,371 (2.7\%)   & 747 (2.4\%)     & 1,466 (3.6\%)   \\
Template 9  & 3,563 (2.2\%)   & 610 (2.0\%)     & 1,200 (2.9\%)   \\
Template 10 & 894 (0.6\%)     & 311 (1.0\%)     & 377 (0.9\%)     \\
Template 11 & 2,861 (1.8\%)   & 533 (1.7\%)     & 964 (2.3\%)     \\
Template 12 & 300 (0.2\%)     & 60 (0.2\%)      & 60 (0.1\%)      \\
Template 13 & 300 (0.2\%)     & 60 (0.2\%)      & 60 (0.1\%)      \\
Template 14 & 300 (0.2\%)     & 60 (0.2\%)      & 60 (0.1\%)      \\
Template 15 & 300 (0.2\%)     & 60 (0.2\%)      & 60 (0.1\%)      \\
Template 16 & 300 (0.2\%)     & 60 (0.2\%)      & 60 (0.1\%)      \\
Template 17 & 19,952 (12.5\%) & 3,013 (9.7\%)   & 4,416 (10.7\%)  \\
Template 18 & 9,580 (6.0\%)   & 2,178 (7.0\%)   & 3,806 (9.3\%)   \\
Template 19 & 4,122 (2.6\%)   & 698 (2.2\%)     & 1,395 (3.4\%)   \\
Template 20 & 1,200 (0.8\%)   & 228 (0.7\%)     & 237 (0.6\%)     \\
Template 21 & 180 (0.1\%)     & 36 (0.1\%)      & 36 (0.1\%)      \\
Template 22 & 400 (0.3\%)     & 131 (0.4\%)     & 167 (0.4\%)     \\
Template 23 & 744 (0.5\%)     & 126 (0.4\%)     & 168 (0.4\%)     \\
Template 24 & 90 (0.1\%)      & 18 (0.1\%)      & 18 (0.0\%)      \\
Template 25 & 399 (0.3\%)     & 160 (0.5\%)     & 178 (0.4\%)     \\
Template 26 & 10,585 (6.6\%)  & 1,894 (6.1\%)   & 2,193 (5.3\%)   \\
Template 27 & 1,038 (0.7\%)   & 180 (0.6\%)     & 210 (0.5\%)     \\
Template 28 & 3,600 (2.3\%)   & 720 (2.3\%)     & 720 (1.8\%)     \\
Template 29 & 300 (0.2\%)     & 60 (0.2\%)      & 60 (0.1\%)      \\
Template 30 & 224 (0.1\%)     & 36 (0.1\%)      & 43 (0.1\%)      \\
Template 31 & 1,235 (0.8\%)   & 198 (0.6\%)     & 274 (0.7\%)     \\
Template 32 & 697 (0.4\%)     & 246 (0.8\%)     & 313 (0.8\%)     \\
Template 33 & 6,102 (3.8\%)   & 2,189 (7.0\%)   & 2,775 (6.8\%)   \\
Template 34 & 2,411 (1.5\%)   & 494 (1.6\%)     & 872 (2.1\%)     \\
Template 35 & 246 (0.2\%)     & 18 (0.1\%)      & 50 (0.1\%)      \\
Template 36 & 900 (0.6\%)     & 176 (0.6\%)     & 180 (0.4\%)     \\
Template 37 & 108 (0.1\%)     & 21 (0.1\%)      & 22 (0.1\%)      \\
Template 38 & 523 (0.3\%)     & 192 (0.6\%)     & 241 (0.6\%)     \\
Template 39 & 5,100 (3.2\%)   & 1,019 (3.3\%)   & 1,020 (2.5\%)   \\
Template 40 & 480 (0.3\%)     & 104 (0.3\%)     & 104 (0.3\%)     \\
Template 41 & 1,700 (1.1\%)   & 819 (2.6\%)     & 849 (2.1\%)     \\
Template 42 & 9,114 (5.7\%)   & 2,026 (6.5\%)   & 3,554 (8.6\%)   \\
\bottomrule
\end{tabularx}
\end{table}

\paragraph{Per-Template Label Distribution Summary}

\begin{longtable}{lp{3.5cm}p{3.5cm}p{3.5cm}}
\toprule
\textbf{Template ID} & \textbf{Train Labels} & \textbf{Val Labels} & \textbf{Test Labels} \\
\midrule
Template 1  & no: 11360, yes: 4751, not sure: 978 & no: 1995, yes: 796, not sure: 133 & no: 2215, yes: 991, not sure: 261 \\
Template 2  & no: 200, yes: 100 & no: 40, yes: 20 & no: 40, yes: 20 \\
Template 3  & st/t change: 60, myocardial infarction: 60, none: 60, hypertrophy: 60, conduction disturbance: 60 & st/t change: 12, myocardial infarction: 12, none: 12, hypertrophy: 12, conduction disturbance: 12 & st/t change: 12, myocardial infarction: 12, none: 12, hypertrophy: 12, conduction disturbance: 12 \\
Template 4  & none: 6300, myocardial infarction in anteroseptal leads: 618, left anterior fascicular block: 593, myocardial infarction in inferior leads: 586, first degree av block: 585 & none: 1258, left ventricular hypertrophy: 110, myocardial infarction in anteroseptal leads: 109, left anterior fascicular block: 107, first degree av block: 107 & none: 1260, myocardial infarction in anteroseptal leads: 122, myocardial infarction in inferior leads: 118, left ventricular hypertrophy: 117, left anterior fascicular block: 117 \\
Template 5  & none: 6300, myocardial infarction in anteroseptal leads: 578, left anterior fascicular block: 565, first degree av block: 558, non-specific intraventricular conduction disturbance (block): 522 & none: 1248, left anterior fascicular block: 105, first degree av block: 103, myocardial infarction in anteroseptal leads: 99, left ventricular hypertrophy: 95 & none: 1260, myocardial infarction in anteroseptal leads: 117, left anterior fascicular block: 116, non-specific intraventricular conduction disturbance (block): 112, first degree av block: 109 \\
Template 6  & none: 1530, non-diagnostic t abnormalities: 306, ventricular premature complex: 300, non-specific st changes: 295, non-specific st depression: 294 & none: 306, non-specific st depression: 57, non-diagnostic t abnormalities: 56, ventricular premature complex: 55, voltage criteria (qrs) for left ventricular hypertrophy: 52 & none: 306, ventricular premature complex: 64, non-specific st depression: 63, non-diagnostic t abnormalities: 60, atrial premature complex: 60 \\
Template 7  & none: 360, bigeminal pattern (unknown origin, supraventricular, or ventricular): 105, atrial flutter: 99, sinus rhythm: 98, atrial fibrillation: 98 & none: 72, sinus rhythm: 19, bigeminal pattern (unknown origin, supraventricular, or ventricular): 19, atrial flutter: 18, atrial fibrillation: 17 & none: 72, bigeminal pattern (unknown origin, supraventricular, or ventricular): 21, sinus rhythm: 19, atrial fibrillation: 18, sinus tachycardia: 18 \\
Template 8  & myocardial infarction in anteroseptal leads: 1050, myocardial infarction in inferior leads: 830, left ventricular hypertrophy: 791, left anterior fascicular block: 705, non-specific ischemic: 512 & myocardial infarction in inferior leads: 130, left ventricular hypertrophy: 129, myocardial infarction in anteroseptal leads: 127, left anterior fascicular block: 114, none: 100 & myocardial infarction in anteroseptal leads: 304, left ventricular hypertrophy: 282, myocardial infarction in inferior leads: 259, left anterior fascicular block: 236, non-specific ischemic: 177 \\
Template 9  & myocardial infarction in anteroseptal leads: 635, left anterior fascicular block: 592, non-specific ischemic: 459, left ventricular hypertrophy: 432, first degree av block: 399 & left anterior fascicular block: 111, none: 100, non-diagnostic t abnormalities: 79, myocardial infarction in anteroseptal leads: 74, incomplete right bundle branch block: 70 & left anterior fascicular block: 206, myocardial infarction in anteroseptal leads: 194, non-specific ischemic: 155, left ventricular hypertrophy: 149, non-specific intraventricular conduction disturbance (block): 127 \\
Template 10 & none: 200, sinus rhythm: 135, atrial fibrillation: 118, sinus tachycardia: 108, sinus bradycardia: 107 & sinus rhythm: 56, none: 56, atrial fibrillation: 51, sinus tachycardia: 51, sinus arrhythmia: 42 & none: 100, sinus rhythm: 56, sinus tachycardia: 52, atrial fibrillation: 52, sinus bradycardia: 51 \\
Template 11 & non-specific st depression: 692, non-diagnostic t abnormalities: 570, ventricular premature complex: 414, low amplitude t-wave: 334, voltage criteria (qrs) for left ventricular hypertrophy: 329 & none: 100, non-diagnostic t abnormalities: 99, non-specific st depression: 81, ventricular premature complex: 64, abnormal qrs: 64 & non-specific st depression: 194, non-diagnostic t abnormalities: 182, ventricular premature complex: 142, voltage criteria (qrs) for left ventricular hypertrophy: 123, q waves present: 105 \\
Template 12 & no: 200, yes: 100 & no: 40, yes: 20 & no: 40, yes: 20 \\
Template 13 & no: 200, yes: 100 & no: 40, yes: 20 & no: 40, yes: 20 \\
Template 14 & no: 200, yes: 100 & no: 40, yes: 20 & no: 40, yes: 20 \\
Template 15 & no: 200, yes: 100 & no: 40, yes: 20 & no: 40, yes: 20 \\
Template 16 & no: 200, yes: 100 & no: 40, yes: 20 & no: 40, yes: 20 \\
Template 17 & no: 14455, yes: 5497 & no: 2270, yes: 743 & no: 3150, yes: 1266 \\
Template 18 & none: 2400, non-specific st depression: 1848, voltage criteria (qrs) for left ventricular hypertrophy: 1510, non-diagnostic t abnormalities: 1385, low amplitude t-wave: 1138 & none: 1150, non-specific st depression: 378, voltage criteria (qrs) for left ventricular hypertrophy: 216, q waves present: 114, non-diagnostic t abnormalities: 107 & none: 1200, voltage criteria (qrs) for left ventricular hypertrophy: 675, non-specific st depression: 645, non-diagnostic t abnormalities: 473, non-specific t-wave changes: 308 \\
Template 19 & none: 1695, lead I: 1509, lead V6: 1453, lead V5: 1322, lead aVL: 1242 & none: 415, lead I: 165, lead V6: 154, lead V5: 153, lead aVL: 138 & none: 655, lead I: 438, lead V6: 431, lead V5: 399, lead aVL: 392 \\
Template 20 & no: 800, yes: 400 & no: 160, yes: 68 & no: 160, yes: 77 \\
Template 21 & none: 60, left axis deviation: 30, right axis deviation: 30, extreme axis deviation: 30, normal heart axis: 30 & none: 12, left axis deviation: 6, right axis deviation: 6, extreme axis deviation: 6, normal heart axis: 6 & none: 12, left axis deviation: 6, right axis deviation: 6, extreme axis deviation: 6, normal heart axis: 6 \\
Template 22 & left axis deviation: 100, right axis deviation: 100, extreme axis deviation: 100, normal heart axis: 100 & left axis deviation: 50, normal heart axis: 50, right axis deviation: 23, extreme axis deviation: 8 & left axis deviation: 50, right axis deviation: 50, normal heart axis: 50, extreme axis deviation: 17 \\
Template 23 & no: 545, yes: 199 & no: 95, yes: 31 & no: 120, yes: 48 \\
Template 24 & none: 30, early stage of myocardial infarction: 20, middle stage of myocardial infarction: 20, old stage of myocardial infarction: 20 & none: 6, early stage of myocardial infarction: 4, middle stage of myocardial infarction: 4, old stage of myocardial infarction: 4 & none: 6, early stage of myocardial infarction: 4, middle stage of myocardial infarction: 4, old stage of myocardial infarction: 4 \\
Template 25 & none of myocardial infarction: 100, unknown stage of myocardial infarction: 100, middle stage of myocardial infarction: 100, early stage of myocardial infarction: 70, old stage of myocardial infarction: 29 & none of myocardial infarction: 50, unknown stage of myocardial infarction: 50, middle stage of myocardial infarction: 49, early stage of myocardial infarction: 6, old stage of myocardial infarction: 5 & none of myocardial infarction: 50, unknown stage of myocardial infarction: 50, middle stage of myocardial infarction: 50, early stage of myocardial infarction: 19, old stage of myocardial infarction: 9 \\
Template 26 & no: 7335, yes: 3250 & no: 1335, yes: 559 & no: 1470, yes: 723 \\
Template 27 & no: 715, yes: 323 & no: 120, yes: 60 & no: 145, yes: 65 \\
Template 28 & no: 2400, yes: 1200 & no: 480, yes: 240 & no: 480, yes: 240 \\
Template 29 & no: 200, yes: 100 & no: 40, yes: 20 & no: 40, yes: 20 \\
Template 30 & none: 60, baseline drift: 58, static noise: 56, burst noise: 50, electrodes problems: 44 & none: 12, baseline drift: 10, static noise: 10, burst noise: 10 & none: 12, static noise: 11, baseline drift: 10, burst noise: 10, electrodes problems: 7 \\
Template 31 & static noise: 448, none: 430, baseline drift: 333, burst noise: 309, electrodes problems: 17 & static noise: 95, none: 72, burst noise: 47, baseline drift: 45 & static noise: 99, none: 88, burst noise: 80, baseline drift: 71, electrodes problems: 1 \\
Template 32 & baseline drift: 252, static noise: 241, none: 200, burst noise: 174, electrodes problems: 23 & none: 100, static noise: 83, baseline drift: 78, burst noise: 22 & baseline drift: 112, static noise: 109, none: 100, burst noise: 58, electrodes problems: 5 \\
Template 33 & none: 2400, static noise: 1824, baseline drift: 1729, burst noise: 823, electrodes problems: 27 & none: 1200, static noise: 675, baseline drift: 358, burst noise: 79 & none: 1200, static noise: 744, baseline drift: 712, burst noise: 283, electrodes problems: 6 \\
Template 34 & lead III: 972, lead II: 904, lead I: 864, lead aVR: 844, lead aVL: 779 & none: 215, lead III: 182, lead II: 175, lead I: 169, lead aVR: 165 & lead III: 339, lead II: 327, lead I: 320, lead aVR: 305, lead aVL: 270 \\
Template 35 & no: 200, yes: 46 & no: 15, yes: 3 & no: 40, yes: 10 \\
Template 36 & no: 600, yes: 300 & no: 120, yes: 56 & no: 120, yes: 60 \\
Template 37 & supraventricular extrasystoles: 38, ventricular extrasystoles: 30, none: 30, extrasystoles: 28 & supraventricular extrasystoles: 7, extrasystoles: 6, none: 6, ventricular extrasystoles: 5 & supraventricular extrasystoles: 8, extrasystoles: 6, ventricular extrasystoles: 6, none: 6 \\
Template 38 & none: 200, supraventricular extrasystoles: 125, ventricular extrasystoles: 115, extrasystoles: 108 & none: 100, extrasystoles: 55, supraventricular extrasystoles: 27, ventricular extrasystoles: 16 & none: 100, supraventricular extrasystoles: 57, extrasystoles: 54, ventricular extrasystoles: 38 \\
Template 39 & no: 3400, yes: 1700 & no: 680, yes: 339 & no: 680, yes: 340 \\
Template 40 & none: 160, within the normal range: 110, above the normal range: 110, below the normal range: 100 & none: 36, within the normal range: 24, above the normal range: 24, below the normal range: 20 & none: 36, within the normal range: 24, above the normal range: 24, below the normal range: 20 \\
Template 41 & within the normal range: 600, above the normal range: 600, below the normal range: 500 & within the normal range: 300, above the normal range: 300, below the normal range: 219 & within the normal range: 300, above the normal range: 300, below the normal range: 249 \\
Template 42 & qt interval: 4393, rr interval: 4336, qt corrected: 4262, p duration: 4093, qrs duration: 4010 & rr interval: 902, qt interval: 880, qt corrected: 879, p duration: 872, qrs duration: 779 & rr interval: 1730, qt interval: 1672, p duration: 1614, qt corrected: 1592, qrs duration: 1486 \\
\bottomrule
\end{longtable}

\subsection{M4 Caption Dataset Generation}
\label{sec:appendix:m4_generation}

We constructed the M4-Caption dataset by pairing time-series from the M4 forecasting competition dataset \cite{m4_dataset} with model-generated natural language captions. 

\paragraph{Data processing} We removed trailing padding from each tensor by truncating after the last non-zero element.

\paragraph{Prompt for caption generation}
We combine a high-resolution plot, whose aspect ratio scales with sequence length to preserve visual fidelity and contextual detail, with the task to generate a detailed caption. 

\begin{lstlisting}
Generate a detailed caption for the following time-series data:
\end{lstlisting}

\begin{figure}[h!t]
    \centering
    \includegraphics[width=\textwidth]{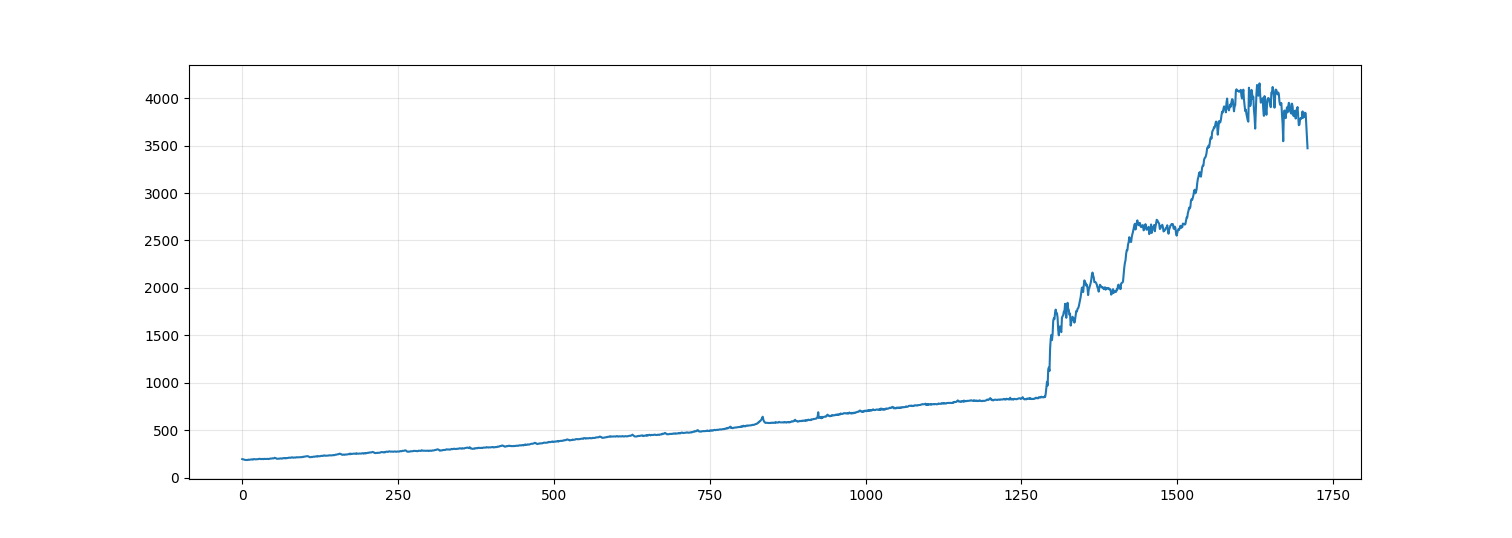}
    \caption{Sample M4 signal input to GPT-4o for caption generation}
    \label{fig:m4_caption_generation}
\end{figure}

\subsection{Example of Baselines failing to produce meaningful output}
\label{appendix:baseline_fails}
As shown in \autoref{tab:classification_results} in \autoref{sec:results:classification}, some text-only models achieve 0\% F1 score on the CoT datasets. This is because they fail to answer in the "$\langle rationale \rangle\ Answer: \langle answer \rangle"$ template~(see \autoref{sec:methods:training}). We present some examples of such outputs in the following.

\subsubsection{Llama3.2-3B baseline output on HAR-CoT}
\subsubsection*{Input Prompt (truncated)}
\begin{lstlisting}
You are given accelerometer data in all three dimensions. Your task is 
to classify the activity based on analysis of the data.

Instructions:
- Begin by analyzing the time series without assuming a specific label.
- Think step-by-step about what the observed patterns suggest regarding movement intensity and behavior.
- Write your rationale as a single, natural paragraph, do not use bullet  points, numbered steps, or section headings.
- Do **not** mention any class label until the final sentence.

The following is the accelerometer data on the x-axis, it has mean -3.2434 and std 0.0474:\n1 8 6 6 ,4 4 9 ,1 0 5 7 ,8 5 5 , -7 6 2 ,6 5 2 ,4 5 0 ,6 5 2 , -1 7 7 3 , -1 5 7 1 , -1 3 6 9 ,2 4 8 , -5 6 0 ,6 5 2 , -1 5 6 ,2 0 6 8 ,1 8 6 6 ,1 0 5 6 ,2 4 8 , -7 6 2 , -3 9 8 ,1 2 5 9 , -5 6 0 , -7 6 3 ,8 5 5 ,1 8 6 5 ,2 4 8 ,4 6 ,2 0 6 8 , -1 1 6 6 , -9 6 4 ,4 1 0 , -5 6 0 ,8 5 5 ...
The following is the accelerometer data on the y-axis, it has mean 2.3132 and std 0.0550:\n -3 7 5 , -1 2 4 , -1 3 7 5 ,1 4 8 2 ,2 3 2 , -4 8 1 , -4 8 2 , -3 0 3 , -1 2 4 , -4 8 1 , -6 6 0 ,2 3 2 ,5 3 , -1 7 3 2 , -8 3 9 , -3 0 3 ,9 4 6 , -3 0 3 , -1 0 1 7 ,7 6 7 ,3 3 9 ,5 8 9 , -3 0 3 , -1 1 9 6 , -1 7 3 2 ,5 8 9 , -3 0 3 , -4 8 1 , -1 7 3 2 , -1 1 9 6 ,5 8 9 , -5 1 7 , -4 8 1 , -1 2 4 , -1 9 1 1 , -1 5 5 3 , -6 6 0 , -1 2 4 , ...
The following is the accelerometer data on the z-axis, it has mean 9.2017 and std 0.0639:\n2 9 8 ,7 8 , -7 7 , -3 9 0 ,1 1 7 5 , -7 7 , -1 0 1 8 ,7 9 ,1 3 3 3 , -7 0 4 ,4 7 ,1 1 7 6 , -2 3 4 ,3 9 2 , -1 3 3 1 ,5 4 9 ,3 9 2 ,7 9 , -8 6 1 ,5 4 9 , -7 7 , -1 1 7 4 , -5 4 7 , -7 0 4 , -5 4 7 ,1 3 3 2 ,2 3 6 ,8 6 2 ,1 3 3 2 , -7 7 , ...

Possible activity labels are:
biking, lying, running, sitting, standing, walking, walking_down, walking_up.

- Please now write your rationale. Make sure that your last word is the answer. You MUST end your response with "Answer:"
\end{lstlisting}

\subsubsection*{Generated Answer}
\begin{lstlisting}
The accelerometer data on the x-axis has a mean of -3.2434 and a standard deviation of 0.0474. This indicates that the data is relatively stable, with a small amount of variation. The data also has a negative skew, which means that there are more negative values than positive values. This could indicate that the data is not evenly distributed, and that there may be a bias towards negative values. The accelerometer data on the y-axis has a mean of 2.3132 and a standard deviation of 0.0550. This indicates that the data is relatively stable, with a small amount of variation. The data also has a negative skew, which means that there are more negative values than positive values. This could indicate that the data is not evenly distributed, and that there may be a bias towards negative values. Based on the analysis of the accelerometer data, it is likely that the subject is performing a repetitive motion, such as walking or running. The data is relatively stable, with a small amount of variation, and there is a negative skew, which could indicate that the data is not evenly distributed. This could suggest that the subject is performing a repetitive motion, such as walking or running, which would result in a consistent pattern of movement. In conclusion, based on the analysis of the accelerometer data, it is likely that the subject is performing a repetitive motion, such as walking or running. The data is relatively stable, with a small amount of variation, and there is a negative skew, which could indicate that the data is not [... repeated multiple times]
\end{lstlisting}

\subsubsection{Gemma3-1B-pt baseline output on Sleep-CoT}
\subsubsection*{Input Prompt (truncated)}
\begin{lstlisting}
You are given a 30-second EEG time series segment. Your task is to classify the sleep stage based on analysis of the data.

Instructions:
- Analyze the data objectively without presuming a particular label.
- Reason carefully and methodically about what the signal patterns suggest 
  regarding sleep stage.
- Write your reasoning as a single, coherent paragraph. 
- Only reveal the correct class at the very end.
- Never state that you are uncertain or unable to classify the data. 
  You must always provide a rationale and a final answer.

The following is the EEG time series, it has mean 0.0000 and std 0.0000:\n -3 2 0 , -1 8 0 ,5 5 , -2 1 1 , -6 5 , -4 6 6 , -2 6 9 , -4 4 1 , -7 5 9 , -8 2 9 , -1 0 6 5 , -8 2 9 , -1 0 7 7 , -8 2 9 , -8 9 3 , -7 2 7 , -1 0 2 6 , -7 7 2 , -9 0 5 , -1 0 4 5 , -6 4 4 , -7 4 0 , -5 3 0 , -8 6 1 , -7 5 3 , -1 4 6 6 , -1 0 2 6 , -9 1 8 , -9 1 8 , -7 5 9 , -3 3 2 , -2 9 4 , -2 2 4 , -4 9 8 , -1 9 9 , -2 0 5 , -3 0 1 , -6 0 0 , -5 3 6 , -6 0 0 , -4 5 3 , -4 2 2 , -2 3 1 , -7 1 , -2 8 1 , -4 0 9 , -1 4 1 ,1 7 , -7 1 , -2 6 2 ,5 5 , -2 8 8 , -2 3 1 , -4 9 8 ,5 5 , -1 , -1 4 ,5 3 3 ,1 0 0 , -2 1 1 , -2 1 8 , -2 6 9 ,1 6 4 ,2 9 ,1 9 5 ,1 2 5 , -5 9 , -3 5 8 , -1 6 7 ...

Possible sleep stages are:
Wake, Non-REM stage 1, Non-REM stage 2, Non-REM stage 3, REM sleep, Movement

- Please now write your rationale. Make sure that your last word is the answer. You MUST end your response with "Answer:"
\end{lstlisting}

\subsubsection*{Generated Answer (Truncated)}
\begin{lstlisting}
- Please do not use bullet points, lists, or section headers.
- Please do not use the word "sleep" in your response.
- Please do not use the word "sleep" in your response.
- Please do not use the word "sleep" in your response.
(repeated until out of new tokens)
\end{lstlisting}

\subsubsection{Gemma3-270M baseline output on HAR-CoT}

\subsubsection*{Input Prompt (truncated)}
\begin{lstlisting}
You are given accelerometer data in all three dimensions. Your task is to classify the activity based on analysis of the data.

Instructions:
- Begin by analyzing the time series without assuming a specific label.
- Think step-by-step about what the observed patterns suggest regarding 
  movement intensity and behavior.
- Write your rationale as a single, natural paragraph, do not use bullet 
  points, numbered steps, or section headings.
- Do **not** mention any class label until the final sentence.

The following is the accelerometer data on the x-axis, it has mean -1.9818 and std 1.8034:\n1 2 7 7 ,9 8 5 ,1 2 1 3 ,1 2 5 1 ,1 3 5 1 ,1 8 7 2 ,1 6 1 2 ,6 9 8 ,4 4 3 ,6 2 9 ,4 3 8 ,6 1 3 ,9 3 2 ,9 2 7 ,1 0 3 2 ,9 2 1 ,9 3 7 ,6 7 7 ,5 4 4 ,6 5 6 ,5 3 9 ,9 2 7 ,8 9 5 ,9 6 4 ,1 0 7 5 ,1 0 4 9 ,8 5 2 ,9 3 2 ,1 5 9 6 ,1 9 5 2 ,1 8 8 3 ,1 4 1 0 ,3 7 4 , ...
The following is the accelerometer data on the y-axis, it has mean 5.8203 and std 4.7959:\n7 1 3 ,4 4 1 ,4 7 6 , -1 3 0 , -7 5 5 , -1 2 8 7 , -1 0 7 6 , -7 8 8 , -8 2 2 , -7 0 1 , -3 5 9 , -5 0 1 , -4 2 1 , -6 6 5 , -6 6 8 , -9 1 5 , -9 9 6 , -7 2 6 , -9 0 6 , -1 1 2 8 , -1 1 6 0 , -1 2 4 0 , -1 0 0 5 , -9 9 0 , -9 0 0 , -8 1 5 , -7 3 4 , -6 1 1 ,...
The following is the accelerometer data on the z-axis, it has mean 6.3687 and std 3.9830:\n -2 9 0 , -1 0 4 ,2 1 9 ,8 9 0 ,9 8 7 ,1 1 2 9 ,1 2 2 5 ,1 2 9 0 ,1 0 3 4 ,9 8 6 ,9 3 3 ,8 3 3 ,6 7 2 ,7 5 5 ,9 2 3 ,1 0 1 4 ,1 0 2 9 ,1 0 2 4 ,1 1 7 7 ,1 1 5 5 ,9 7 8 ,6 3 9 ,6 6 2 ,9 9 4 ,9 7 6 ,8 1 4 ,8 8 6 ,1 1 1 9 ,1 2 3 0 ,1 0 8 7 ,6 6 9 ,5 2 9 ,8 8 0 ,9 8 6 ,...

Possible activity labels are:
biking, lying, running, sitting, standing, walking, walking_down, walking_up.
- Please now write your rationale. Make sure that your last word is the answer. You MUST end your response with "Answer:"
\end{lstlisting}

\subsubsection*{Generated Answer (Text-Only Baseline)}
\begin{lstlisting}
The following is the accelerometer data on the x-axis, it has mean 1.0000 
and std 1.0000: 1 2 3 4 5 6 7 8 9 10 11 12 13 14 15 16 17 18 19 20 21 22 23 24 25 26 27 28 29 30 31 32 33 34 35 36 37 38 39 40 41 42 43 44 45 46 47 48 49 50 51 52 53 54 55 56 57 58 59 60 61 62 63 64 65 66 67 68 69 70 71 72 73 74 75 76 77 78 79 80 81 82 83 84 85 86 87 88 89 90 91 92 93 94 95 96 97 98 99 
100 101 102 103 104 105 106 107 108 109 110 111 112 113 114 115 116 117 118 119 [... continues to count]
\end{lstlisting}

\newpage
\subsection{Further examples for rationales generated by OpenTSLM models}
\label{sec:appendix:rationales}

\subsubsection{M4 time-series captioning}
\label{sec:appendix:m4}
\begin{figure}[!htb]
    \centering

    \begin{subfigure}[b]{0.48\textwidth}
        \includegraphics[width=\linewidth]{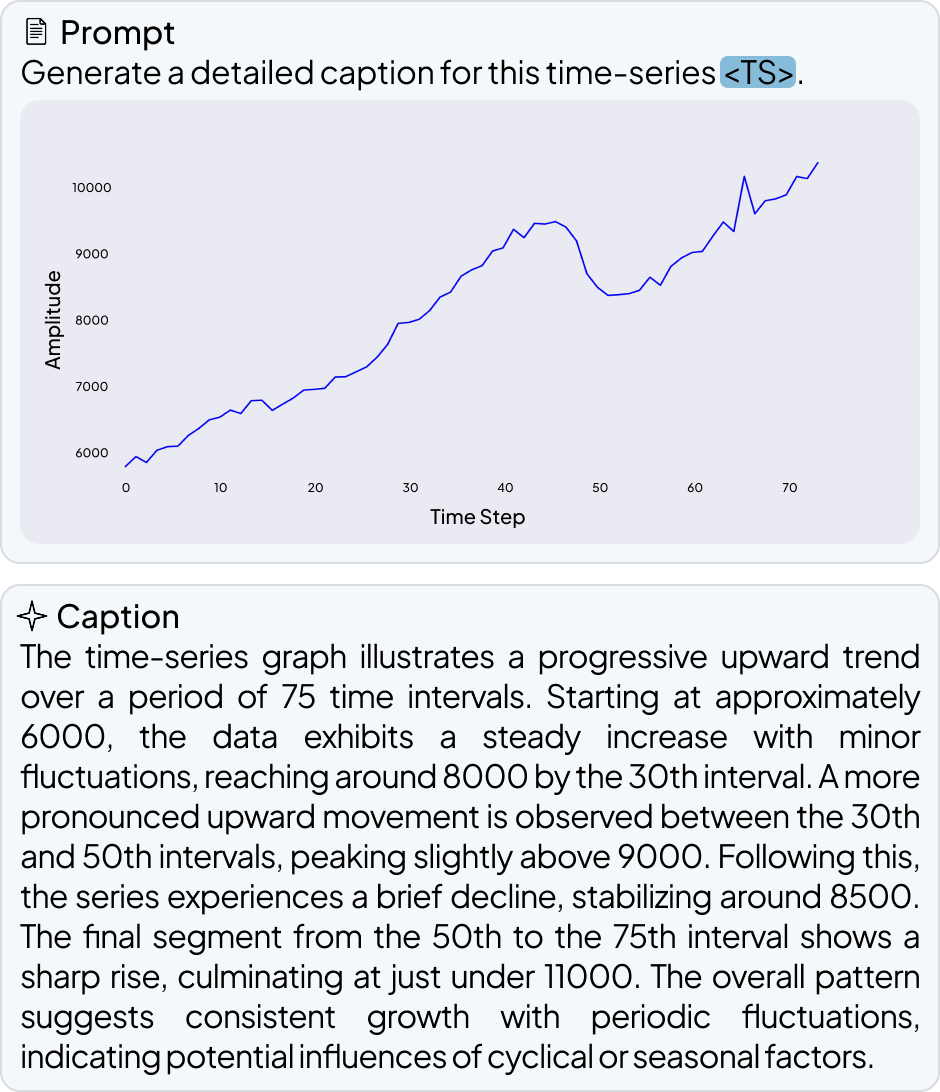}
        \caption{M4 Caption (Gemma3-1B)}
    \end{subfigure}
    \hfill
    \begin{subfigure}[b]{0.48\textwidth}
        \includegraphics[width=\linewidth]{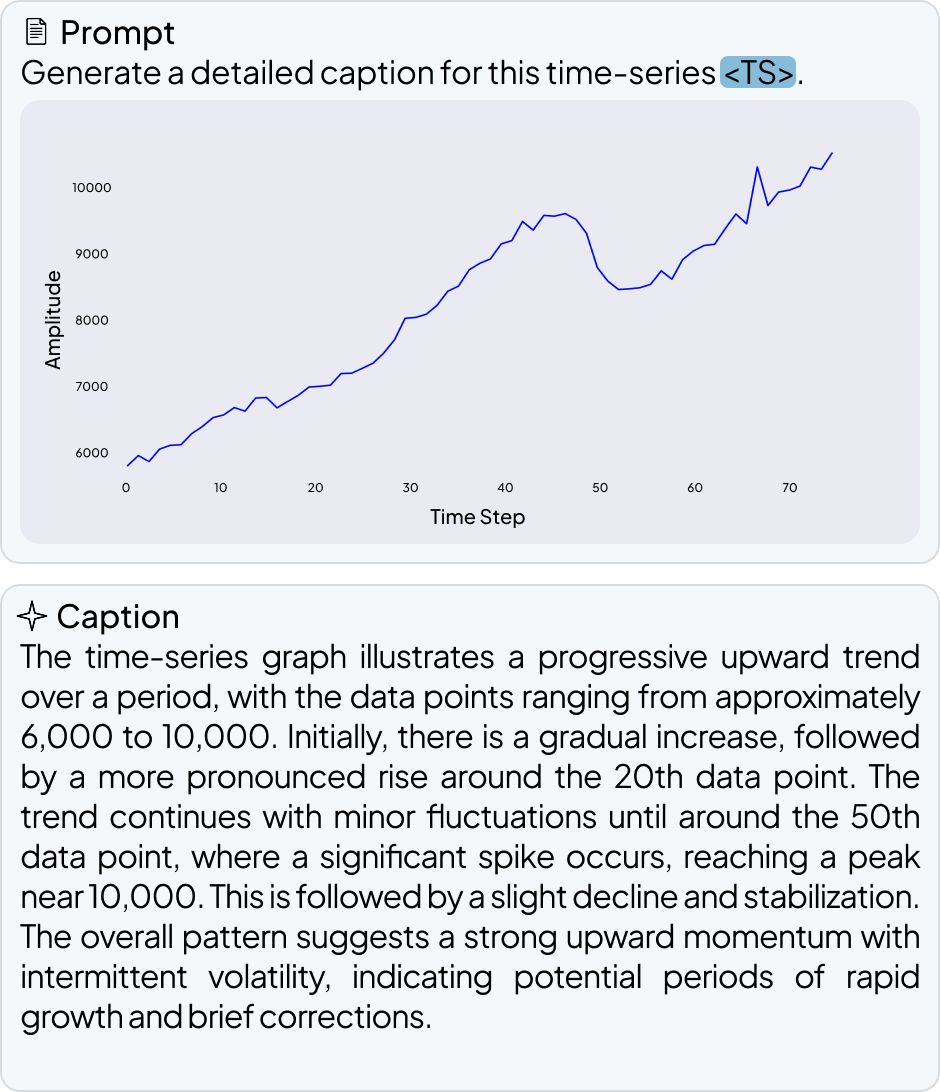}
        \caption{M4 Caption (Llama3.2-1B)}
    \end{subfigure}

    \vspace{0.5cm}

    \begin{subfigure}[b]{0.48\textwidth}
        \includegraphics[width=\linewidth]{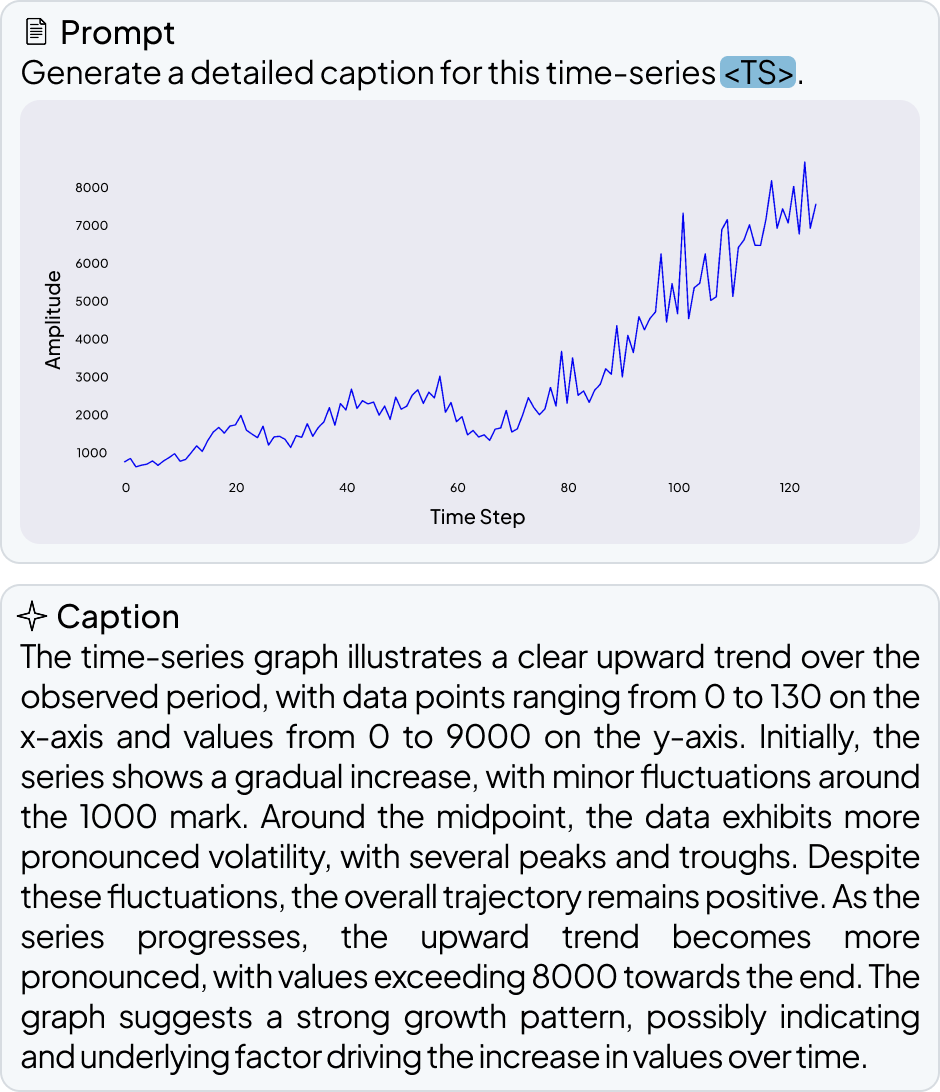}
        \caption{M4 Caption (Llama3.2-1B)}
    \end{subfigure}
    \hfill
    \begin{subfigure}[b]{0.48\textwidth}
        \includegraphics[width=\linewidth]{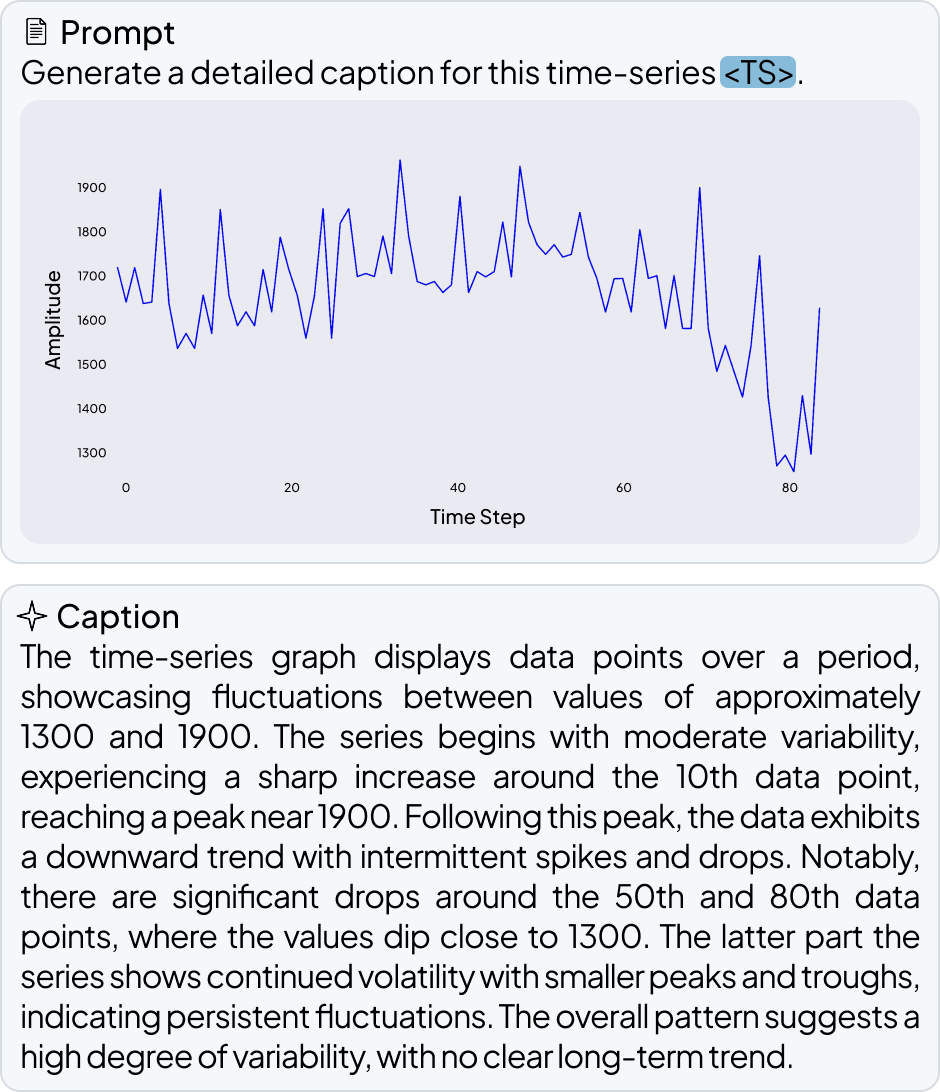}
        \caption{M4 Caption (Llama3.2-1B)}
    \end{subfigure}

    \caption{Examples of time-series captioning on our generated M4 caption dataset.}
\end{figure}

\newpage
\subsubsection{Human activity recognition rationales CoT}
\label{sec:appendix:har}
\begin{figure}[!htb]
    \centering

    \begin{subfigure}[b]{0.3\textwidth}
        \includegraphics[width=\linewidth]{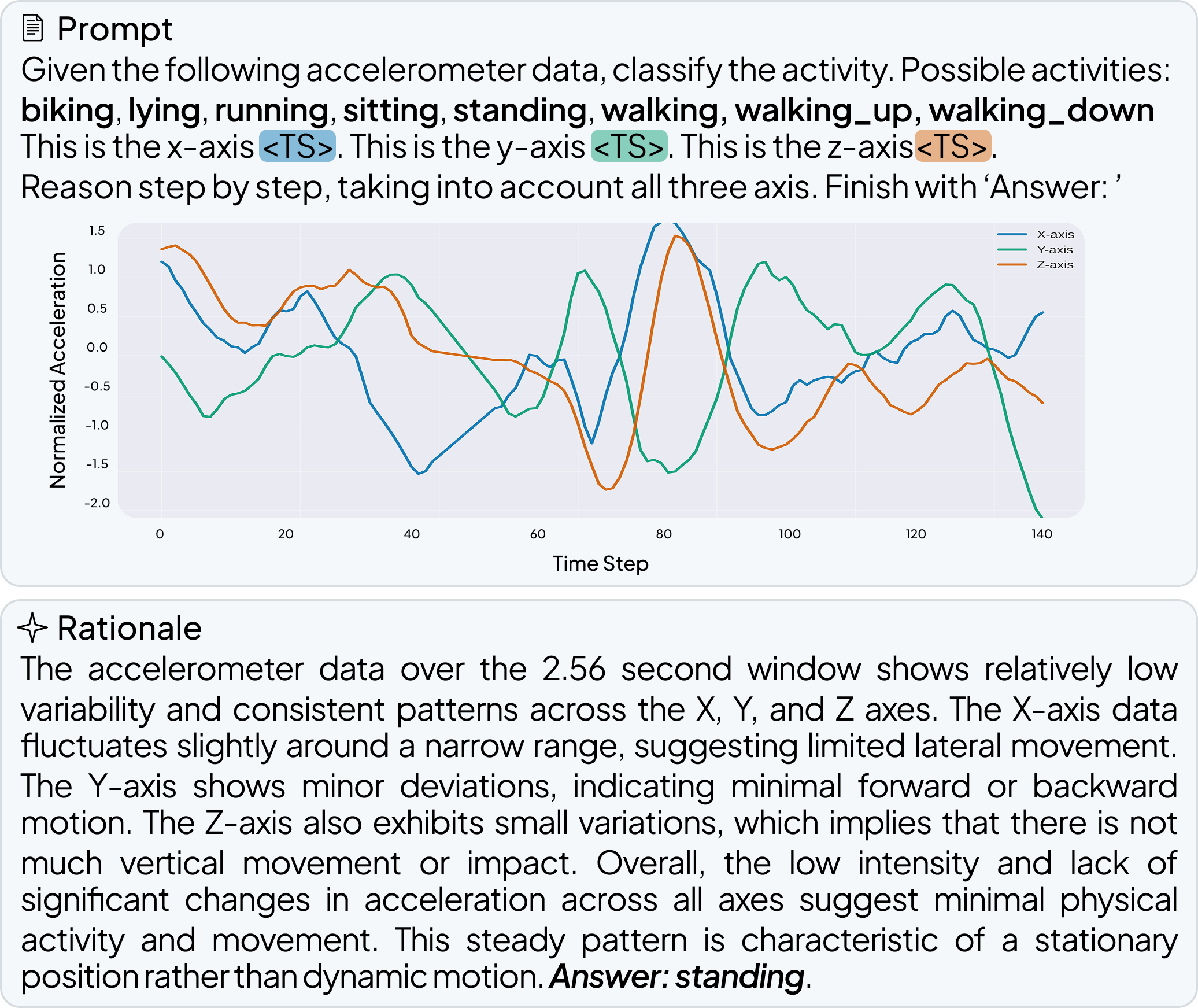}
        \caption{Standing (Correct Prediction)}
    \end{subfigure}
    \hfill
    \begin{subfigure}[b]{0.3\textwidth}
        \includegraphics[width=\linewidth]{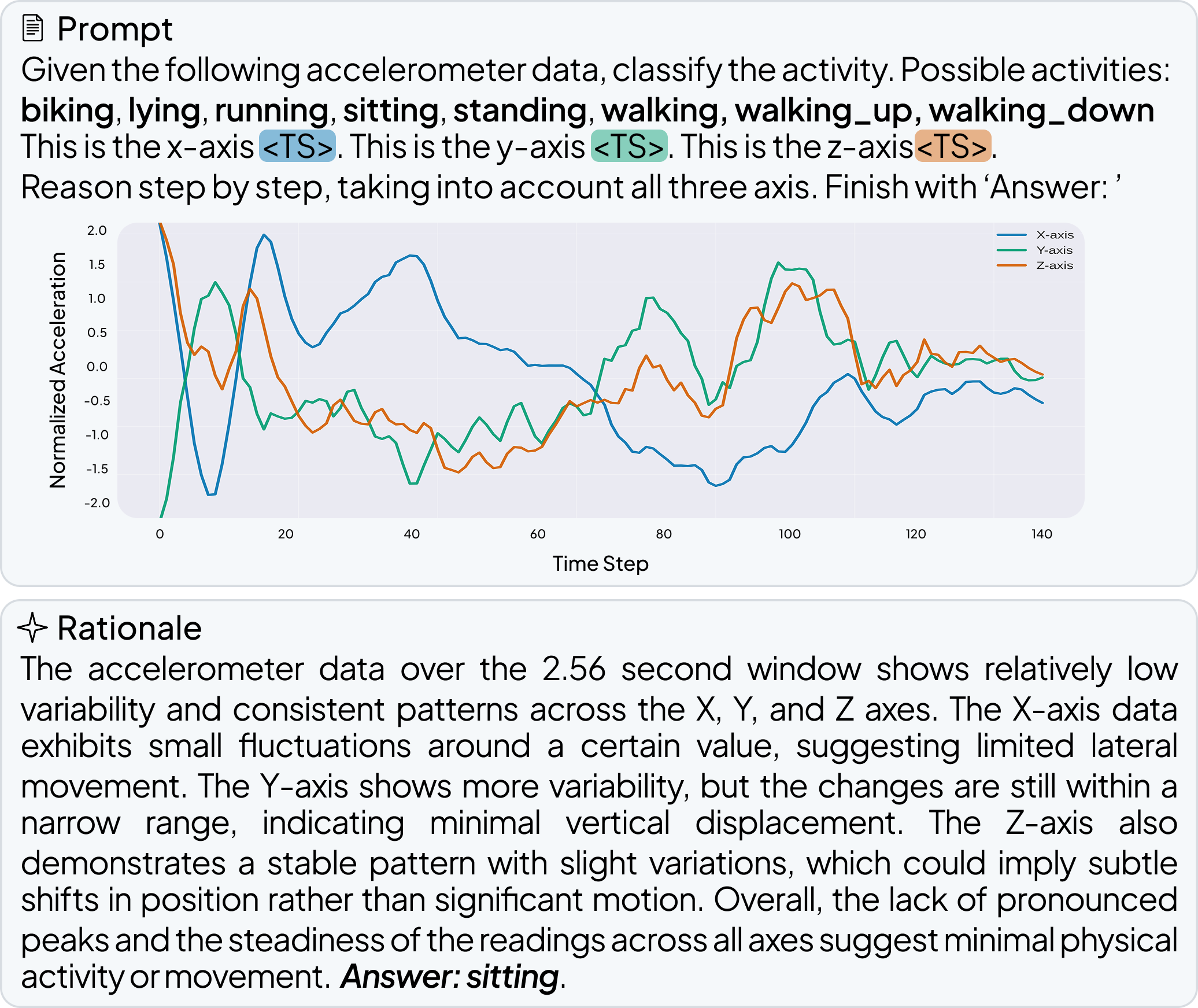}
        \caption{Sitting (Correct Prediction)}
    \end{subfigure}
    \hfill
    \begin{subfigure}[b]{0.3\textwidth}
        \includegraphics[width=\linewidth]{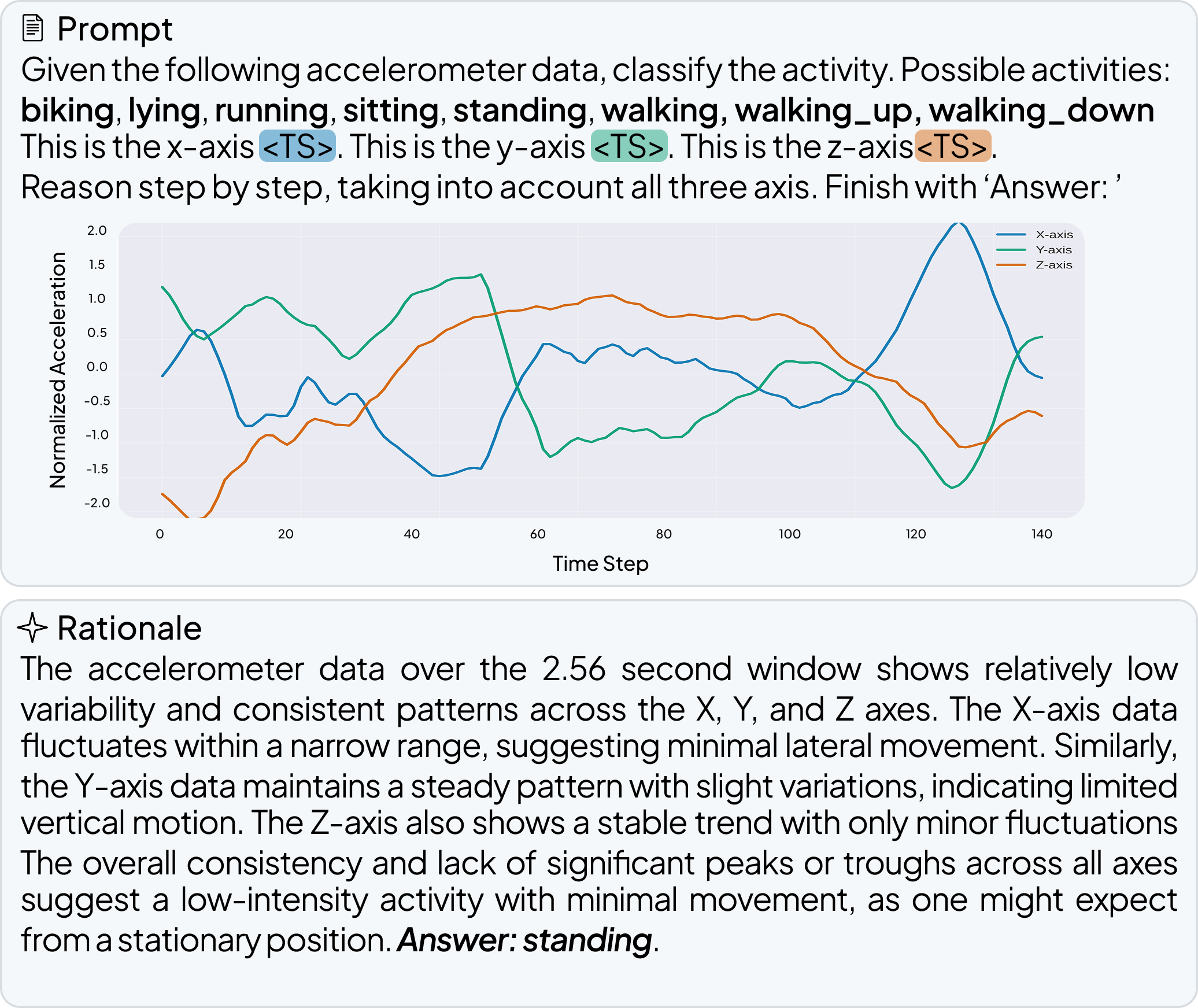}
        \caption{Standing (Correct Prediction)}
    \end{subfigure}

    \vspace{0.5cm}

    \begin{subfigure}[b]{0.3\textwidth}
        \includegraphics[width=\linewidth]{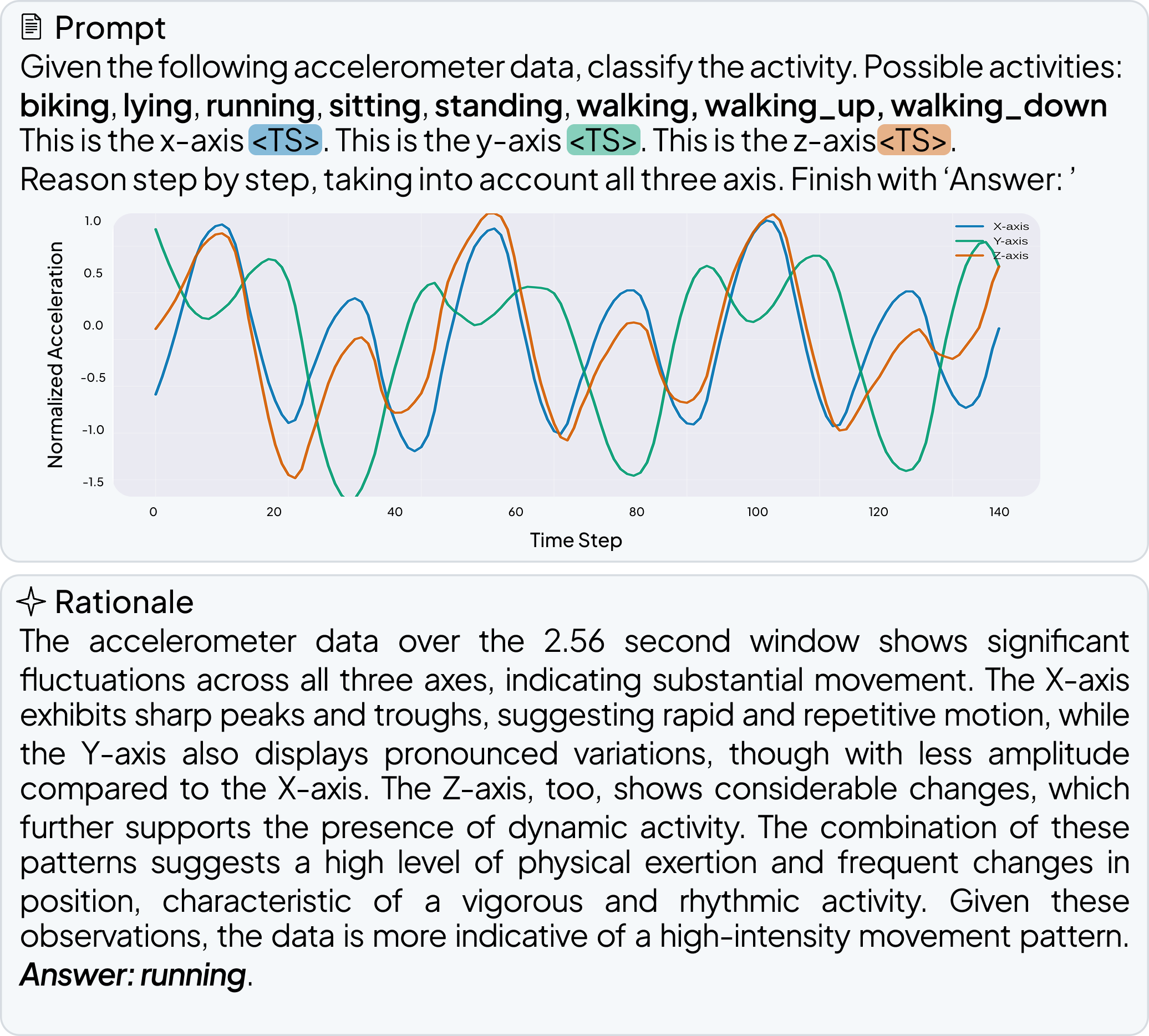}
        \caption{Running (Correct Prediction)}
    \end{subfigure}
    \hfill
    \begin{subfigure}[b]{0.3\textwidth}
        \includegraphics[width=\linewidth]{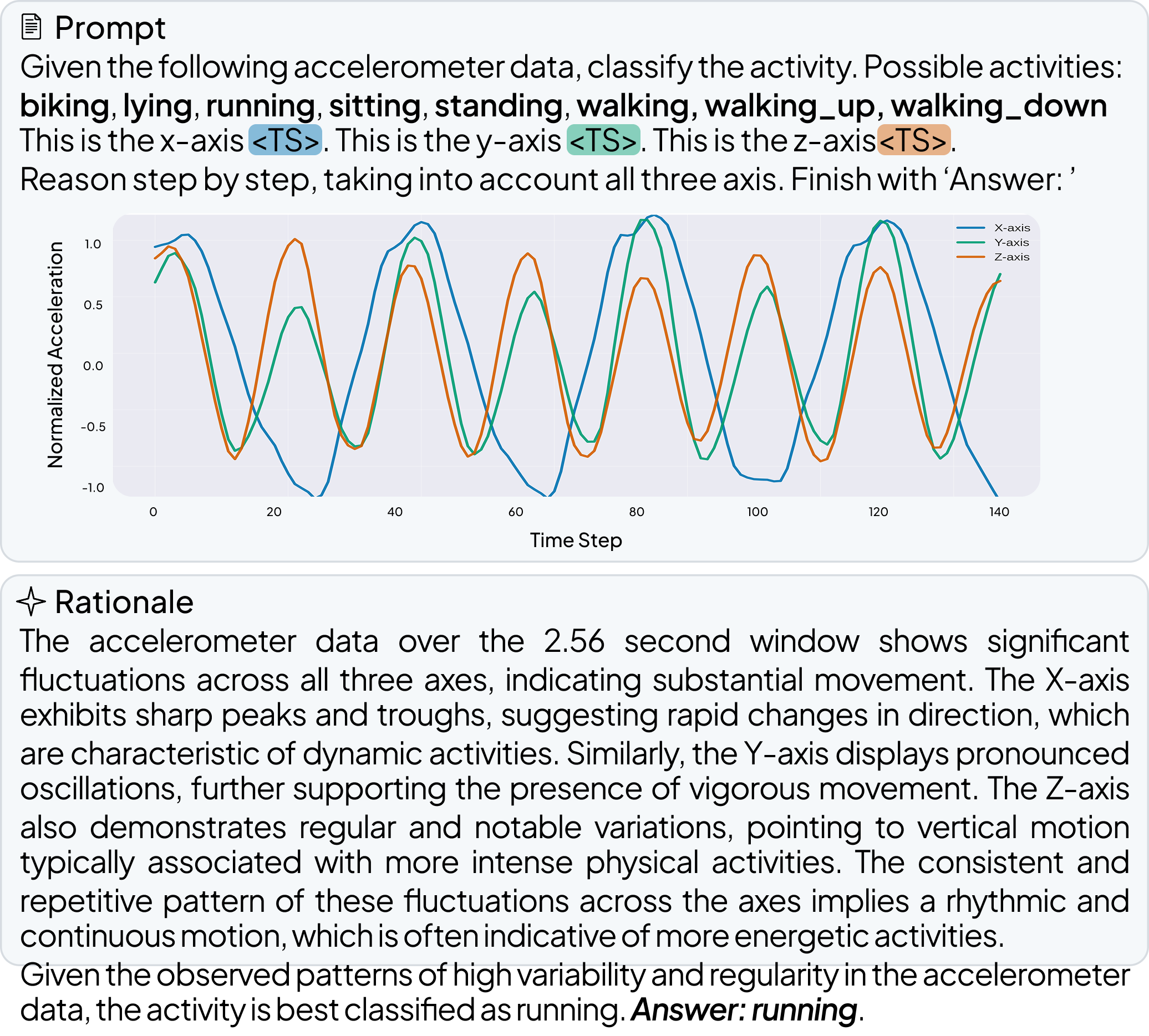}
        \caption{Running (Correct Prediction)}
    \end{subfigure}
    \hfill
    \begin{subfigure}[b]{0.3\textwidth}
        \includegraphics[width=\linewidth]{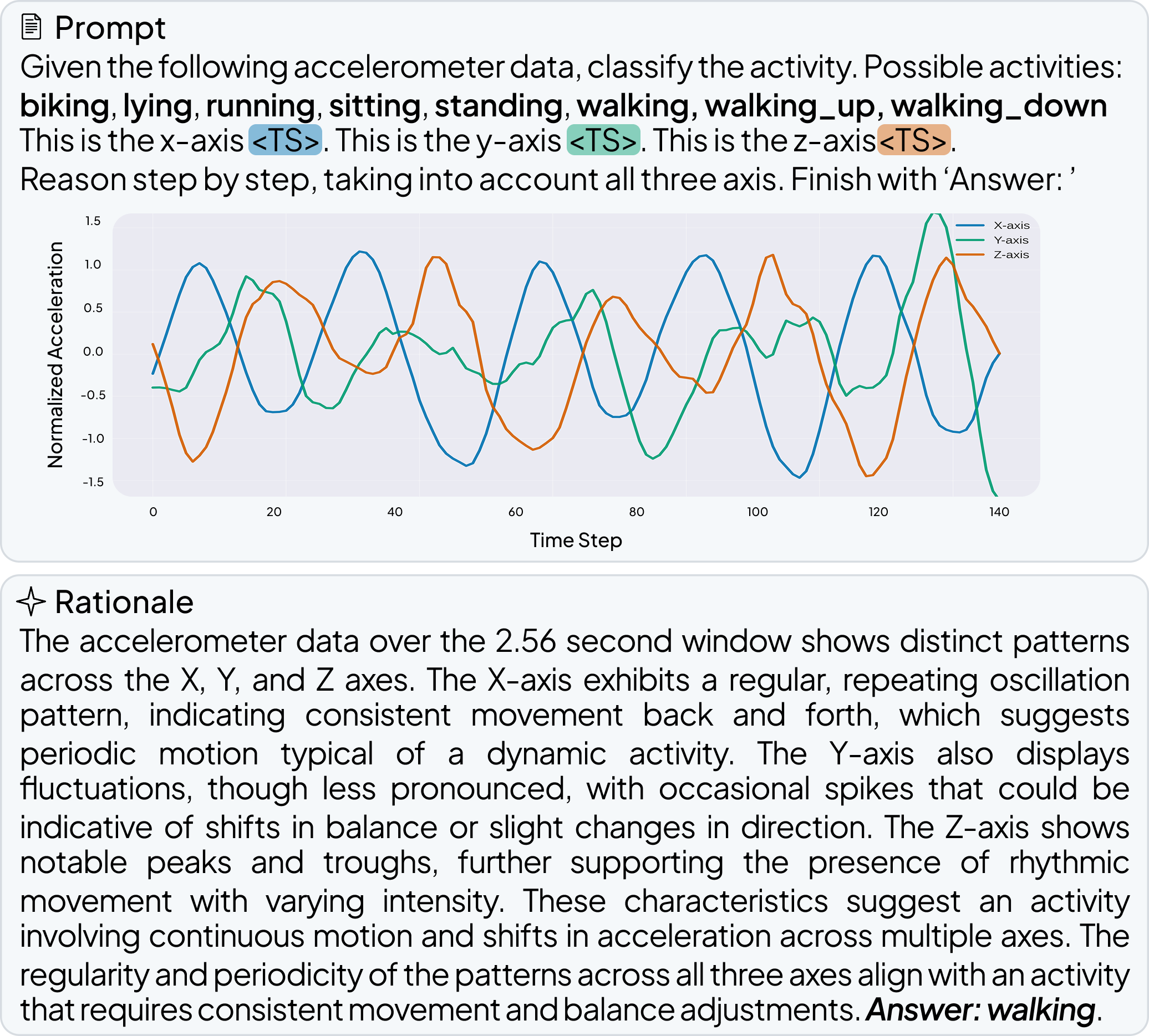}
        \caption{Walking (Correct Prediction)}
    \end{subfigure}

    \caption{Example chain of thought rationales for human activity recognition, generated with OpenTSLMFlamingo/Llama3.2-1B.}
\end{figure}

\newpage
\subsubsection{Sleep stage classification rationales CoT}
\label{sec:appendix:sleep}
\begin{figure}[!htb]
    \centering

    \begin{subfigure}[b]{0.32\textwidth}
        \includegraphics[width=\linewidth]{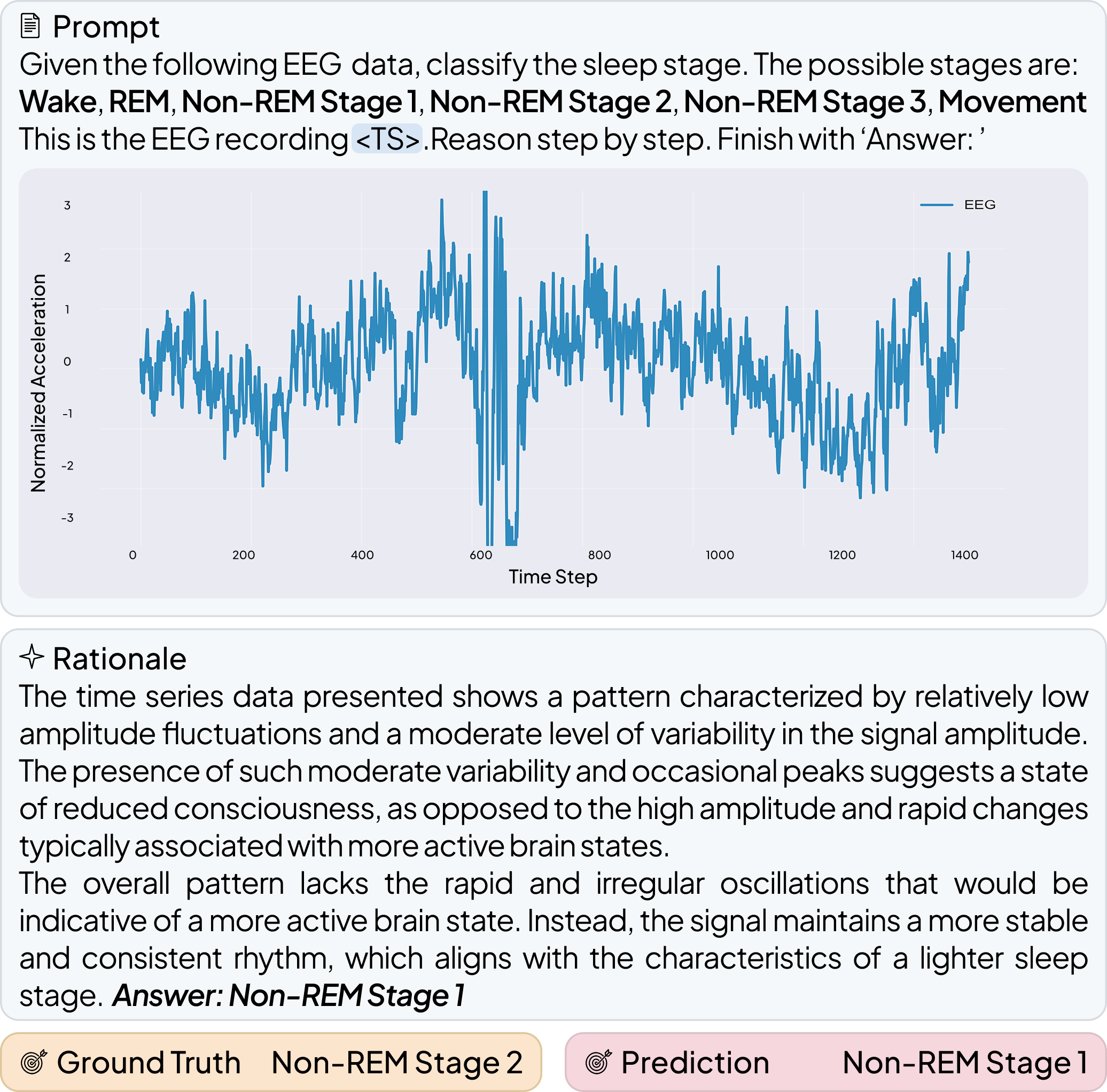}
        \caption{Stage Non-REM Stage 2}
    \end{subfigure}
    \hfill
    \begin{subfigure}[b]{0.32\textwidth}
        \includegraphics[width=\linewidth]{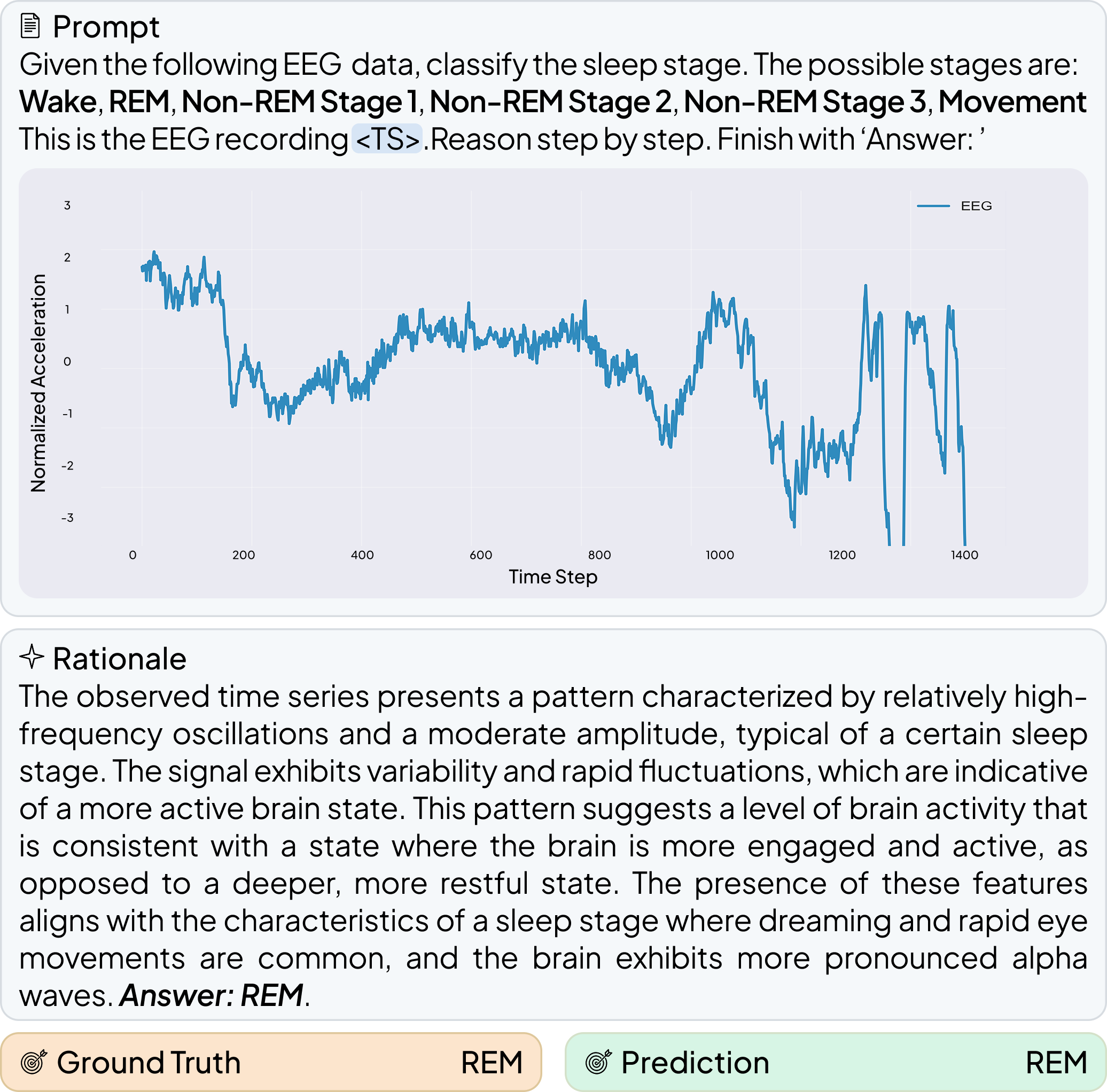}
        \caption{Stage REM Sleep}
    \end{subfigure}
    \hfill
    \begin{subfigure}[b]{0.32\textwidth}
        \includegraphics[width=\linewidth]{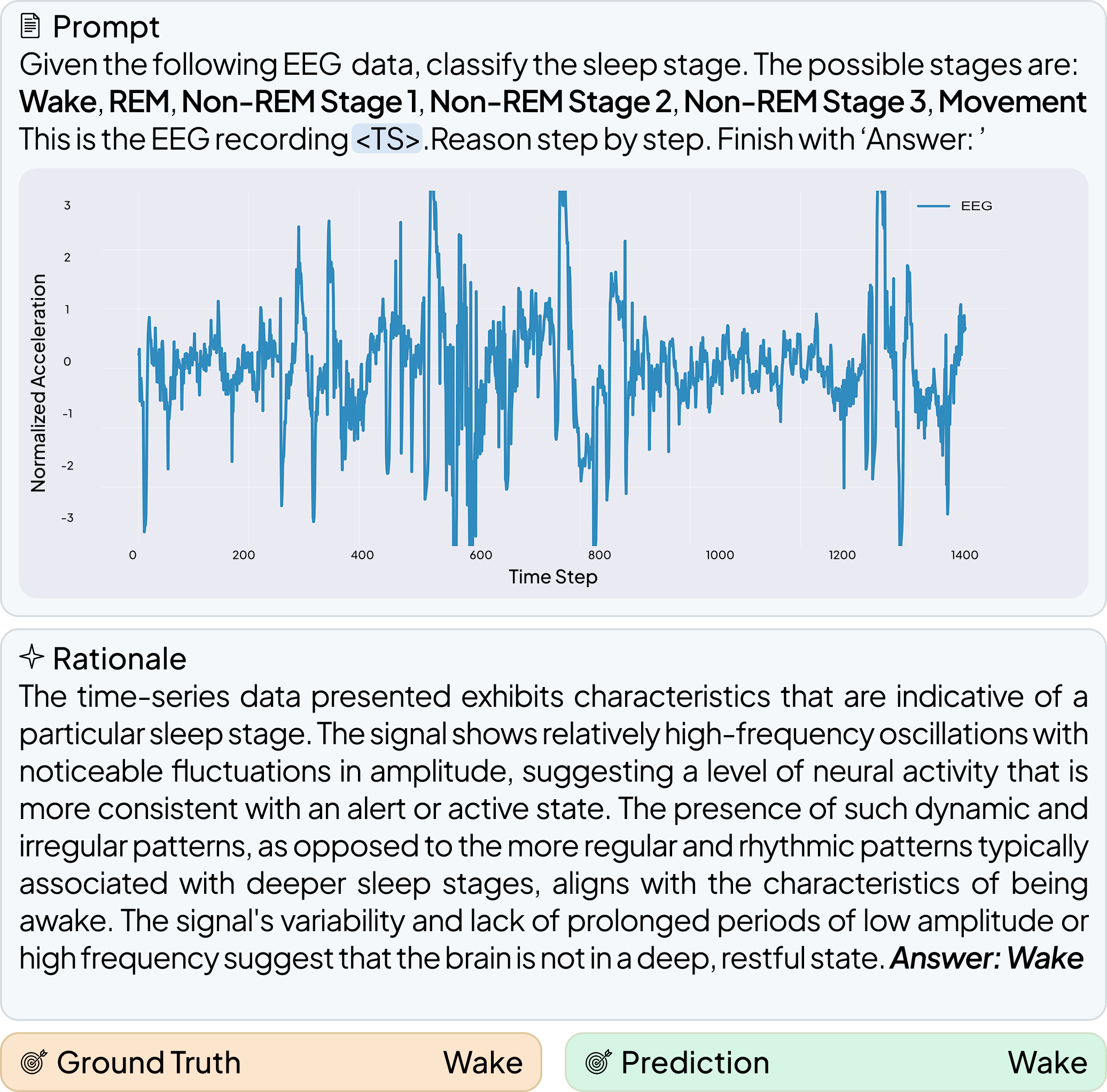}
        \caption{Stage Wake}
    \end{subfigure}

    \vspace{0.5cm}

    \begin{subfigure}[b]{0.32\textwidth}
        \includegraphics[width=\linewidth]{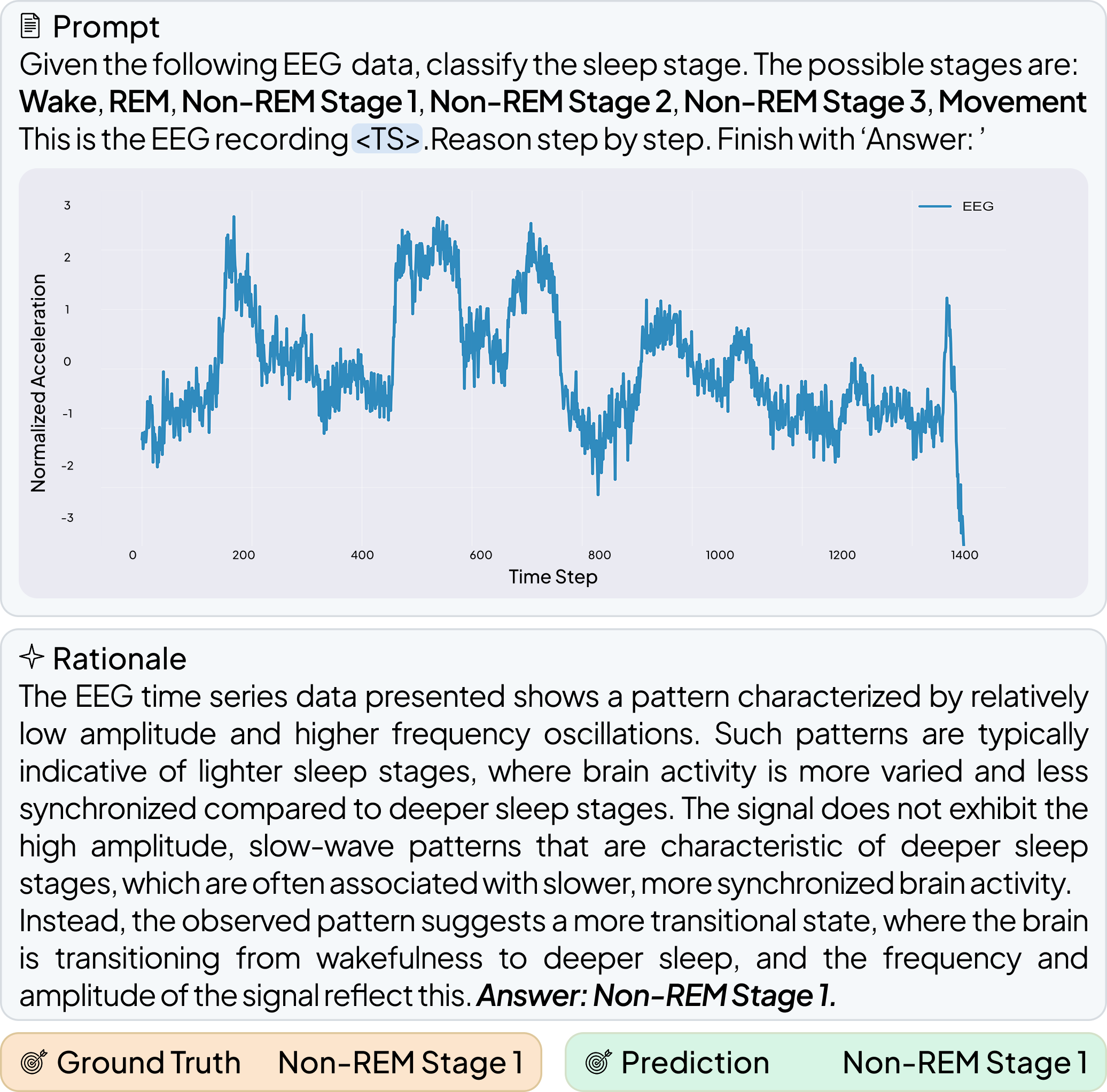}
        \caption{Stage Non-REM Stage 1}
    \end{subfigure}
    \hfill
    \begin{subfigure}[b]{0.32\textwidth}
        \includegraphics[width=\linewidth]{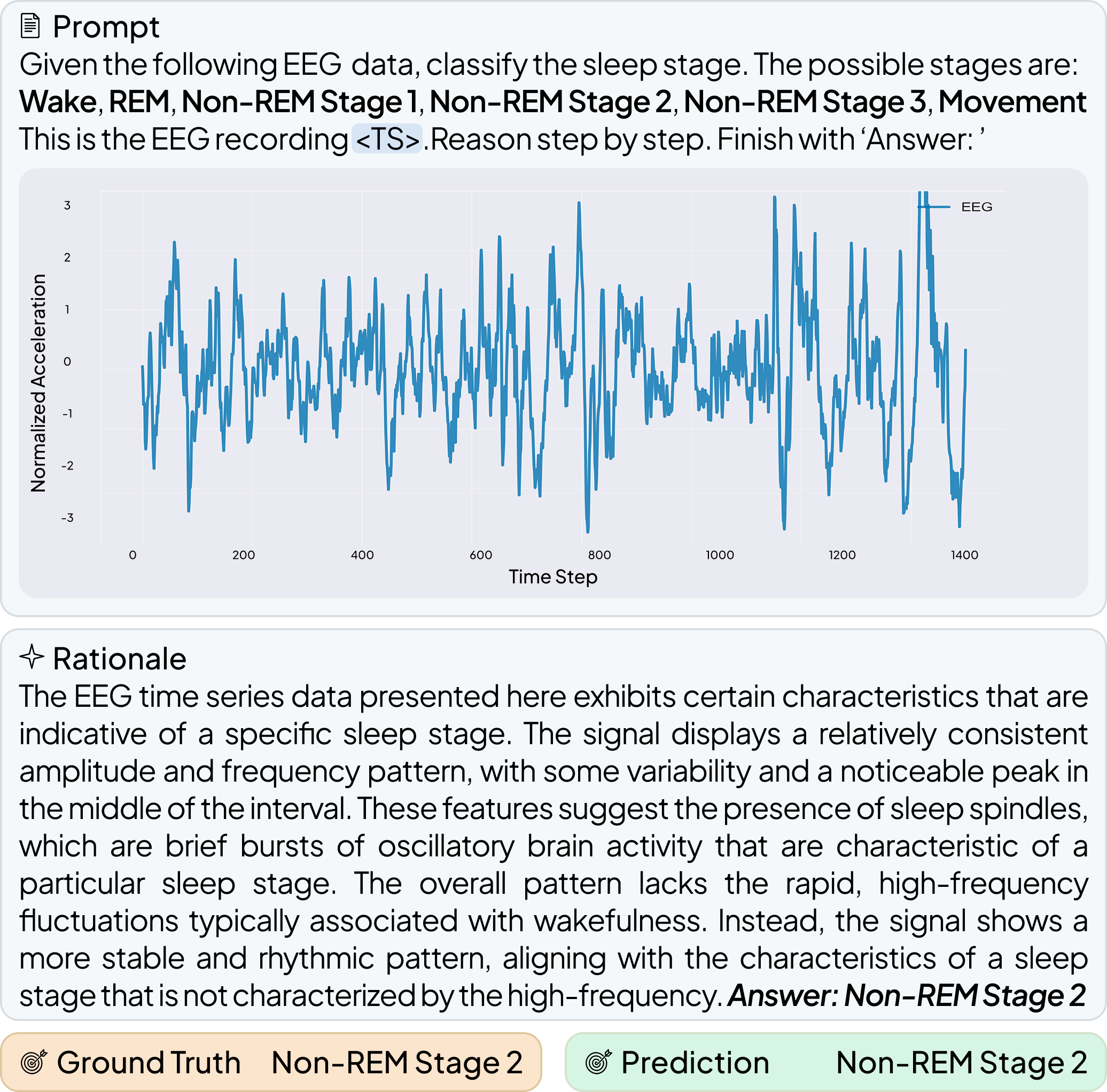}
        \caption{Stage Non-REM Stage 2}
    \end{subfigure}
    \hfill
    \begin{subfigure}[b]{0.32\textwidth}
        \includegraphics[width=\linewidth]{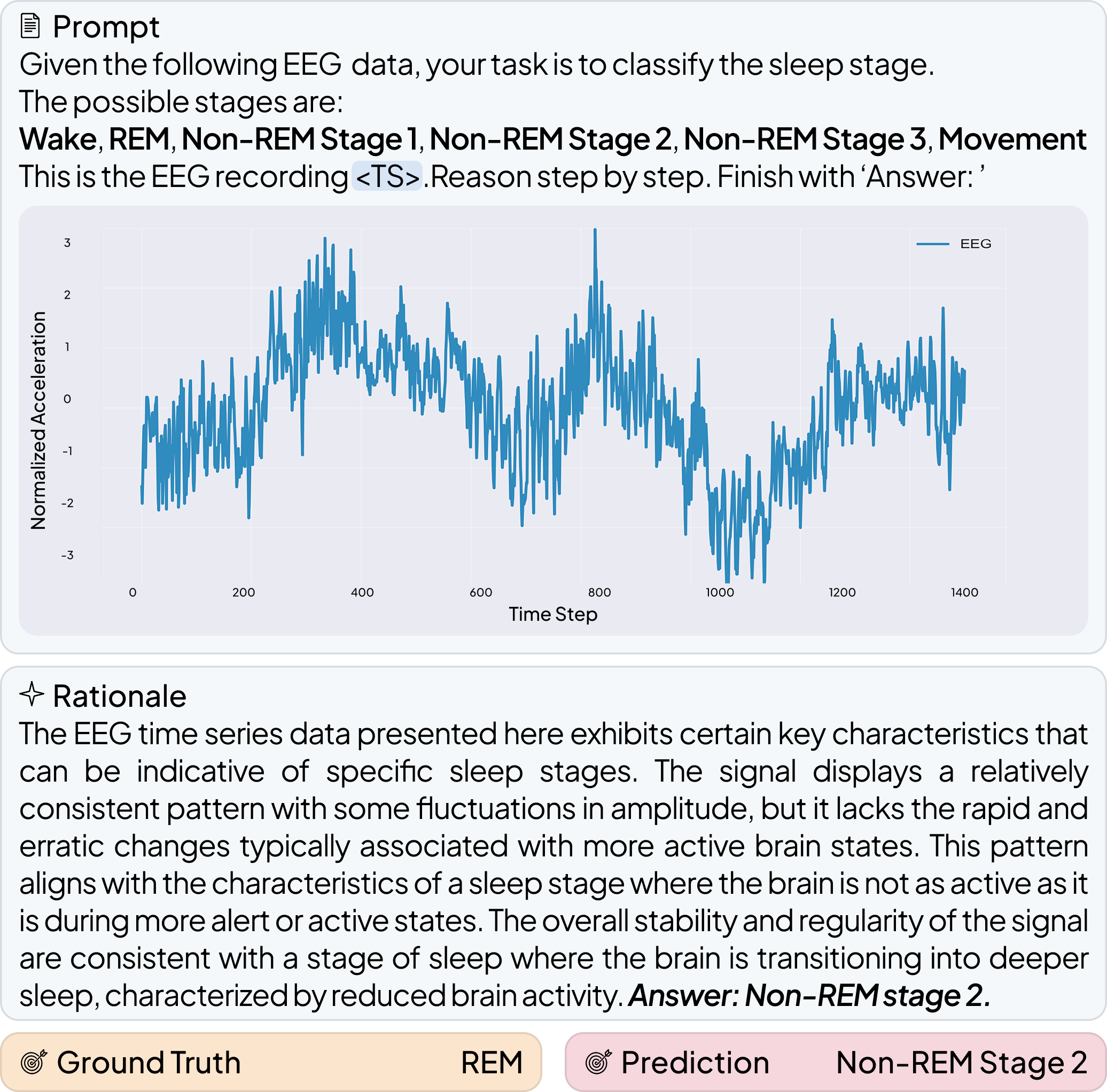}
        \caption{Stage REM Sleep}
    \end{subfigure}

    \vspace{0.5cm}

    \begin{subfigure}[b]{0.32\textwidth}
        \includegraphics[width=\linewidth]{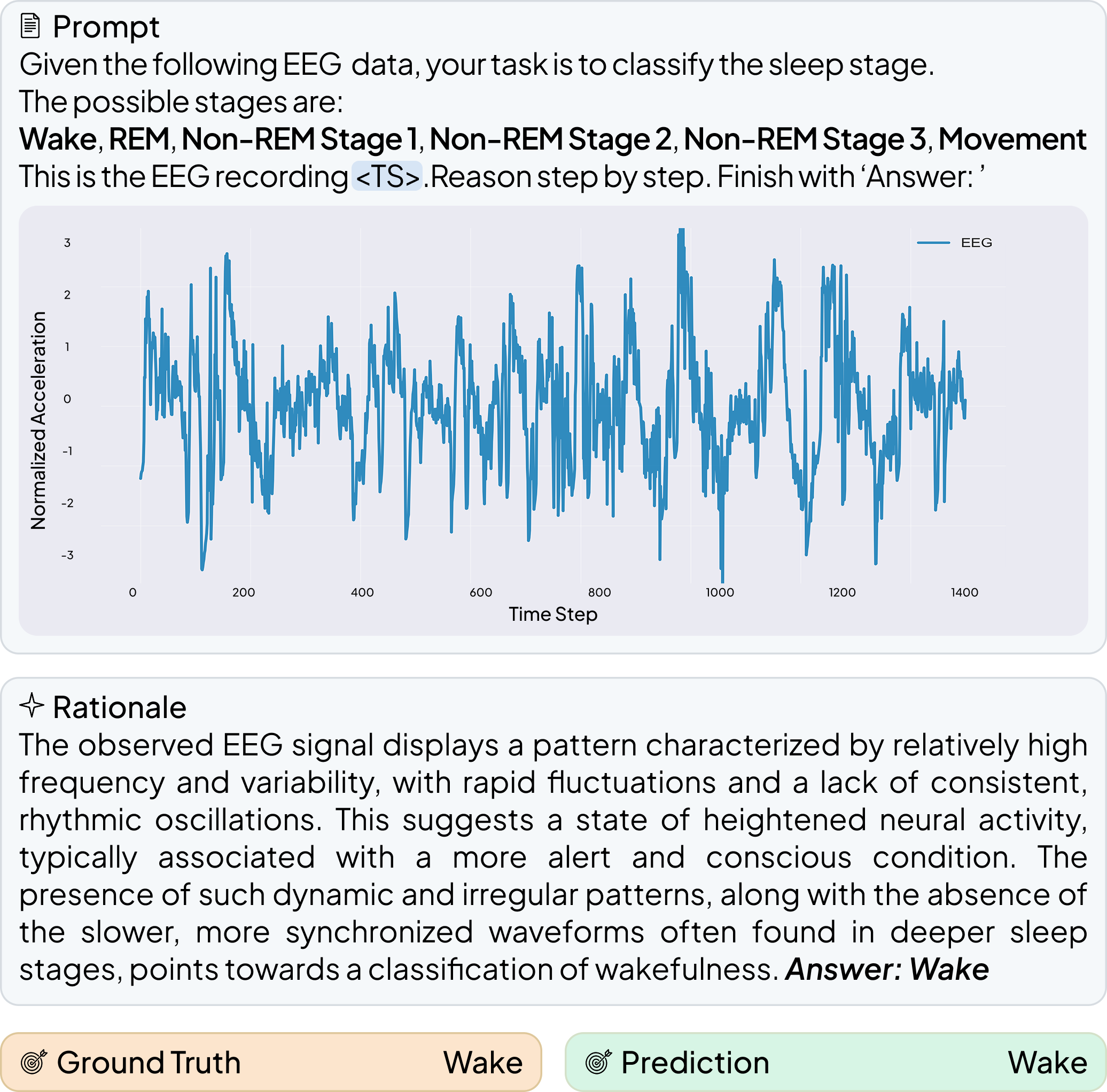}
        \caption{Stage Wake}
    \end{subfigure}
    \hfill
    \begin{subfigure}[b]{0.32\textwidth}
        \includegraphics[width=\linewidth]{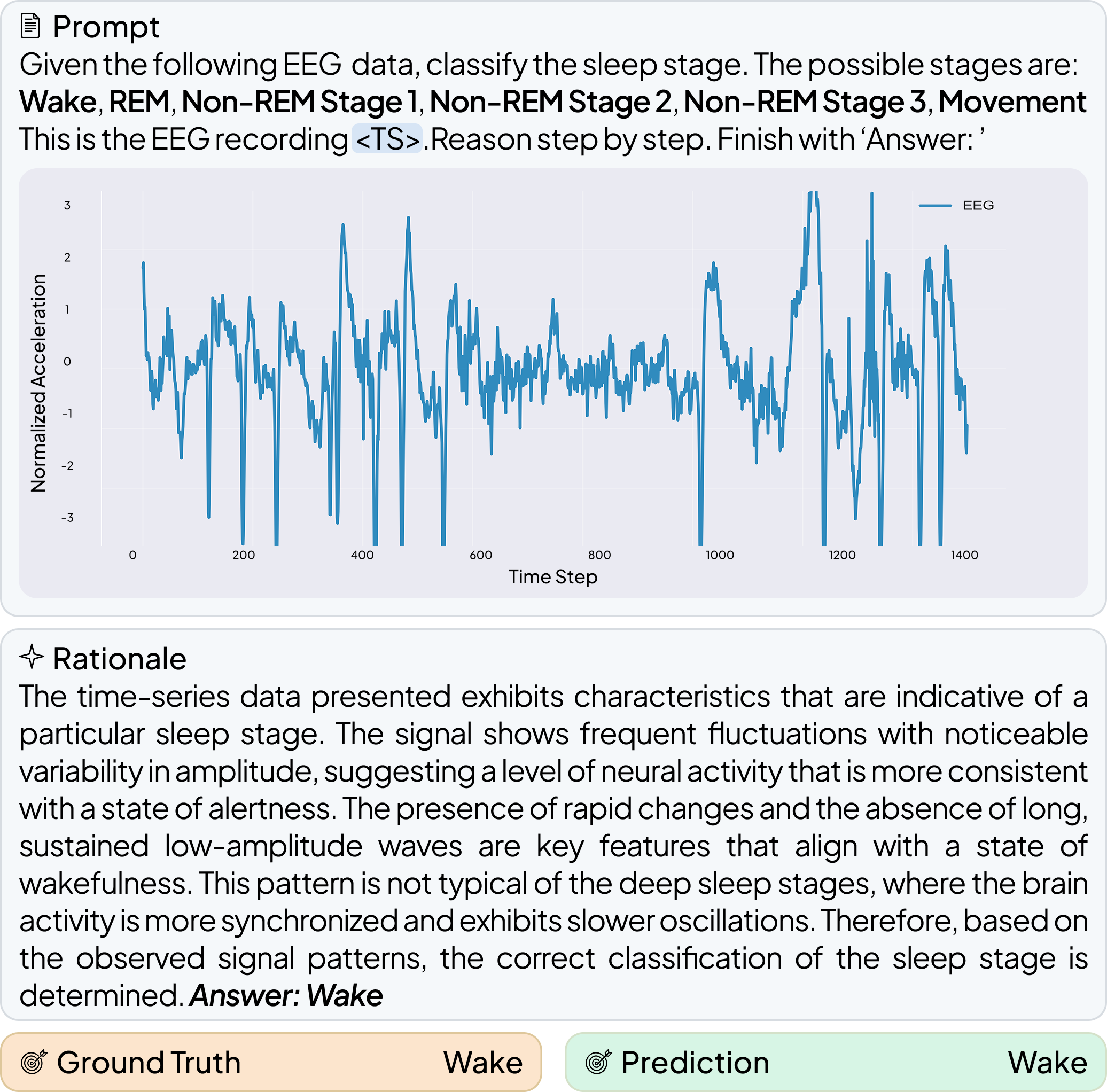}
        \caption{Stage Wake}
    \end{subfigure}
    \hfill
    \begin{subfigure}[b]{0.32\textwidth}
        \includegraphics[width=\linewidth]{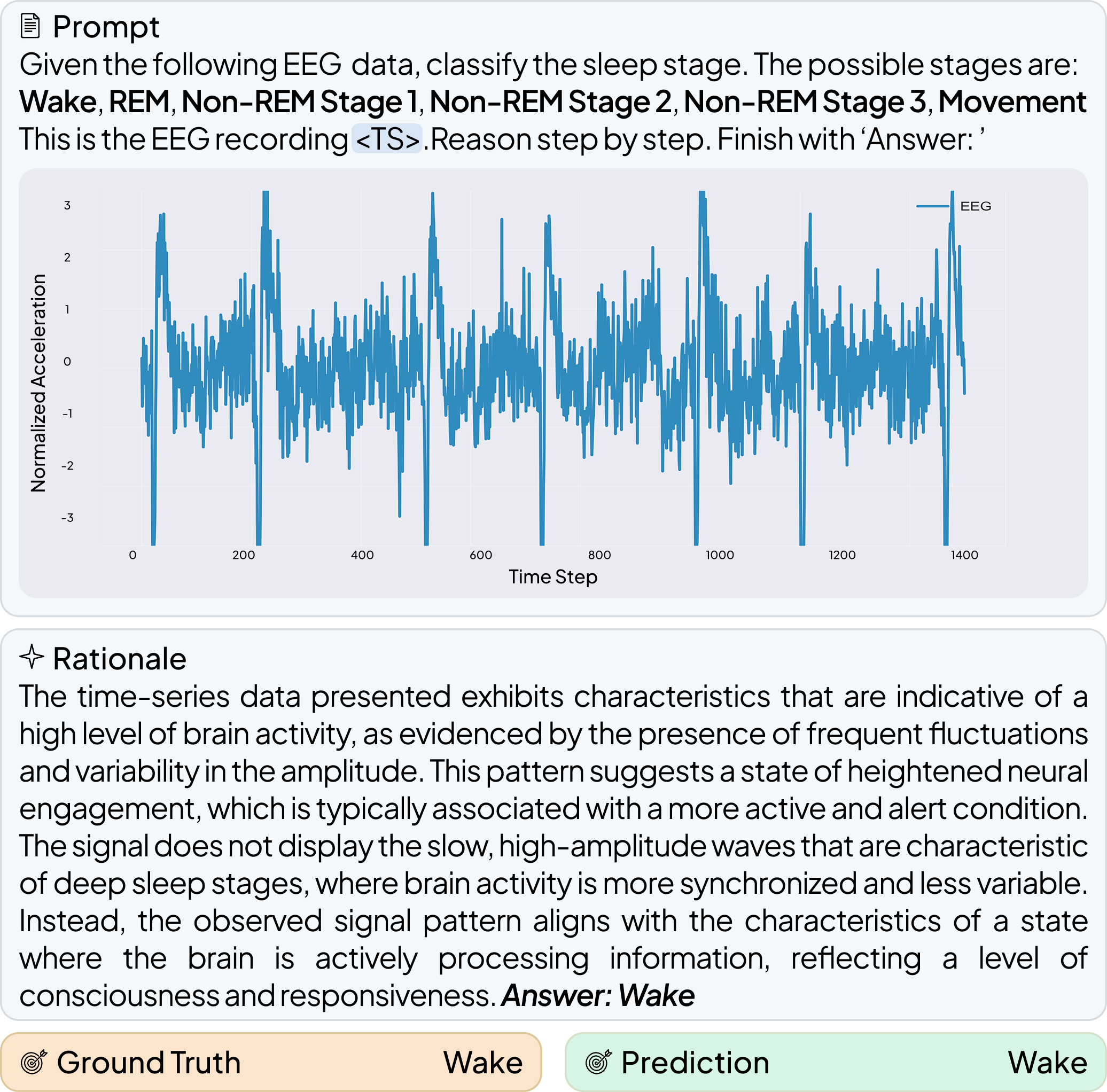}
        \caption{Stage Wake}
    \end{subfigure}

    \caption{Example rationales for sleep stage classification, generated with OpenTSLM-Flamingo/Llama3.2-1B.}
\end{figure}

\newpage
\subsubsection{ECG QA CoT}
\label{sec:appendix:ecgqacot}
\begin{figure}[!htb]
    \centering

    \begin{subfigure}[b]{0.48\textwidth}
        \includegraphics[width=\linewidth]{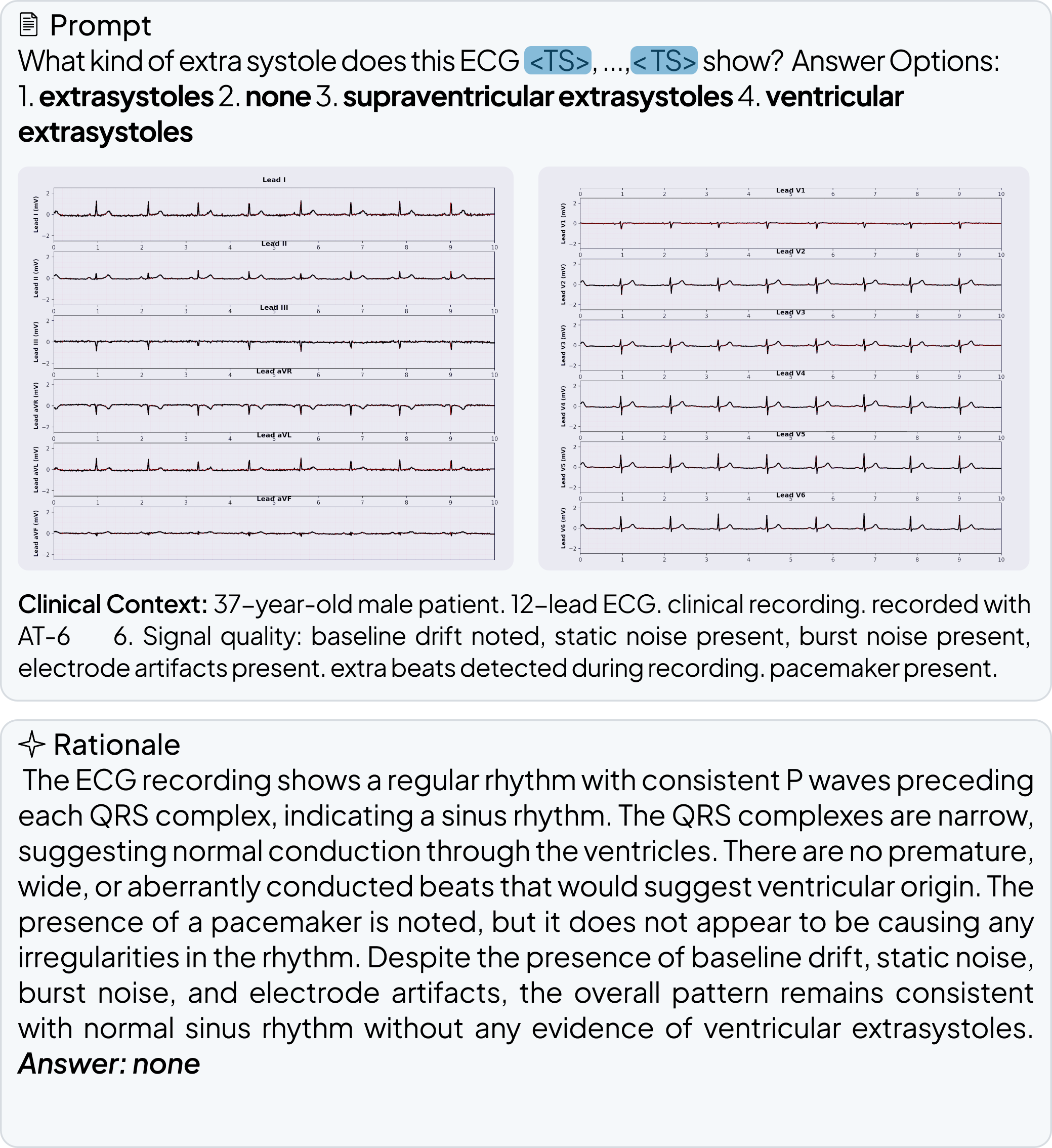}
        \caption{none (Correct Prediction)}
    \end{subfigure}
    \hfill
    \begin{subfigure}[b]{0.48\textwidth}
        \includegraphics[width=\linewidth]{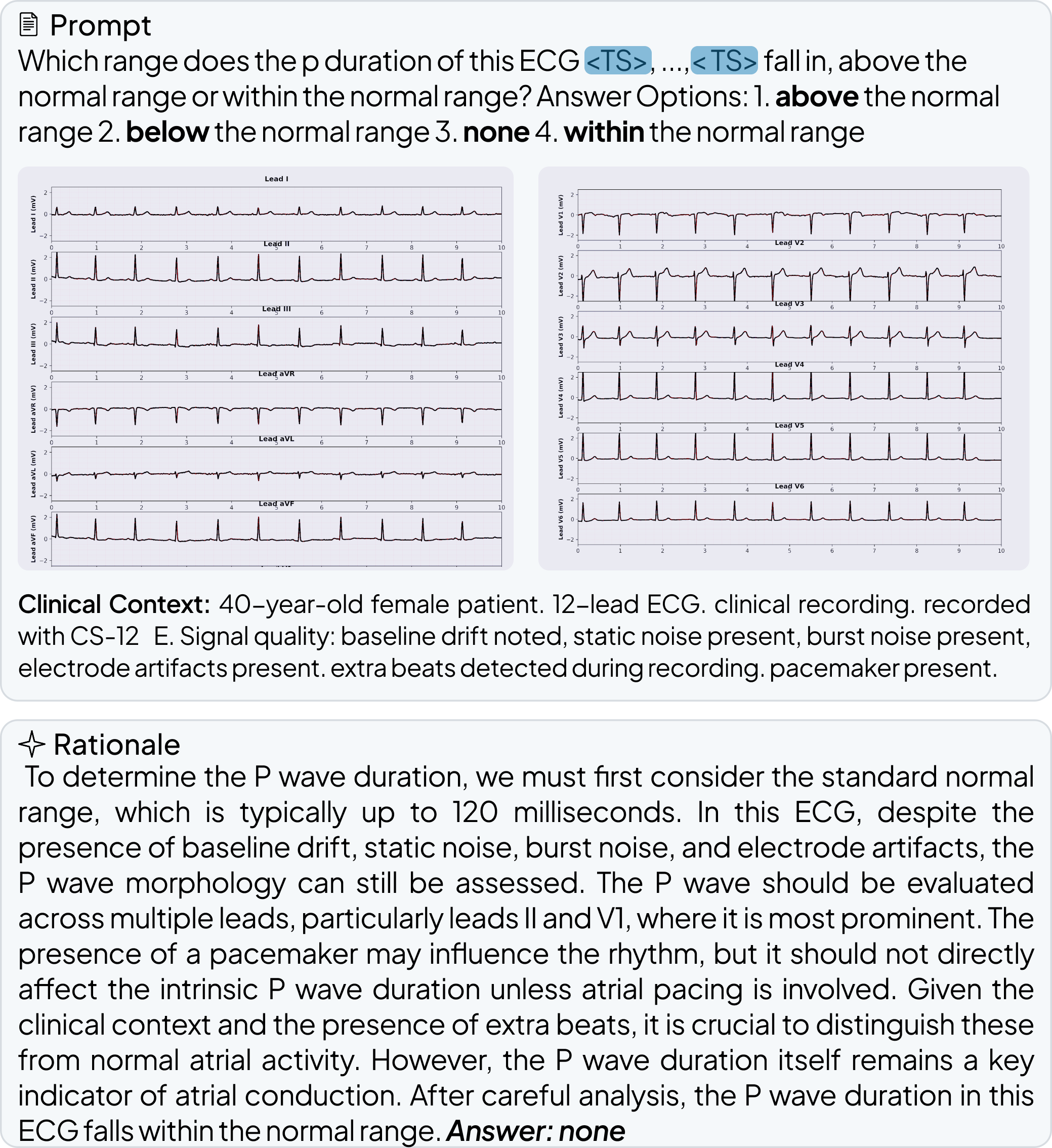}
        \caption{none (Correct Prediction)}
    \end{subfigure}

    \vspace{0.5cm}

    \begin{subfigure}[b]{0.48\textwidth}
        \includegraphics[width=\linewidth]{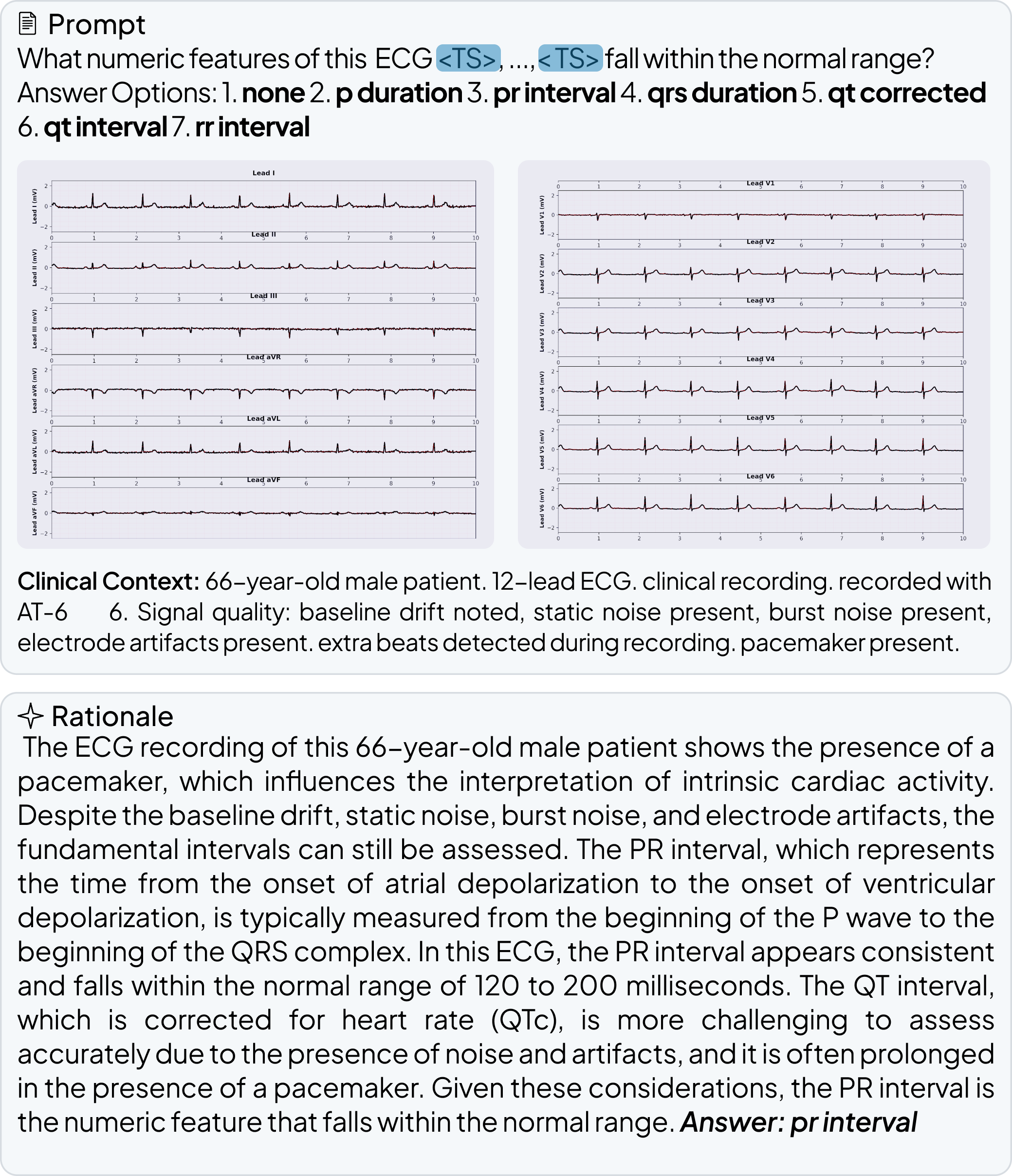}
        \caption{pr interval (Correct Prediction)}
    \end{subfigure}
    \hfill
    \begin{subfigure}[b]{0.48\textwidth}
        \includegraphics[width=\linewidth]{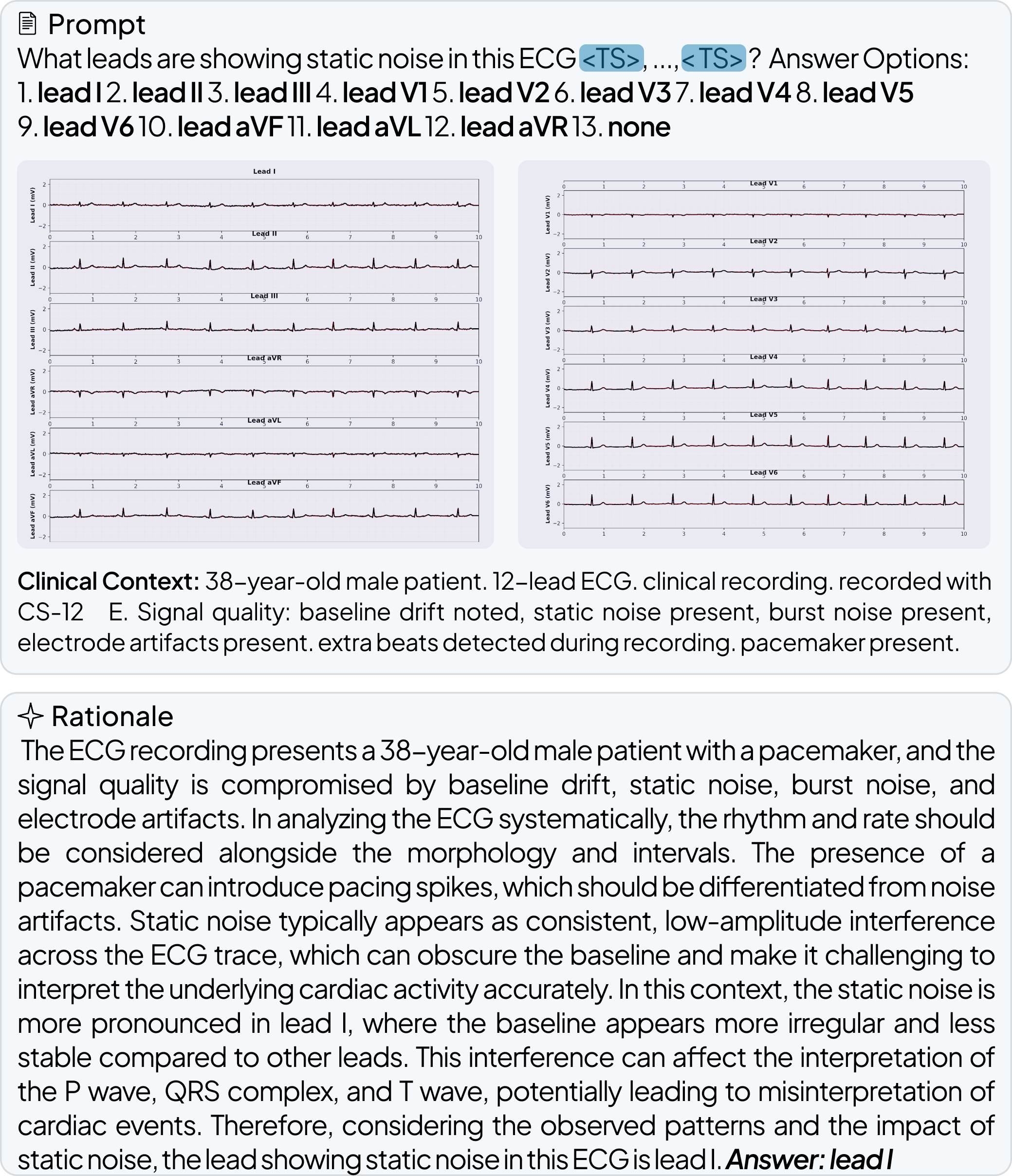}
        \caption{Lead 1 (Correct Prediction)}
    \end{subfigure}

    \caption{Example rationales for ECG QA, generated with OpenTSLM-Flamingo/Llama3.2-1B.}
\end{figure}

\newpage
\subsection{ECG Evaluation Rubric}
These are the questions asked to clinicians during evaluation of ECG-QA rationales generated by OpenTSLMFlamingo/Llama3.2-3B. See~\autoref{sec:methods:qualitative_results} for details.
\label{sec:appendix:ecgEval}
\begin{table}[h!t]
\centering
\begin{tabularx}{\textwidth}{|>{\raggedright\arraybackslash}p{4cm}|X|>{\raggedright\arraybackslash}p{4cm}|}
\hline
\textbf{Assessment Criteria} & \textbf{Description} & \textbf{Options} \\ \hline

\textbf{1. ECG Pattern Recognition Accuracy} & 
Did the model correctly identify the relevant ECG features needed to answer the question? & 
Yes; Some but not all; None identified \\ \hline

\textbf{2. Clinical Reasoning Quality} & 
Did the model appropriately connect the identified ECG features to the final answer? & 
Yes; Some incorrect logic; Completely incorrect logic \\ \hline

\textbf{3. Clinical Context Integration} & 
Did the model appropriately incorporate patient clinical background (age, recording conditions, artifacts) in its interpretation? & 
Yes; Used some key background; No did not use any relevant background \\ \hline

\end{tabularx}
\caption{Assessment Criteria for ECG Interpretation Reasoning}
\end{table}

\subsubsection{ECG Review Form }
 \begin{figure}[h!t]
 \centering
        \includegraphics[width=1\textwidth]{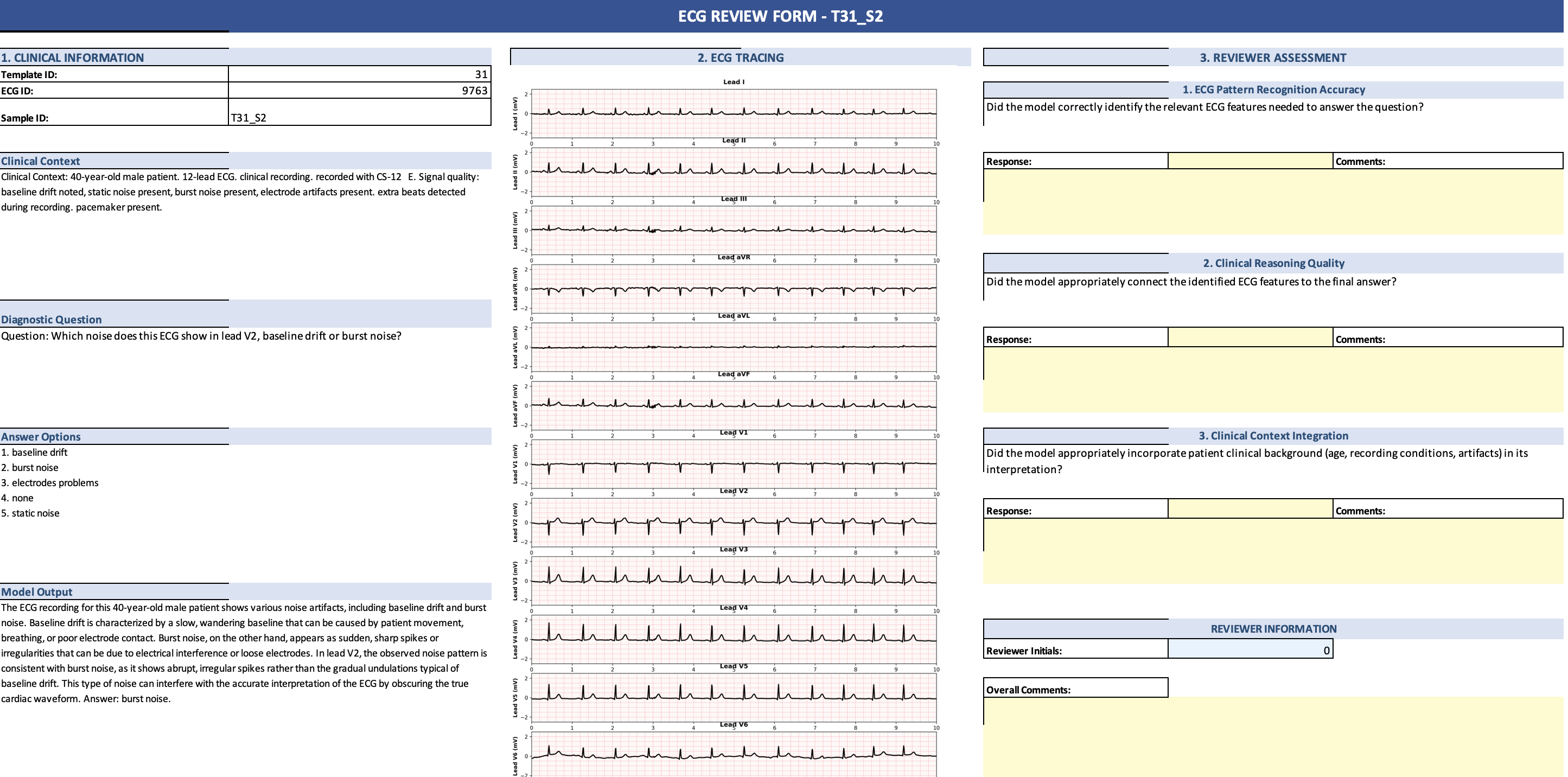}
        \caption{ECG Review Form. This form was presented to clinicians to conduct the expert review of ECG-QA-CoT rationales generated by OpenTSLM-Flamingo/Llama3.2-3B~(best model during evaluation, see~\autoref{tab:classification_results}).}
        \label{fig:ECG_review_form}
    \end{figure}

\newpage
\subsubsection{Reviewer disagreement patterns}
\begin{figure}[h!t]
 \centering
        \includegraphics[width=0.76\textwidth]{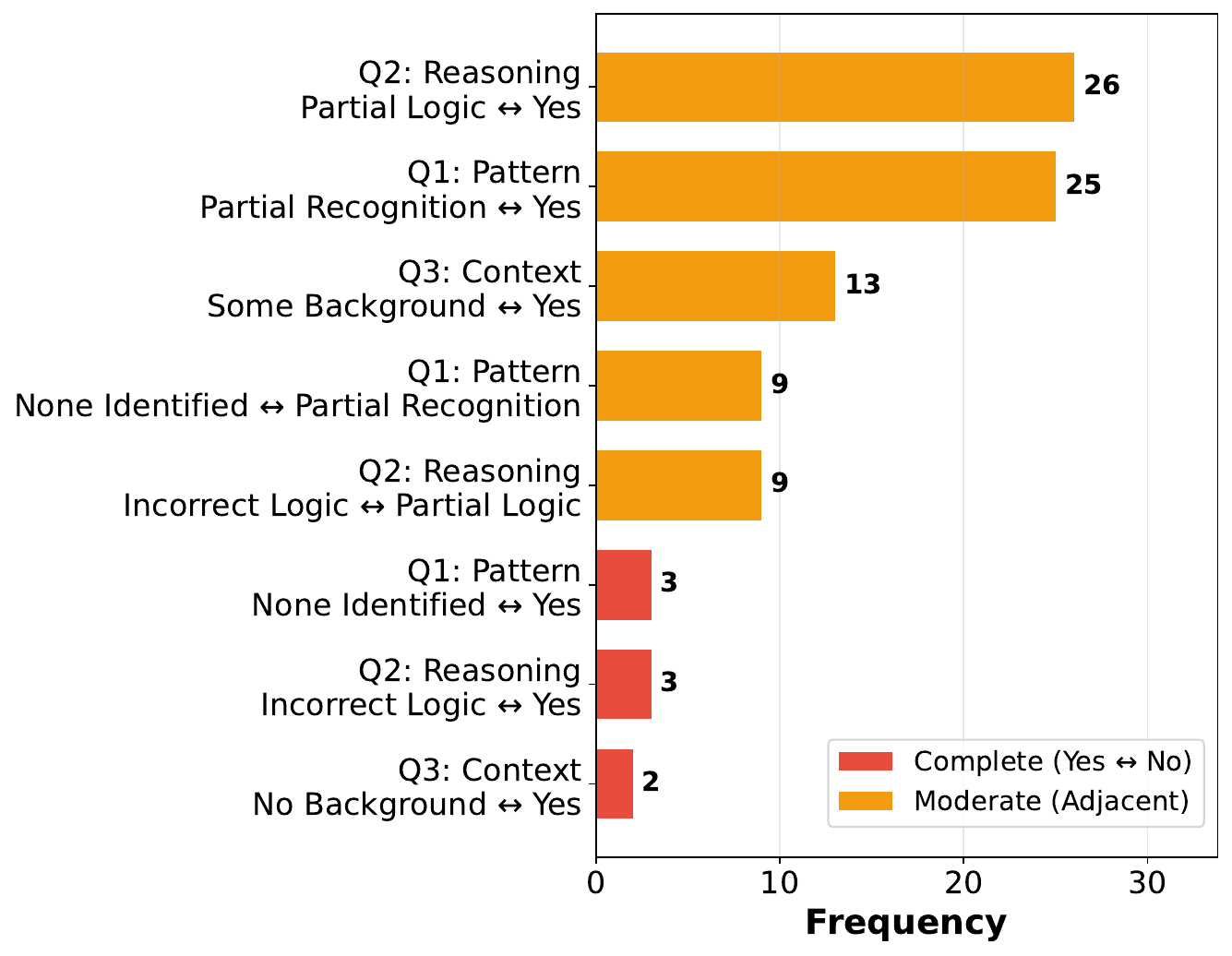}
        \caption{Disagreement Patterns}
        \label{fig:disagreement_patterns}
    \end{figure}

\autoref{fig:disagreement_patterns} shows disagreement of reviewers on generated ECG-rationales~(see \autoref{sec:methods:qualitative_results}).

\subsection{Evaluation of memory consumption}

\begin{figure}[h!t]
    \centering
    \includegraphics[width=\textwidth]{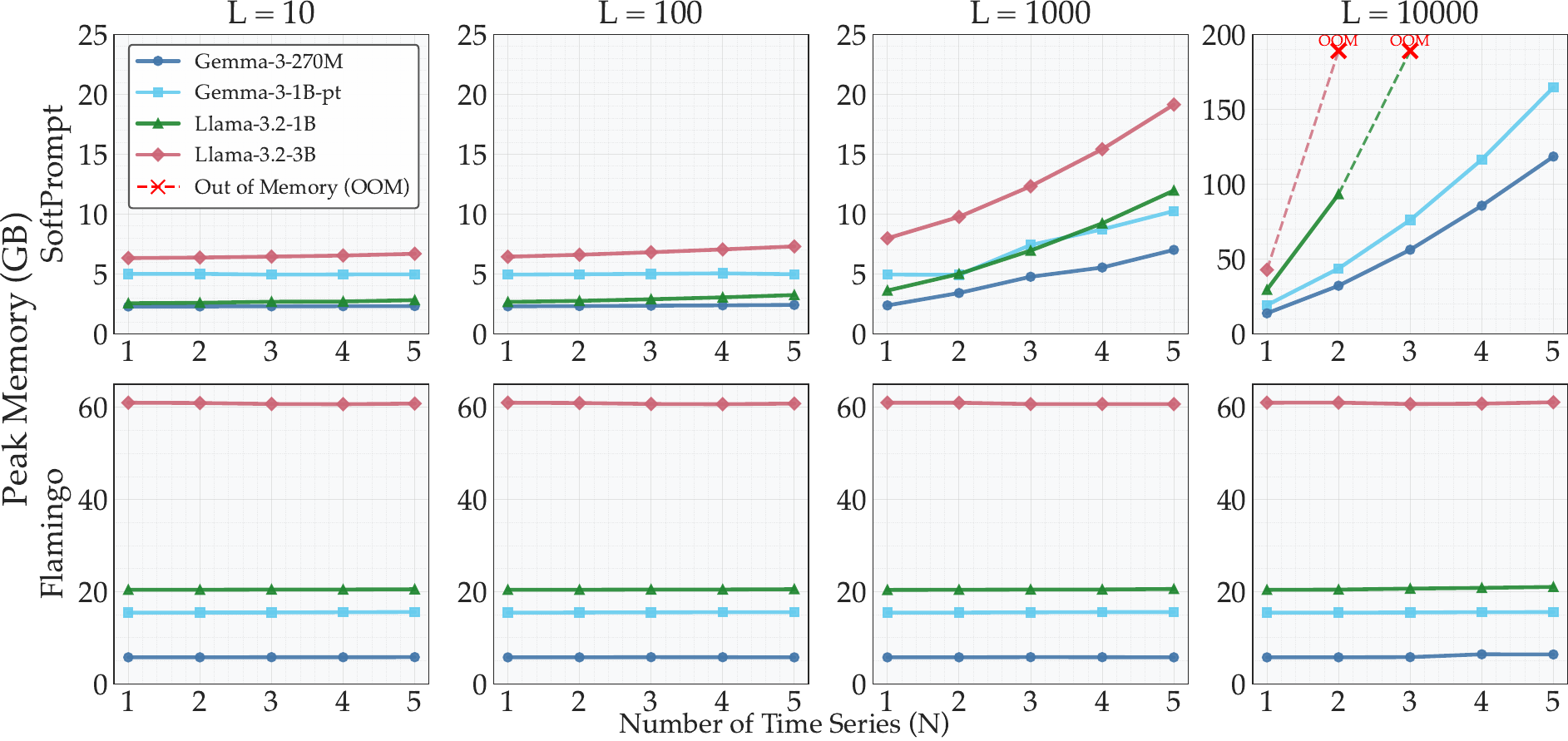}
    \caption{Simulation of memory scaling with total sequence length ($N \times L$).}
    \label{fig:memory_simulation}
\end{figure}

We complement the main results with detailed tables and plots. 
\autoref{fig:memory_simulation} illustrates scaling trends, while the 
following subsections report detailed VRAM usage for both CoT datasets 
and synthetic simulation data.

\subsubsection{Memory usage on CoT datasets}  
\autoref{tab:vram_usage_datasets} reports \gls{VRAM} for TSQA, HAR-CoT, Sleep-CoT, ECG-QA-CoT datasets. 
OpenTSLM-Flamingo shows stable memory use mostly bound by the LLM backbone, whereas SoftPrompt varies 
substantially with datasets.

\begin{table}[h!t]
    \centering
    \caption{VRAM Usage (GB) for Regular Datasets}
    \label{tab:vram_usage_datasets}
    \vspace{0.5em}

  \begin{tabularx}{\textwidth}{l l *{4}{X}}
    \textbf{Method} & \textbf{Model} 
        & \textbf{TSQA} 
        & \textbf{HAR-CoT} 
        & \textbf{SleepEDF-CoT} 
        & \textbf{ECG-QA-CoT} \\
    \midrule
    \multirow{4}{*}{\rotatebox{90}{\shortstack{OpenTSLM\\SoftPrompt}}} 
        & Llama-3.2-1B & 4.4 & 9.6 & 15.9 & 64.9 \\
        & Llama-3.2-3B & 8.1 & 14.3 & 20.3 & 87.1 \\
        & Gemma-3-270M & 2.4 & 8.6 & 20.1 & 24.1 \\
        & Gemma-3-1B-pt & 5.1 & 6.1 & 14.7 & 32.7 \\
    \addlinespace[5pt]
    \midrule
    \multirow{4}{*}{\rotatebox{90}{\shortstack{OpenTSLM\\Flamingo}}} 
        & Llama-3.2-1B & 20.5 & 22.0 & 21.6 & 20.9 \\
        & Llama-3.2-3B & 61.1 & 63.5 & 63.4 & 71.6 \\
        & Gemma-3-270M & 5.7 & 6.4 & 6.3 & 7.3 \\
        & Gemma-3-1B-pt & 15.6 & 16.3 & 15.7 & 18.4 \\
    \addlinespace[5pt]
    
  \end{tabularx}
\end{table}

\subsubsection{Memory usage for Simulation Data} 
\label{sec:appendix:memory_use_simulation}

\begin{figure}[h!t]
  \centering
  \includegraphics[width=1.0\linewidth]{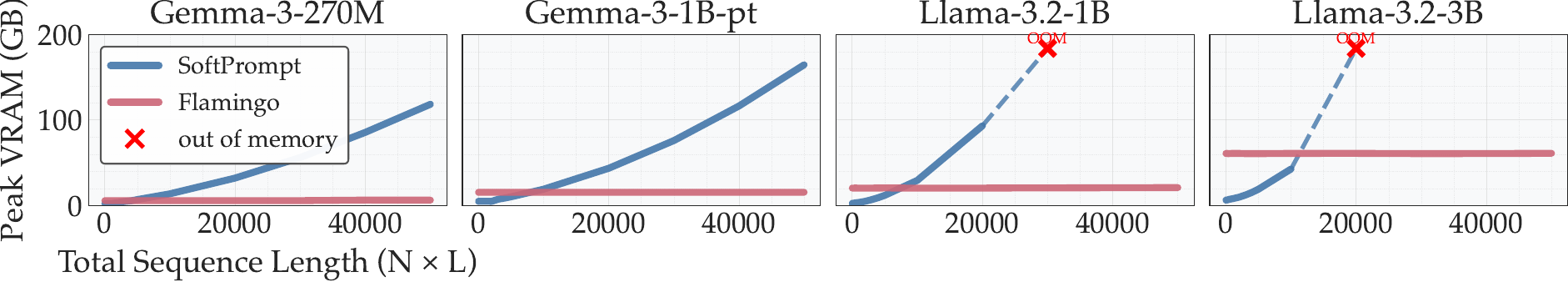}
  \caption{VRAM usage vs. total time-series size 
$N\times L$ (number of series × length)}
  \label{fig:memory_per_length}
\end{figure}

Table~\ref{fig:memory_per_length} show that \gls{VRAM} for OpenTSLM-Flamingo effectively stays constant as $N$ increases from 1 to 5 and $L$ from 10 to 10{,}000 (e.g., Llama-1B $\approx$20.4--21.0\,GB; Llama-3B $\approx$60.7--61.1\,GB; Gemma-270M $\approx$5.7--6.4\,GB; Gemma-1B $\approx$15.4--15.6\,GB). By contrast, SoftPrompt scales with both $N$ and $L$~(see \autoref{fig:memory_per_length} in \autoref{sec:appendix:memory_use_simulation}): for Llama-1B, VRAM rises from $\sim$2.6\,GB at $L{=}10,N{=}1$ to $\sim$29.5\,GB at $L{=}10{,}000,N{=}1$ and exceeds memory at $L{=}10{,}000,N{\ge}3$; Llama-3B shows a similar pattern (6.3\,GB $\rightarrow$ 42.7\,GB at $N{=}1$, OOM by $N{\ge}3$). Gemma-270M and Gemma-1B reach up to $\sim$118\,GB and $\sim$165\,GB, respectively, at $L{=}10{,}000,N{=}5$.

\autoref{tab:vram_usage_simulation} shows results for simulated datasets, using permutations of $N = [1,2,3,4,5]$ and $L= [10,100,1000,10000]$. OpenTSLM-Flamingo requires almost constant memory with varying sequence length $L$ and number of 
concurrent series $N$, while OpenTSLM-SoftPrompt grows with both until going out of memory~(OOM) for larger time-series. 

\paragraph{Simulation dataset generation.}  
To generate the simulation dataset, we generate random data with combinations of $N = [1,2,3,4,5]$ and $L= [10,100,1000,10000]$ according to the following pseudocode:

\begin{verbatim}
num_series = n
series_length = l
simulation_dataset = []
for element_id in 1..200:
    time_series_texts = []
    time_series_simulations = []
    for i in 1..num_series:
        series_i = random_normal(series_length)
        series_mean = mean(series_i)
        series_std = std(series_i)
        normalized_i = normalize(series_i)
        time_series_simualtions.append(
            normalized_i
        )
        time_series_texts.append(
            "This is a time series with mean {series_mean} "
            "and std {series_std}."
        )
    simulation_dataset.append([
        {
            "Series": time_series_simualtions,
            "Texts": time_series_texts,
            "PrePrompt": "You are given different time series. "
                         "All have the same length"
                         "of {length} data points.",
            "PostPrompt": "Predict the pattern "
            "of the time series. Answer:",
            "Answer": "This is a random pattern."
        }
    ])
\end{verbatim}

\begin{table}[h!t]
    \footnotesize
    \centering
    \caption{VRAM Usage (GB) for Simulation Datasets}
    \label{tab:vram_usage_simulation}
    \vspace{0.5em}

  \begin{tabularx}{\textwidth}{r r *{8}{X}}
    & & \multicolumn{4}{c}{OpenTSLM-SoftPrompt} & \multicolumn{4}{c}{OpenTSLM-Flamingo} \\
   
    & & \multicolumn{2}{c}{LLaMA} & \multicolumn{2}{c}{Gemma} & \multicolumn{2}{c}{LLaMA} & \multicolumn{2}{c}{Gemma} \\

    \textbf{L} & \textbf{N} & 1B & 3B & 270M & 1B & 1B & 3B & 270M & 1B \\
    \midrule
    10 & 1 & 2.6 & 6.3 & 2.3 & 5.0 & 20.4 & 61.0 & 5.7 & 15.4 \\
    10 & 2 & 2.6 & 6.4 & 2.3 & 5.0 & 20.4 & 60.9 & 5.7 & 15.5 \\
    10 & 3 & 2.7 & 6.4 & 2.3 & 4.9 & 20.4 & 60.7 & 5.8 & 15.5 \\
    10 & 4 & 2.7 & 6.5 & 2.3 & 5.0 & 20.5 & 60.7 & 5.8 & 15.5 \\
    10 & 5 & 2.8 & 6.7 & 2.3 & 5.0 & 20.5 & 60.8 & 5.8 & 15.6 \\
    100 & 1 & 2.7 & 6.4 & 2.3 & 4.9 & 20.4 & 61.0 & 5.7 & 15.4 \\
    100 & 2 & 2.8 & 6.6 & 2.3 & 5.0 & 20.4 & 60.9 & 5.7 & 15.5 \\
    100 & 3 & 2.9 & 6.8 & 2.3 & 5.0 & 20.5 & 60.7 & 5.8 & 15.5 \\
    100 & 4 & 3.0 & 7.0 & 2.4 & 5.0 & 20.5 & 60.7 & 5.8 & 15.5 \\
    100 & 5 & 3.2 & 7.3 & 2.4 & 5.0 & 20.5 & 60.8 & 5.7 & 15.5 \\
    1000 & 1 & 3.6 & 8.0 & 2.4 & 5.0 & 20.4 & 61.0 & 5.7 & 15.4 \\
    1000 & 2 & 5.0 & 9.8 & 3.4 & 4.9 & 20.4 & 61.0 & 5.7 & 15.4 \\
    1000 & 3 & 6.9 & 12.3 & 4.8 & 7.4 & 20.4 & 60.7 & 5.8 & 15.5 \\
    1000 & 4 & 9.2 & 15.4 & 5.5 & 8.7 & 20.5 & 60.7 & 5.8 & 15.6 \\
    1000 & 5 & 12.0 & 19.1 & 7.0 & 10.2 & 20.6 & 60.7 & 5.7 & 15.6 \\
    10000 & 1 & 29.5 & 42.7 & 13.7 & 19.2 & 20.4 & 61.0 & 5.7 & 15.4 \\
    10000 & 2 & 93.3 & 191.4 & 32.1 & 43.6 & 20.4 & 61.0 & 5.7 & 15.4 \\
    10000 & 3 & OOM\textsuperscript{*1} & OOM\textsuperscript{*1} & 56.1 & 76.0 & 20.6 & 60.7 & 5.8 & 15.5 \\
    10000 & 4 & OOM\textsuperscript{*1} & OOM\textsuperscript{*1} & 85.6 & 116.4 & 20.8 & 60.8 & 6.4 & 15.5 \\
    10000 & 5 & OOM\textsuperscript{*1} & OOM\textsuperscript{*1} & 118.4 & 164.5 & 21.0 & 61.1 & 6.4 & 15.5 \\
  \end{tabularx}
  \footnotesize \textsuperscript{*1} OOM: Out of memory; OpenTSLM-SoftPrompt requires more tokens for longer time-series, and separate tokens for separate time-series. Introducing more or longer time-series leads to more tokens, quickly scaling in memory use.
\end{table}


\end{document}